%% file: main.tex
\pdfoutput=1

\documentclass[11pt]{article}

\usepackage{acl}

\usepackage{times}
\usepackage{latexsym}
\usepackage{amsmath}
\usepackage{amsfonts}
\usepackage{amssymb}
\usepackage{subcaption}
\usepackage{caption}
\usepackage{graphicx}
\usepackage{booktabs}
\usepackage{multirow}
\usepackage{comment}
\usepackage[capitalize]{cleveref}
\crefname{section}{Sect.}{Sect.}
\Crefname{equation}{Eq.}{Eqs.}
\Crefname{figure}{Fig.}{Figs.}
\Crefname{table}{Tab.}{Tabs.}
\crefname{equation}{Eq.}{Eqs.}
\crefname{figure}{Fig.}{Figs.}
\crefname{table}{Tab.}{Tabs.}

\usepackage[T1]{fontenc}

\usepackage[utf8]{inputenc}

\usepackage{microtype}
\usepackage{inconsolata}
\usepackage{xcolor}
\usepackage{listings}

\usepackage{xspace}
\newcommand{\recode}{\texttt{ReCode}\xspace}

\title{\texttt{ReCode}: Robustness Evaluation of Code Generation Models}

\author{Shiqi Wang$^{1,*,\ddagger}$ \quad Zheng Li$^{2,*,\dagger}$ \quad Haifeng Qian$^1$ \quad Chenghao Yang$^{3,\dagger}$ \textbf{Zijian Wang}$^1$ \\ \quad \textbf{Mingyue Shang}$^1$  \quad \textbf{Varun Kumar}$^1$ \quad \textbf{Samson Tan}$^4$ \quad \textbf{Baishakhi Ray}$^{1}$ \quad \textbf{Parminder Bhatia}$^1$ \\ \textbf{Ramesh Nallapati}$^1$ \quad \textbf{Murali Krishna Ramanathan}$^1$ \quad \textbf{Dan Roth}$^1$  \quad \textbf{Bing Xiang}$^1$ \\
         $^{1}$AWS AI Labs \quad $^{2}$Cornell University \quad $^{3}$University of Chicago
        \quad $^{4}$ AWS AI Research \& Education \\
\texttt{wshiqi@amazon.com} \quad
\texttt{zl634@cornell.edu} \quad
\texttt{qianhf@amazon.com} \quad
\texttt{yangalan1996@gmail.com}
 \\
\texttt{\{zijwan,myshang,kuvrun,samson,rabaisha,parmib,rnallapa,mkraman,drot,bxiang\}@amazon.com}}

\begin{document}
\maketitle
\def\thefootnote{$\ddagger$}\footnotetext{Corresponding author.}\def\thefootnote{\arabic{footnote}}
\def\thefootnote{*}\footnotetext{Equal contribution.}\def\thefootnote{\arabic{footnote}}
\def\thefootnote{$\dagger$}\footnotetext{Work done while the authors were at Amazon.}\def\thefootnote{\arabic{footnote}}

\input{abstract}

\input{introduction}

\input{background}

\input{methodology}

\input{evaluation}

\input{conclusion}

\section*{Ethics Statement}

Our \texttt{ReCode} robustness benchmark aims to provide a comprehensive robustness evaluation framework for any code-generation models, which we believe is critical towards building robust and user-friendly language models for code. With the new robustness evaluation metrics, users can rely on \texttt{ReCode} and assess model predictions with more confidence. The model trainers, on the other hand, will be aware of the potential vulnerabilities that might cause mis-predictions in practice and mitigate them before deployments. Therefore, we believe our \texttt{ReCode} benchmark is beneficial in terms of broader impact.

\bibliography{anthology,custom,percy_liang_cite}

\newpage
\appendix
\input{appendix}

\end{document}

%% file: abstract.tex
\begin{abstract}
Code generation models have achieved impressive performance. However, they tend to be brittle as slight edits to a prompt could lead to very different generations; these robustness properties, critical for user experience when deployed in real-life applications, are not well understood. Most existing works on robustness in text or code tasks have focused on classification, while robustness in generation tasks is an uncharted area and to date there is no comprehensive benchmark for robustness in code generation. 
In this paper, we propose \texttt{ReCode}, a comprehensive robustness evaluation benchmark for code generation models. We  customize over 30 transformations specifically for code on docstrings, function and variable names, code syntax, and code format. They are carefully designed to be natural in real-life coding practice, preserve the original semantic meaning, and thus provide multifaceted assessments of a model's robustness performance. With human annotators, we verified that over 90\% of the perturbed prompts do not alter the semantic meaning of the original prompt. 
In addition, we define robustness metrics for code generation models considering the worst-case behavior under each type of perturbation, taking advantage of the fact that executing the generated code can serve as objective evaluation.
We demonstrate \texttt{ReCode} on SOTA models using HumanEval, MBPP, as well as function completion tasks derived from them. Interesting observations include: better robustness for CodeGen over InCoder and GPT-J;  models are most sensitive to syntax perturbations; more challenging robustness evaluation on MBPP over HumanEval.

\end{abstract}

%% file: introduction.tex
\section{Introduction}
\label{sec:introduction}

Code generation has emerged as an important AI application.
Multiple models~\citep{CodeGen, incoder, gpt-j} have been proposed and achieved impressive performance on generating code using a natural-language description, on completing partial lines and functions, and even on solving complex coding-contest problems. They can offer real-life help to software engineers and enhance their productivity, and multiple commercial offerings exist today for AI-powered code generation~\citep{chen2021evaluating}.

\begin{figure*}
    \centering

    \begin{lstlisting}[language={Python}]
    \end{lstlisting}
    \begin{subfigure}[b]{0.45\textwidth}
        \includegraphics[width=\linewidth]{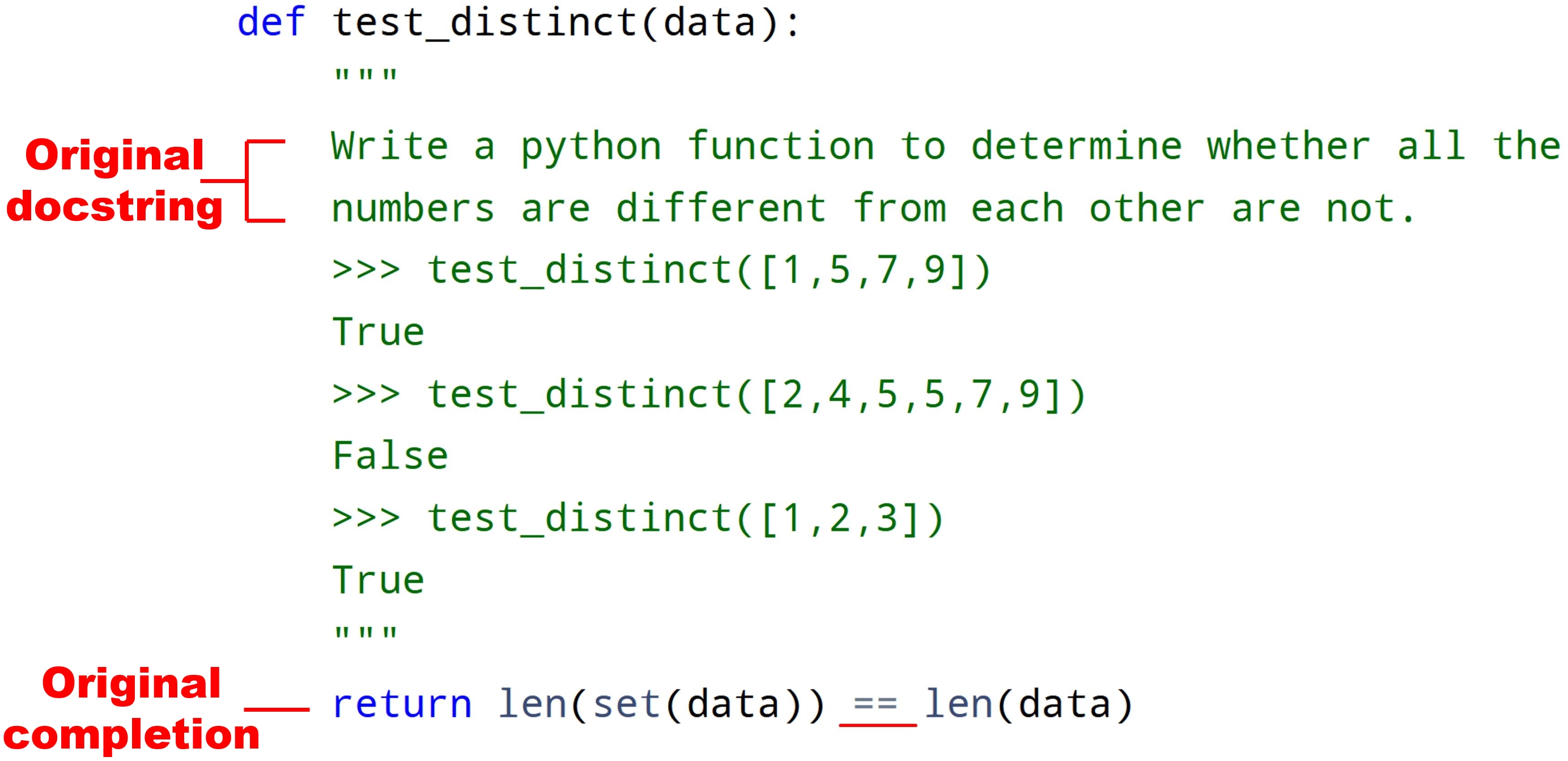}
    \end{subfigure}
    \begin{subfigure}[b]{0.39\textwidth}
        \includegraphics[width=\linewidth]{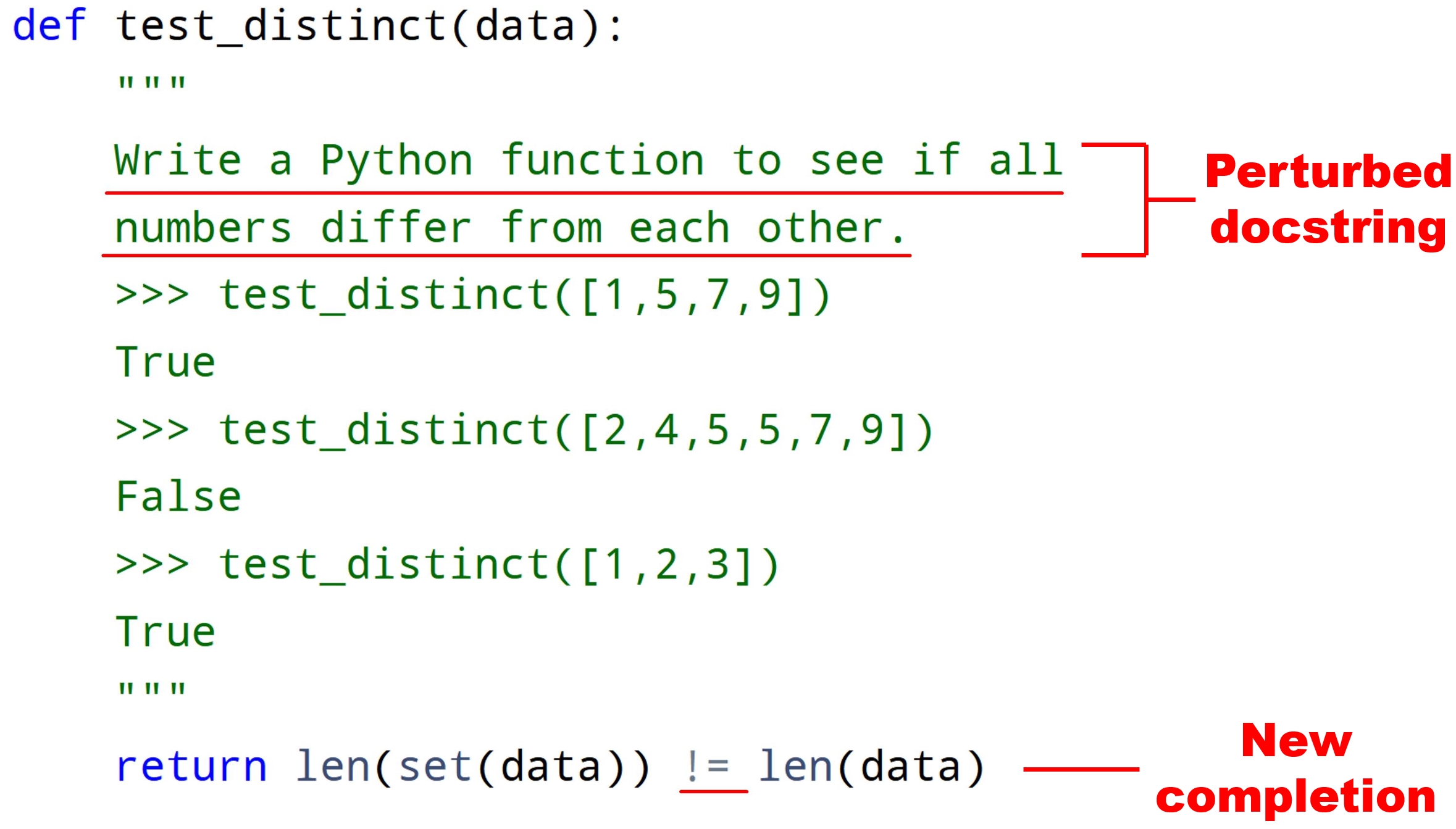}
    \end{subfigure}
    \vspace{-5pt}
        \caption{InCoder-6B predicts correctly on nominal prompt (left) but fails on the prompt where docstrings are paraphrasing with BackTranslation (right). We underline the perturbed positions and wrong model completions. }
    \label{fig: motivating_examples1}
\vspace{-5pt}
\end{figure*}
\begin{figure*}
    \centering

    \begin{subfigure}[b]{0.46\textwidth}
        \includegraphics[width=\linewidth]{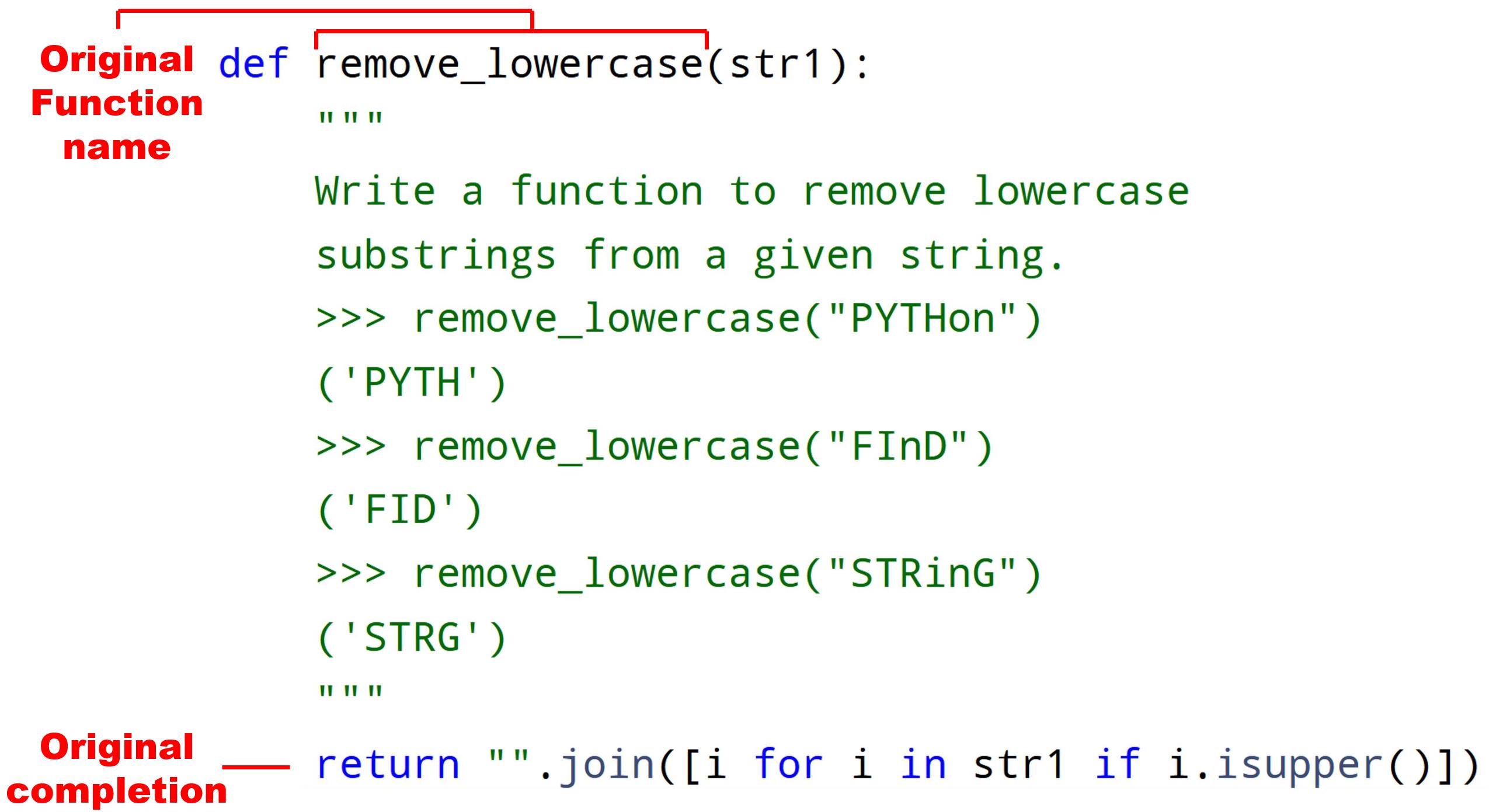}
    \end{subfigure}
    \begin{subfigure}[b]{0.38\textwidth}
        \includegraphics[width=\linewidth]{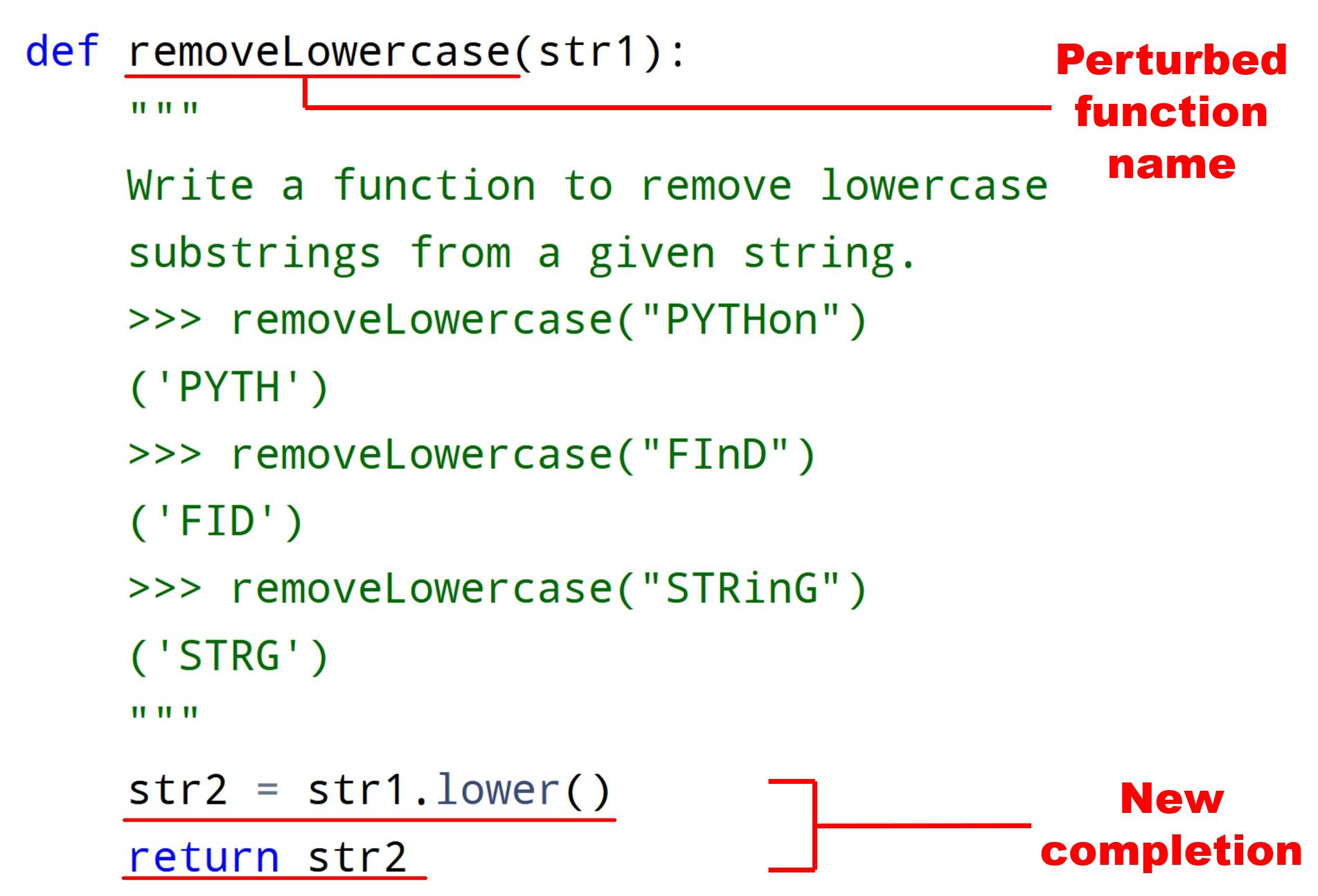}
    \end{subfigure}
    \vspace{-5pt}
        \caption{CodeGen-16B-mono is correct on nominal prompt (left) but fails when function name is perturbed (right).}%
    \label{fig: motivating_examples2}
    \vspace{-1em}
\end{figure*}

However, one important aspect, robustness of the code generation models, is commonly overlooked.
Anecdotally, people know that these models are sensitive to perturbations over prompts: sometimes just an extra space in a line or a slight change to a function name would lead to completely different generations, with potentially negative impacts to usability.
In \cref{fig: motivating_examples1} and \cref{fig: motivating_examples2}, we show two failure cases on InCoder-6B~\citep{incoder} and CodeGen-16B-mono~\cite{CodeGen} where
they perform correctly on regular prompts but fail on our perturbed ones after docstring paraphrasing and function camel case renaming in our \texttt{ReCode} benchmark. The perturbed prompts are natural and retain the original meaning, indicating weakness of these models if deployed in real-life applications.

There exists no comprehensive and quantitative robustness benchmark for code generation models. \citet{li2022competition} includes a brief study on robustness but it has limited perturbation types and is in a setting with massive numbers of samples, unrealistic in practice. 
Other existing works on robustness in text or code tasks have focused on classification and are not directly applicable to code generation~\cite{zhang2020adversarial,jha2022codeattack}.

In this paper, we present \textbf{\recode}, a \textbf{R}obustness \textbf{E}valuation framework for \textbf{Code}, aiming to provide comprehensive assessment for robustness of code generation models. ReCode includes only transformations that (1) appear naturally in practice and (2) preserve the semantic meaning of the original inputs.
We carefully collect and customize a comprehensive list of natural transformations on docstrings, function and variable names, code syntax, and code format, providing multifaceted assessments of a model’s robustness performance. We verify the quality
of the perturbed data using both human evaluation and objective similarity scores.
We take advantage of the fact that executing the generated code can serve as objective evaluation and define three robustness evaluation metrics that aggregate a model's correctness across randomized transformations and transformation types.
These metrics quantify a model's accuracy on perturbed prompts, its relative accuracy drop from original prompts, as well as its general instability.

We summarize our contributions below:
\begin{itemize}
    \item We present the first robustness evaluation benchmark \texttt{ReCode} for code generation tasks. Our evaluation framework  is general and can be easily extended to any code generation datasets and models. \footnote{Code and datasets released at \url{https://github.com/amazon-science/recode}.}
    
    \item We collect and customize over 30 natural transformations from the aspects of docstrings, function and variable names, code syntax, and code format. Human evaluation shows that most of the perturbed prompts do not alter the semantic meaning
    and that their level of naturalness is close to the originals. Quantitative similarity metrics confirm the same.
    
    \item We propose robustness evaluation metrics for code-generation tasks: Robust Pass$_{s}$@k, Robust Drop$_{s}$@k, and Robust Relative$_{s}$@k.
    
    \item We demonstrate the \recode benchmark on HumanEval
    and MBPP
    datasets and present extensive empirical robustness comparisons on state-of-the-art models including CodeGen, InCoder, and GPT-J across different sizes. We find that 1) diverse pretraining corpus and larger model size can help improve the model worst-case robustness, but models may learn to generalize in a non-robust way; 2) code generation models are most sensitive to syntax perturbations; %
    3) due to diversity, MBPP poses greater changes than HumanEval.

\end{itemize}

%% file: background.tex
\section{Related Work}
\label{sec: background}

\paragraph{Robustness for NLP.}

Recent research have identified the severe robustness problem in Pretrained Language Models (PLMs) using adversarial examples. For example, PLMs can be easily fooled by synonym replacement  \cite{Jin2019textfooler,zang2020word}. 
To better illustrate the severity of adversarial robustness problems for NLP models, and encourage people to explore more to build robust and trustworthy models, existing works~\citep{nie2020adversarial, gardner2020evaluating, kiela2021dynabench, wang2021adversarial} build robustness benchmark. 
\citet{zhang2020adversarial} presents a comprehensive overview of works in this field.
Most existing works in this field focuses on \textbf{classification tasks} rather than \textbf{generation tasks}. The main challenge for benchmarking robustness over generation tasks is that the evaluation of text generation is highly subjective and is usually hard to quantify. However, code generation provides a special opportunity because we can do objective and quantitative evaluation on generated codes, and code generation models use similar model architecture as NLP models.

\paragraph{Robustness for code.}
There are a series of previous work on different aspects of robustness problems for code. Specifically, \citet{bielik2020adversarial} studies the adversarial robustness problem for type inference in programming languages. \citet{yang2022natural} focuses on improving the naturalness of adversarial examples in code vulnerability prediction, clone detection and authorship attribution. \citet{zhou2022adversarial} focuses on the adversarial robustness problems of source code comment generation and \citep{jha2022codeattack} focuses on code translation, repair and summarization. These papers mainly focus on proposing attack and defense methods for different tasks in code domain, but there is no previous work on a comprehensive robustness benchmark for code generation domain. 

\paragraph{Code generation.}
Code generation, also known as program synthesis, is a task of generating code based on natural language statements or code from context. Researchers have adapted transformer-based large language models to the code generation field. Various architectures have been explored: For example, CodeBERT~\citep{feng2020codebert}, PLBART~\citep{ahmad2021unified}, CodeGPT~\citep{lu2021codexglue} explores BERT, BART and GPT architectures for language models pretrained on code corpus.
There are also works that propose to incorporate code structures for models to better understand the semantic information, including GraphCodeBERT~\citep{guo2020graphcodebert} and CodeT5~\cite{wang-etal-2021-codet5}.
Most recently, models with much larger size (i.e., billion-scale parameter numbers) are shown to significantly improve the performance on code generation benchmarks. Codex-12B~\citep{chen2021evaluating} and CodeGen-16B~\citep{CodeGen} are two representative very large pretrained code generation models and have established new state of the arts. However, few works have systematically explored robustness in code generation.

%% file: methodology.tex
\section{Methodology}
\label{sec: methodology}

In this section, we introduce the transformations to perturb prompts to both text (docstring) and code. We then propose new robust evaluation metrics.

\subsection{Problem Formulation}

We consider the end-to-end model-based code generation task.
The input prompt can include natural language statements that describe the functionality, signature of the function to generate, helper functions, and possibly a half-written function.
The goal is left-to-right generation that creates or completes the function.
This setting is agnostic to model architectures and is applicable to encoder-decoder or decoder-only models.

We perturb the input prompt with transformations.
We focus on natural transformations that preserve the semantic meaning of the original prompt and that are likely to appear in practice, e.g., frequent typos in docstrings, tab to four spaces, function name style changes, and many more.
We do not consider adversarial attacks that require model feedbacks in this paper because it is non-trivial to control the naturalness of adversarial attacks and they often require higher computational cost.
Instead, we randomly generate perturbed prompts based on the restrictions for each type of perturbations and propose new metrics to evaluate model robustness based on these prompts.
We leave adversarial attacks for future work.

\begin{table*}[t]
\footnotesize
\centering

\setlength{\tabcolsep}{1pt}
\scalebox{0.8}{
\begin{tabular}{l|r} \toprule
Perturbations & \multicolumn{1}{c}{MBPP Docstrings}                                                                             \\ \midrule
Nominal                      & Write a function to find all words which are at least 4 characters long in a string by using regex.          \\
BackTranslation              & Write a function to find all words \textbf{in a string at least 4 characters long} using regex.                       \\
ButterFingers    & Wri\textbf{h}e a function to find all words which are a\textbf{r} leas\textbf{v} 4 characters long in a string by using regex.          \\
ChangeCharCase               & Wri\textbf{T}e a f\textbf{U}ncti\textbf{O}n to find All wo\textbf{R}ds whic\textbf{H} are at le\textbf{A}st 4 \textbf{C}ha\textbf{R}acter\textbf{S} \textbf{L}on\textbf{G} in a string by u\textbf{SI}ng re\textbf{G}ex.          \\
EnglishInflectionalVariation & \textbf{Writes} a \textbf{functions} to \textbf{found} all \textbf{word} which \textbf{was} at least 4 \textbf{character} long in a string by use regex.  \\
SwapCharacters   & \textbf{rW}ite a function to find all words which are at \textbf{el}ast 4 ch\textbf{ra}acters long in a string by \textbf{su}ing regex. \\
SynonymInsertion             & Write a function to find \textbf{discover} all words which are at least 4 characters long in a string by using regex. \\
SynonymSubstitution          & Write a function to find all words which \textbf{equal} at least 4 character long in a \textbf{chain} by using regex.          \\
TenseTransformationPast      & Write a function to find all words which \textbf{was} at least 4 characters long in a string by using regex.          \\
TenseTransformationFuture    & Write a function to find all words which \textbf{will be} at least 4 characters long in a string by using regex.      \\
Whitespace       & Write a function to find all words \textbf{w h}ic\textbf{ha}re at least 4 characters long in a string by using regex.
\\ \bottomrule
\end{tabular}
}
\vspace{-10pt}
\caption{Illustrations for docstring perturbations on a MBPP sample.}
\label{tab: docstring_example}
\vspace{-10pt}
\end{table*}

\subsection{Natural Transformations on Docstrings}
\label{subsec: text}

Docstring 
describes the target function to generate.
Since docstrings
can vary greatly when written by different users, robustness against changes in docstrings is critical for usability in applications.

For docstrings, we use the NL-Augmenter~\cite{dhole2021nl} library which is designed for data augmentation and robustness evaluation on text.\footnote{\scriptsize{\url{https://github.com/GEM-benchmark/NL-Augmenter}}}
We carefully select ten transformations, including character-level, word-level and sentence-level ones, that are likely to preserve semantic similarity.
The selected perturbations include \texttt{CharCaseChange}, where random characters are replaced with their upper cases, \texttt{SynonymSubstitution}, where random words are substituted with their WordNet synonyms~\cite{miller1995wordnet}, \texttt{BackTranslation}, where sentences are translated to a different language (e.g., German by default) then back to English for paraphrasing the whole sentence~\cite{li2019improving,sugiyama2019data}, and more.
To perform perturbations, we extract docstring sentences from the input prompt and then put the perturbed version back to the prompt. See \cref{appd: transformation} for details.

We observe that directly applying NL-Augmenter to docstrings without constraints can potentially lead to low quality due to keywords in the programming languages. 
For example, "Create a list a[][]" could be perturbed by 
"Create a list \textbf{[a][]}" by character case swap, which is not natural. Therefore, to guarantee naturalness of perturbations, we use tree-sitter to parse the whole code snippet (the prompt \& the canonical solution) to extract any existing function names, variable names ("a"), and type names ("list"). We then exclude them from being perturbed by the transformations. In \cref{tab: docstring_example}, we list all ten transformations that are customized from NL-Augmenter and are included in our robustness benchmark along with sample illustrations.

\begin{figure}[!hbt]
    \begin{subfigure}[b]{0.33\textwidth}
        \includegraphics[width=\linewidth]{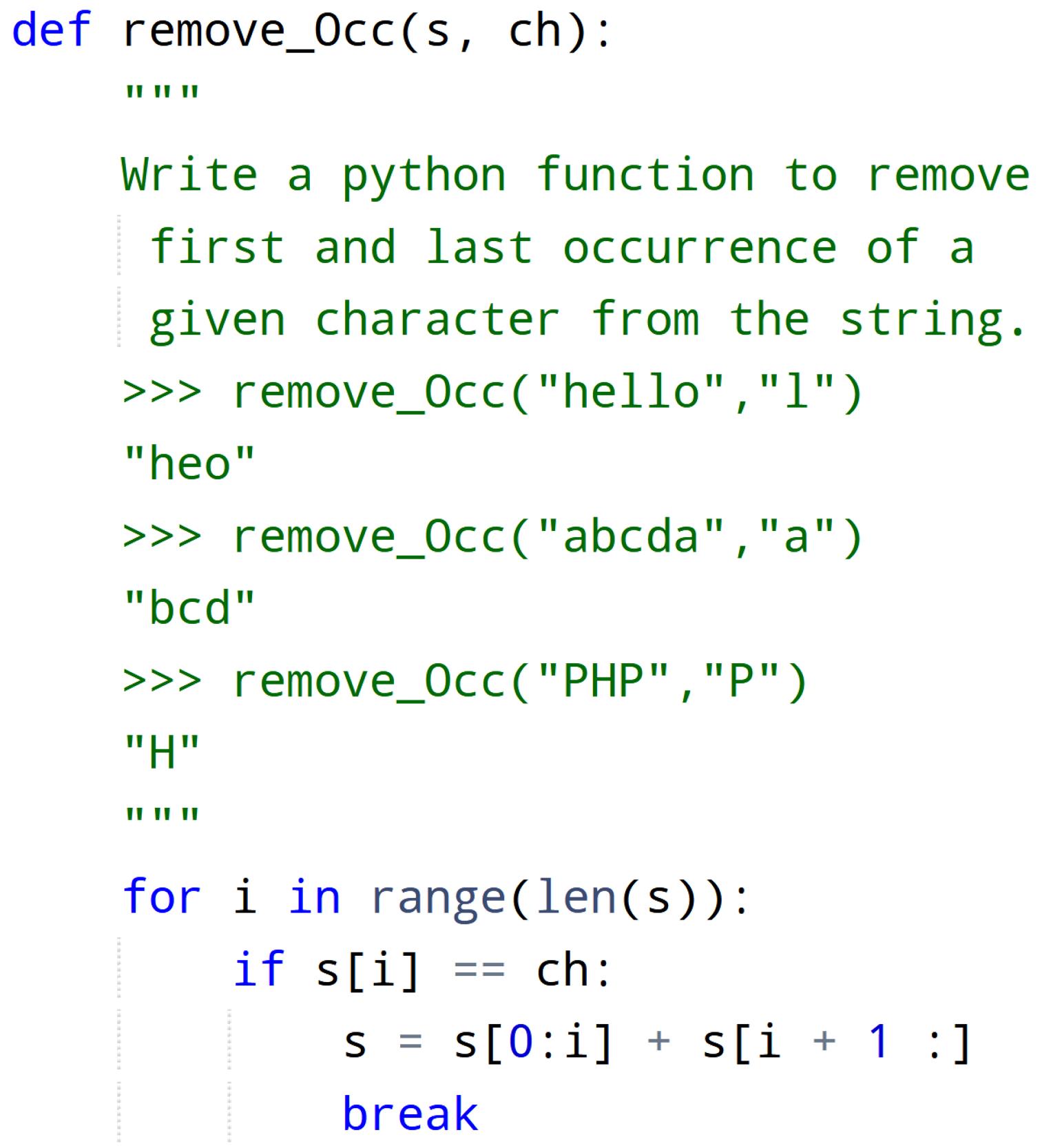}
            \caption{Baseline Partial Code}
            \label{subfig: baseline_parital}
    \end{subfigure}
    \begin{subfigure}[b]{0.3\textwidth}
        \includegraphics[width=\linewidth]{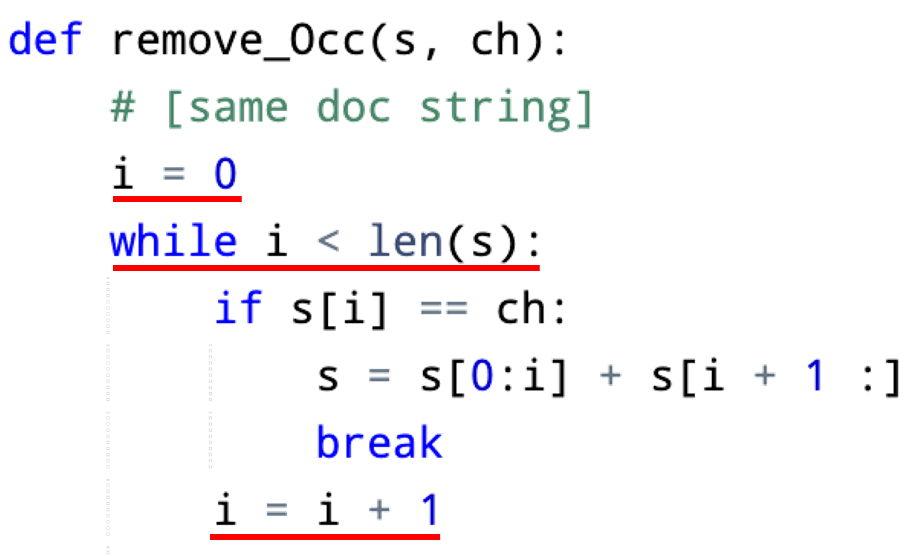}
                \caption{For-While Switch}
    \end{subfigure}
    \begin{subfigure}[b]{0.42\textwidth}
        \includegraphics[width=\linewidth]{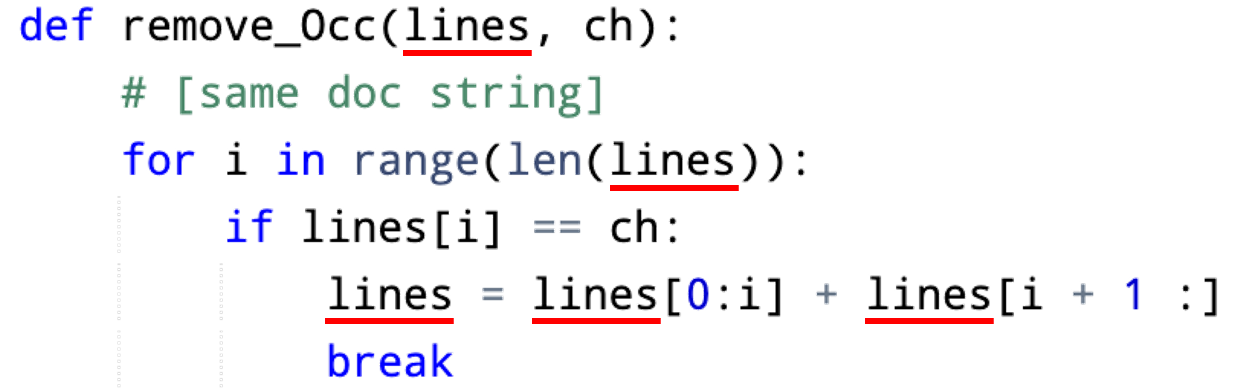}
                \caption{Variable Renaming with CodeBERT}
    \end{subfigure}
    \vspace{-10pt}
    \caption{An original prompt with partial code (a) and its perturbed versions (b, c).}
    \label{fig: natgen_example}
\end{figure}

\subsection{Natural Transformations on Function Names}
\label{subsec: func}

Perturbing function names also results in performance drops for code generation models. We summarize our perturbations in \cref{tab: func_examples}.

Some perturbations switch function names between naming conventions.
For example, the perturbation called \texttt{CamelCase} transform function names between camel-case (e.g., "findCharLong") and snake-case ("find\_char\_long").

Other perturbations apply character-level or word-level natural text transformations on component words in a function name, including \texttt{ChangeCharCase}, \texttt{InflectionalVariation}, and \texttt{SynonymSubstition} as discussed in \cref{subsec: text}. %

\begin{table}[!hbt]
\footnotesize
\centering

\scalebox{0.80}{
\begin{tabular}{l|r} \toprule
Perturbations on Function Names                           & \multicolumn{1}{c}{MBPP } \\ \midrule
Nominal                                 & find\_char\_long                       \\ \midrule
CamelCase             & \textbf{findCharLong}                  \\
ButterFingers         & fin\textbf{f}\_char\_long                       \\
SwapCharacters       & find\_c\textbf{ah}r\_long                       \\
ChangeCharCase     & fin\textbf{D}\_cha\textbf{R}\_long                       \\
InflectionalVariation & \textbf{\textbf{found}\_\textbf{chars}\_long}            \\
SynonymSubstition    & \textbf{discover}\_char\_long          \\ \bottomrule
\end{tabular}
}
\vspace{-10pt}
\caption{Illustrations for function name perturbations on a MBPP sample.}
\label{tab: func_examples}
\vspace{-1em}
\end{table}

\subsection{Natural Transformations on Code Syntax}
\label{subsec: code_syntax}

Code generation models are often used on function completion task where the prompt includes a partial implementation of the target function and the goal is to complete it.
In such scenarios, the partial code in prompt is work in progress and can be subject to frequent editing, and ideally a model should be robust with respect to perturbations in the partial code.
For this evaluation, we derive new customized datasets from HumanEval and MBPP by adding half\footnote{add first $\lfloor k/2 \rfloor$ lines given a $k$-line canonical solution.} of the canonical solutions to the prompts (\cref{subfig: baseline_parital}).
Then we perturb such partial code inside prompts.
Details and examples for each perturbations can be found in \cref{appd: transformation}.

Transformations on partial code must be syntactically correct and must not alter semantic meaning.
The next section will address code format, and let us first focus on code refactoring: these are syntactic changes that are semantically invariant. %

We adopt three transformations from NatGen~\cite{chakraborty2022natgen}: (1) \texttt{Deadcode Insertion} where dummy loops (0 iterations) or if conditions are randomly inserted;
(2) \texttt{Operand Swap} where we randomly swap one operation (e.g., \texttt{a<b} to \texttt{b>a}); (3) \texttt{For-While Switch} where we randomly transform one for-loop structure in code to equivalent while-loop structure and vice versa.

Additionally, we implement three different schemes of variable renaming.
We select the most frequent variable in the partial code and replace it using: (1) using CodeBERT~\cite{feng2020codebert} predictions with highest aggregated scores according to the context around all its appearance, a method inspired by~\cite{jha2022codeattack, li-etal-2020-bert-attack}, (2) using NatGen style renaming as "VAR\_0", and (3) random name generation with half alphabetic and half numeric characters. The first strategy tends to provide more natural variable names, yet names from the other two strategies are also plausible. %

\subsection{Natural Transformations on Code Format}
\label{subsec: code_format}
A natural way to perturb partial code is by code format transformations as they preserve the original semantic meaning. We implement following code format transformations in \texttt{ReCode}.

\textbf{Newline Insertion:} We consider three methods of new line insertions: (1) empty lines at randomly selected positions, (2) an empty line inserted between docstring and partial code, and (3) an empty line inserted after partial code.

\textbf{Tab-Indent:} We randomly replace any space indent with tab or replace tab with 4 spaces for indent-sensitive languages like Python.

\textbf{Line Split:} We select the longest line of code and split it into two lines in the middle.

\textbf{Docstrings to Comments:} We convert docstrings to comments (e.g., \texttt{""" docstring """} to \texttt{\# docstring} for Python).

\subsection{Evaluation Metrics}
\label{subsec: metric}

Many proposed transformations are randomized operations. Hence, we need to measure model robustness over multiple samples to reduce variance. 
Specifically, for each transformation and each prompt, we create $s$ randomly perturbed prompts. The model under evaluation generates outputs for each of them. We measure the worst-case performance across each group of $s$ perturbed prompts: the model is considered robust on a prompt if and only if it generates a correct solution for \textbf{all} $s$ perturbed prompts, where correctness is measured by executing associated unit tests.

Based on such worst-case measurements, we propose three new metrics for robustness evaluation.

\paragraph{Robust Pass$_{s}$@k (RP$_{s}$@k):} Pass@k is a widely used metric for measuring the performance of code generation tasks~\cite{chen2021evaluating}. We extend its definition to Robust Pass$_{s}$@k (RP$_s$@k) with $s$ random perturbations. 
For an original prompt $x$ and for each transformation, let the perturbed prompts be $x_1,\cdots, x_s$.
We sample $n$ generations by the model for each prompt, and in total there are $n \cdot s$ generations $f_i(x_j)$, where $1\leq i \leq n$ and $1\leq j \leq s$.
Instead of regular pass@k, we first consider the worst-case correctness across $f_i(x_1),...,f_i(x_s)$ for $1\leq i \leq n$: Let $c_{i,s}(x)=1$ if $f_i(x_1),...,f_i(x_s)$ are all correct and $c_{i,s}(x)=0$ otherwise.
Let $rc_s(x) = \sum_{i=1}^n c_{i, s}(x)$. 
Following definition of pass@k, we define the RP$_{s}$@k metric as \cref{eq: rp}.

\begin{equation}
\small
    \text{RP}{}_{s}@k:= \mathbb{E}_x \left[1- \frac{{n-rc_s(x) \choose k}}{{n \choose k}}\right] 
    \label{eq: rp}
\end{equation}

\paragraph{Robust Drop$_{s}$@k (RD$_{s}$@k):} 
RP$_s@k$ directly measure worst-case robustness in absolute values. It provides a worst-case estimation for models under certain perturbation. But in some applications, users may care more about \textbf{relative performance change} to compare worst-case performance and average-case performance.
We propose Robust Drop$_{s}$@k defined in \cref{eq: rd} as another important robustness metric to quantify relative changes.
\begin{equation}
\small
    \text{RD}{}_{s}@k:= \frac{\text{Pass@k} - \text{Robust Pass}{}_{s}@k}{\text{Pass@k}}
    \label{eq: rd}
\end{equation}

\paragraph{Robust Relative$_{s}$@k (RR$_{s}$@k):} Lastly, there are cases where models generate incorrect code on original prompts yet predict correctly on perturbed ones. This can (arguably) be considered as non-robust behavior that we should include when reporting model robustness. 
Let's first consider the case of greedy decoding with $n=k=1$.
Let $RC_{s}^{[-]}$ denote the number of correct-to-incorrect changes under the worst-case measurement as discussed.
Symmetrically, let $RC_{s}^{[+]}$ denote the number of incorrect-to-correct changes under best-case measurement: if the prediction with the original prompt is incorrect yet is correct for any of the $s$ perturbed prompts.
We define the Robust Relative$_{s}$@1 metric as the fraction of changes in both directions out of the size of the dataset ($N$):
\vspace{-5pt}
\begin{equation} 
\small
    \text{RR}{}_{s}@1:= \frac{RC_{s}^{[+]} + RC_{s}^{[-]}}{N}
    \label{eq: rr}
\end{equation}
This definition can be generalized to sampling.
Let $rc_{s}^{[-]}\left(x\right)$ and $rc_{s}^{[+]}\left(x\right)$ be similarly defined as $RC_{s}^{[-]}$ and $RC_{s}^{[+]}$ except that they are the number of changes within $n$ samples for a prompt $x$ instead of counting across the dataset.
We define
\begin{align}
\small
    \text{RR}{}_{s}@k:= & \mathbb{E}_x \left[2- \frac{{n-rc_s^{[-]}(x) \choose k}}{{n \choose k}}
                                     - \frac{{n-rc_s^{[+]}(x) \choose k}}{{n \choose k}}
                                     \right] 
    \label{eq: rr_k}
\end{align}
\cref{eq: rr_k} falls back to \cref{eq: rr} when $n=k=1$.

\vspace{-5pt}
\paragraph{Discussion.}
RP$_s@k$, RD$_s@k$ and RR$_s@k$ focus on different robustness requirements in practice. High RP$_s@k$ does not necessarily mean low RD$_s@k$ or RR$_s@k$, because the model may learn to utilize spurious correlation in the datasets to demonstrate better Pass$@k$ or RP$@k$, which is not robust.  
We advocate to report all of them to provide a comprehensive estimation of model robustness.

%% file: evaluation.tex
\section{Evaluation}
\label{sec: evaluation}

\paragraph{Evaluation setup.} In this work, we use execution-based code generation benchmarks HumanEval~\cite{chen2021evaluating} and MBPP~\cite{google_mbpp} to demonstrate our \recode robustness evaluation framework. We perform a comprehensive study of robustness evaluation on popular public models including CodeGen~\cite{CodeGen}, InCoder~\cite{incoder}, and GPT-J~\cite{gpt-j} to show the robustness comparisons across different model architectures and sizes. 
The perturbations and metrics implemented in \recode are general and applicable to any code generation datasets and models.

\begin{table*}[t]
\centering
\footnotesize
\setlength{\tabcolsep}{4.5pt}

\scalebox{0.8}{
\begin{tabular}{r|r|rr|rr|rr|rr|r} \toprule
\multirow{2}{*}{HumanEval}   & \multirow{2}{*}{Metric} & CodeGen & CodeGen  & CodeGen & CodeGen  & CodeGen  & CodeGen   & InCoder & InCoder & GPT-J \\
                             &                         & 2B mono & 2B multi & 6B mono & 6B multi & 16B mono & 16B multi & 1B      & 6B      & 6B    \\ \midrule
\multirow{4}{*}{Docstring} & Nominal$\uparrow$                 & 0.232   & 0.140    & 0.262   & 0.195    & \textbf{0.305}    & 0.195     & 0.104   & 0.152   & 0.122  \\
                           & RP$_5$@1$\uparrow$           & 0.122   & 0.049    & 0.104   & 0.073    & \textbf{0.128}    & 0.098     & 0.024   & 0.067   & 0.037  \\
                           & RD$_5$@1(\%)$\downarrow$               & \textbf{47.37}  & 65.28   & 60.47  & 62.50   & 58.00   & 50.00    & 76.47  & 56.00  & 70.00 \\
                           & RR$_5$@1(\%)$\downarrow$           & 20.73  & 14.63   & 27.44  & 18.90   & 35.37   & 18.90    & 14.63  & 15.85  & \textbf{10.98} \\ \midrule
\multirow{4}{*}{Function}  & Nominal$\uparrow$                 & 0.232   & 0.140    & 0.262   & 0.195    & \textbf{0.305}    & 0.195     & 0.104   & 0.152   & 0.122  \\
                           & RP$_5$@1$\uparrow$           & 0.140   & 0.061    & 0.146   & 0.116    & \textbf{0.213}    & 0.116     & 0.055   & 0.098   & 0.073  \\
                           & RD$_5$@1(\%)$\downarrow$               & 39.47  & 56.52   & 44.19  & 40.63   & \textbf{30.00}   & 40.63    & 47.06  & 36.00  & 40.00 \\
                           & RR$_5$@1(\%)$\downarrow$           & 14.02  & 10.37   & 18.90  & 12.20   & 19.51   & 9.146     & 8.537   & 9.756   & \textbf{6.098}  \\ \midrule
\multirow{4}{*}{Syntax}    & Nominal$\uparrow$                 & 0.402   & 0.293    & 0.518   & 0.366    & \textbf{0.549}    & 0.390     & 0.189   & 0.323   & 0.250  \\
                           & RP$_5$@1$\uparrow$           & 0.110   & 0.067    & 0.152   & 0.110    & \textbf{0.159}    & 0.091     & 0.043   & 0.079   & 0.079  \\
                           & RD$_5$@1(\%)$\downarrow$               & 72.73  & 77.08   & 70.59  & 70.00   & 71.11   & 76.56    & 77.42  & 75.47  & \textbf{68.29} \\
                           & RR$_5$@1(\%)$\downarrow$           & 41.46  & 32.93   & 44.51  & 36.59   & 46.95   & 39.02    & \textbf{21.34}  & 34.76  & 30.49 \\ \midrule
\multirow{4}{*}{Format}    & Nominal$\uparrow$                 & 0.402   & 0.293    & 0.518   & 0.366    & \textbf{0.549}    & 0.390     & 0.189   & 0.323   & 0.250  \\
                           & RP$_5$@1$\uparrow$           & 0.268   & 0.207    & 0.274   & 0.195    & \textbf{0.354}    & 0.232     & 0.091   & 0.171   & 0.104  \\
                           & RD$_5$@1(\%)$\downarrow$               & 33.33  & \textbf{29.17}   & 47.06  & 46.67   & 35.56   & 40.63    & 51.61  & 47.17  & 58.54 \\
                           & RR$_5$@1(\%)$\downarrow$           & 23.17  & 16.46   & 32.93  & 23.78   & 25.00   & 22.56    & \textbf{14.63}  & 23.78  & 21.95 \\ \bottomrule
\end{tabular}
}
\vspace{-10pt}
\caption{\recode benchmark robustness evaluation on popular code generation models for HumanEval. }
\label{tab: main_humaneval}
\end{table*}

\begin{table*}[t]
\centering
\footnotesize
\setlength{\tabcolsep}{4.5pt}

\scalebox{0.8}{
\begin{tabular}{r|r|rr|rr|rr|rr|r} \toprule
\multirow{2}{*}{MBPP}        & \multirow{2}{*}{Metric} & CodeGen & CodeGen  & CodeGen & CodeGen  & CodeGen  & CodeGen   & InCoder & InCoder & GPT-J \\
                             &                         & 2B mono & 2B multi & 6B mono & 6B multi & 16B mono & 16B multi & 1B      & 6B      & 6B    \\ \midrule

\multirow{4}{*}{Docstring} & Nominal$\uparrow$                 & 0.317   & 0.191    & 0.361   & 0.221    & \textbf{0.407}    & 0.241     & 0.128   & 0.199   & 0.133  \\
                           & RP$_5$@1$\uparrow$           & 0.137   & 0.050    & 0.147   & 0.042    & \textbf{0.163}    & 0.045     & 0.011   & 0.031   & 0.013  \\
                           & RD$_5$@1(\%)$\downarrow$               & \textbf{56.96}  & 73.66   & 59.38  & 80.93   & 59.85   & 81.28    & 91.20  & 84.54  & 90.00 \\
                           & RR$_5$@1(\%)$\downarrow$           & 36.86  & 34.39   & 41.89  & 36.76   & 46.72   & 44.66    & \textbf{25.57}  & 35.32  & 30.08 \\ \midrule
\multirow{4}{*}{Function}  & Nominal$\uparrow$                 & 0.317   & 0.191    & 0.361   & 0.221    & \textbf{0.407}    & 0.241     & 0.128   & 0.199   & 0.133  \\
                           & RP$_5$@1$\uparrow$           & 0.221   & 0.101    & 0.252   & 0.110    & \textbf{0.279}    & 0.139     & 0.047   & 0.087   & 0.043  \\
                           & RD$_5$@1(\%)$\downarrow$               & 30.42  & 47.31   & \textbf{30.40}  & 50.23   & 31.31   & 42.55    & 63.20  & 56.19  & 67.69 \\
                           & RR$_5$@1(\%)$\downarrow$           & 19.51  & 20.43   & 24.13  & 22.79   & 24.95   & 23.51    & \textbf{16.22}  & 20.02  & 17.56 \\ \midrule
\multirow{4}{*}{Syntax}    & Nominal$\uparrow$                 & 0.450   & 0.285    & 0.535   & 0.331    & \textbf{0.571}    & 0.379     & 0.219   & 0.292   & 0.176  \\
                           & RP$_5$@1$\uparrow$           & 0.027   & 0.008    & 0.027   & 0.008    & \textbf{0.038}    & 0.017     & 0.008   & 0.006   & 0.004  \\
                           & RD$_5$@1(\%)$\downarrow$               & \textbf{94.06}  & 97.12   & 95.01  & 97.52   & 93.34   & 95.39    & 96.24  & 97.89  & 97.66 \\
                           & RR$_5$@1(\%)$\downarrow$           & 59.03  & 45.07   & 64.17  & 47.74   & 67.04   & 54.21    & 35.42  & 45.79  & \textbf{30.60} \\ \midrule
\multirow{4}{*}{Format}    & Nominal$\uparrow$                 & 0.450   & 0.285    & 0.535   & 0.331    & \textbf{0.571}    & 0.379     & 0.219   & 0.292   & 0.176  \\
                           & RP$_5$@1$\uparrow$           & 0.333   & 0.146    & 0.289   & 0.166    & \textbf{0.403}    & 0.214     & 0.091   & 0.130   & 0.080  \\
                           & RD$_5$@1(\%)$\downarrow$               & \textbf{26.03}  & 48.92   & 46.07  & 49.69   & 29.32   & 43.63    & 58.22  & 55.28  & 54.39 \\
                           & RR$_5$@1(\%)$\downarrow$           & 19.82  & 25.15   & 31.11  & 27.00   & 25.26   & 26.59    & 19.61  & 28.54  & \textbf{18.28} \\ \bottomrule

\end{tabular}
}
\vspace{-10pt}
\caption{\recode benchmark robustness evaluation on popular code generation models for MBPP.}
\label{tab: main_mbpp}
\vspace{-10pt}
\end{table*}

\subsection{Code Generation Robustness Evaluation}
\label{subsec: main_eval}
\cref{tab: main_humaneval} and \cref{tab: main_mbpp} show the general perturbation performances on all the models in terms of the four general perturbation categories including transformations on docstrings, function names, code syntax, and code format. The nominal baselines are the pass@k on nonperturbed datasets for docstrings and function name perturbations. For perturbations on code syntax and format, the nominal baseline is the pass@k on nonperturbed customized datasets with partial code (see \cref{subsec: code_syntax}). We use greedy sampling for all the models to eliminate randomness effect and enable fair comparisons. We consider $s=5$, i.e., we generate five different datasets with different random seeds for each type of perturbations and evaluate worst-case robustness performance according to the robustness evaluation metric defined in \cref{subsec: metric}. To evaluate and compare model robustness in a unified fashion, we aggregate the worst 
performance across different perturbations under each category.  In specific, we say the model is robust on an input under docstring perturbations only when the model predicts correctly on all the $s$ perturbed datasets for each transformation listed in \cref{tab: docstring_example}. We present detailed numbers for each perturbation in \cref{appd: additiona_results}, \cref{appd: tab_doc_humaneval}-\ref{appd: tab_format_mbpp}.

(1) \textbf{Diverse pretraining corpus helps with both generalization and worst-case robustness.} Comparing all code generation models with the same size 6B, CodeGen models have much better nominal performance, and have better robustness on RP$_5@1$, a very strict worst-case robustness metric. That is possibly because CodeGen models are pretrained over a more diverse corpus than InCoder and GPT-J and thus have more capacity to deal with unseen instances and perturbations. Although CodeGen models have worse performance on RD$_5@1$ and RR$_5@1$, two robustness metrics relative to nominal performance, indicating that CodeGen models cannot generalize in a robust way (e.g., may learn to use spurious features in data). \footnote{Although these models may have some subtle architecture-wise differences in details, we follow the benchmarking and evaluation strategies in previous works to focus more on pretraining parts and model sizes (e.g., BigBench~\cite{srivastava2022beyond}, HELM~\cite{liang2022holistic}). We leave further ablation study for future work.}

(2) \textbf{Larger model size brings improvement in worst-case robustness, but may risk overfitting.} In general, we observe higher RP$_5@1$ for larger models  within the same model family (e.g., improved from $0.174$ to $0.217$ for CodeGen-mono 2B to 16B on average across all perturbations),
indicating larger model helps improve worst-case robustness. Similarly, we observe that larger models usually have larger RR$_5@1$ (e.g., increased from $27.90$\% to $35.91$\% for CodeGen-mono 2B to 16B on average), indicating that larger models may risk overfitting as the relative performance drops under perturbations are significant.

(3) \textbf{Code generation models are most sensitive to syntax perturbation.} Among all perturbation types and across MBPP and HumanEval, we observe that syntax perturbations often result in the most performance drops. That reveals a significant limitation of syntax understanding ability of the state-of-the-art code generation models.

(4) \textbf{Datasets having more variances in code style poses more challenges on model robustness.} In \cref{tab: across_dataset}, we can see that models show better robustness on HumanEval over MBPP on average. MBPP has more variances in code style (e.g., indent with 1 space), closer to natural code distribution hence more challenging for model robustness.

\begin{table}[ht]
\centering
\footnotesize
\scalebox{0.8}{
\begin{tabular}{l|r|r|r} \toprule
Category                   & {Metric} & {HumanEval} & MBPP           \\ \midrule
\multirow{3}{*}{Docstring} & RP$_5$@1$\uparrow$              & \textbf{0.078}     & 0.071          \\
                           & RD$_5$@1$(\%)\downarrow$                  & \textbf{60.67}     & 75.31          \\
                           & RR$_5$@1$(\%)\downarrow$              & \textbf{19.72}     & {36.92} \\ \midrule
\multirow{3}{*}{Function}  & RP$_5$@1$\uparrow$              & {0.113}     & \textbf{0.142} \\
                           & RD$_5$@1$(\%)\downarrow$                  & \textbf{41.61}     & 46.59          \\
                           & RR$_5$@1$(\%)\downarrow$              & \textbf{12.06}     & 21.01          \\ \midrule
\multirow{3}{*}{Syntax}    & RP$_5$@1$\uparrow$              & \textbf{0.100}     & 0.025          \\
                           & RD$_5$@1$(\%)\downarrow$                  & \textbf{72.58}     & 93.40          \\
                           & RR$_5$@1$(\%)\downarrow$              & \textbf{33.88}     & 47.86          \\ \midrule
\multirow{3}{*}{Format}    & RP$_5$@1$\uparrow$              & \textbf{0.211}     & 0.206          \\
                           & RD$_5$@1$(\%)\downarrow$                  & \textbf{43.30}     & 45.73          \\
                           & RR$_5$@1$(\%)\downarrow$              & \textbf{22.70}     & 24.60    \\ \bottomrule      
\end{tabular}
}
\vspace{-5pt}
\caption{Average robustness numbers across all models. MBPP is more challenging for robustness evaluation.}
\label{tab: across_dataset}
\vspace{-15pt}
\end{table}

\subsection{Ablation Study}
\label{subsec: ablation}

\begin{figure}
    \centering
    \includegraphics[width=0.48\linewidth]{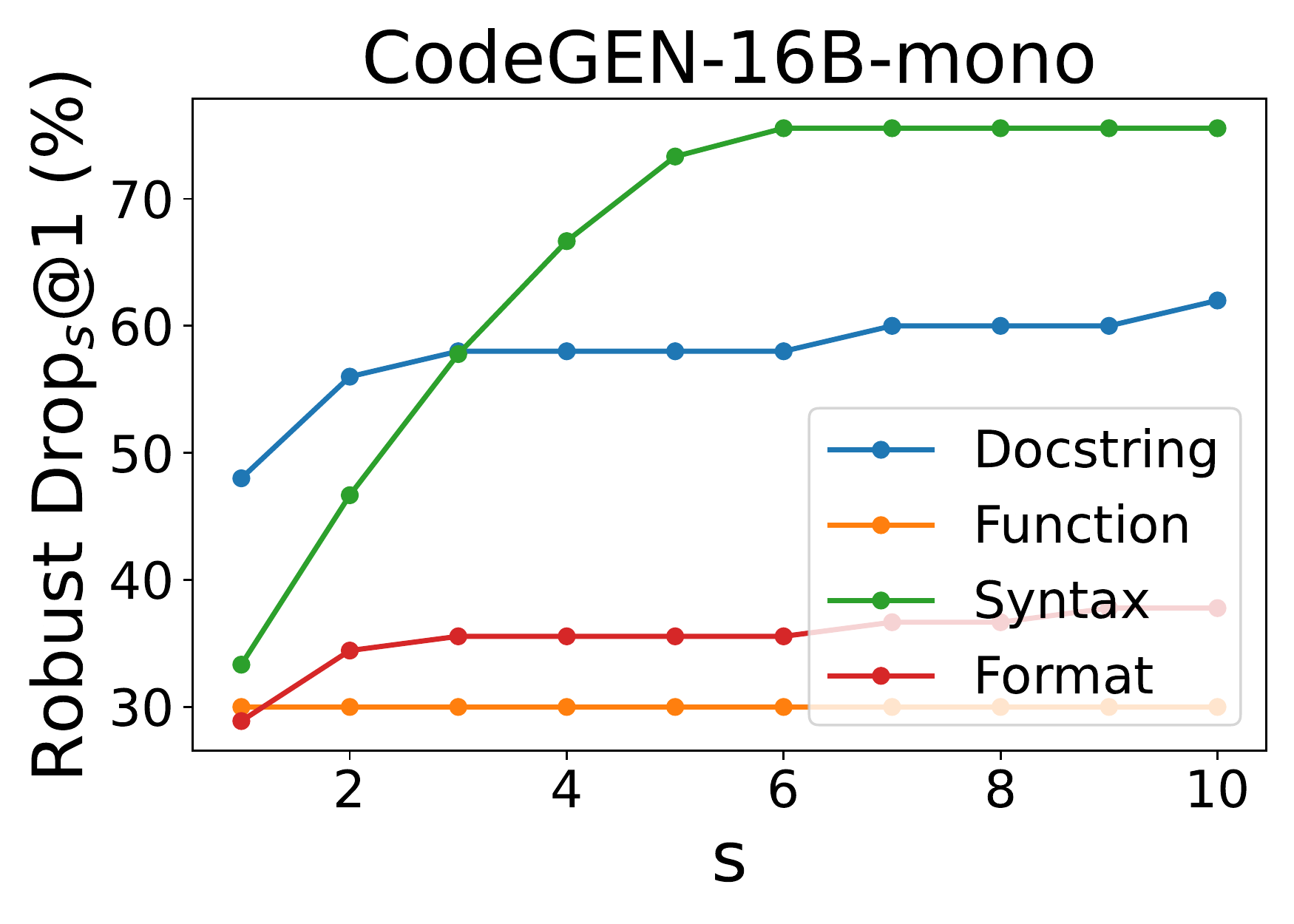}
    \includegraphics[width=0.48\linewidth]{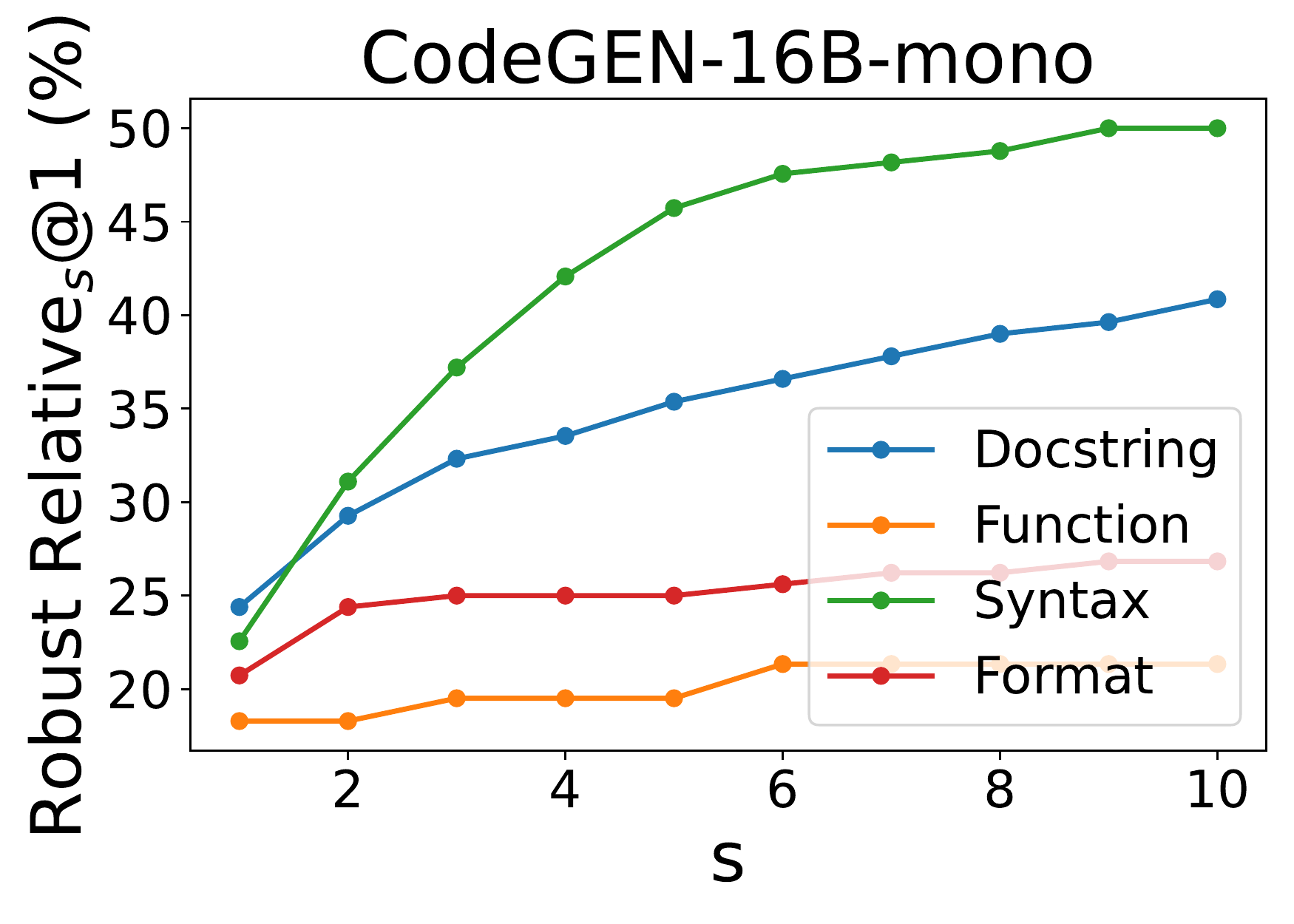}
    \vspace{-10pt}
    \caption{Robust Drop$_s$@1 and Robust Relative$_s$@1 under different $s$. Larger $s$ indicates stronger perturbations evaluated and larger performance drops.}
    \label{fig: s}
    \vspace{-10pt}
\end{figure}

\paragraph{Robustness with $s$ perturbed datasets.} As described in \cref{subsec: metric}, our robustness metrics consider worst-case performance across $s$ perturbed datasets for each perturbations. Larger $s$ leads to stronger perturbations evaluated, larger performance drops, and more extensive coverage to practical failures. The performance drops will start converging when large enough $s$ evaluated. We can clearly see such trends in \cref{fig: s} where we evaluate CodeGen-16B-mono RD$_s$@1 and RR$_s$@1 under greedy sampling with $s=1,...,10$. Perturbation categories like docstring and syntax that involve larger searching space and more randomness tend to benefit more with larger $s$ (see \cref{appd: transformation} for details). As a trade-off, evaluation cost linearly increase with $s$. Thus, we recommend $s=5$ as a good balance between cost and evaluation strength.

\begin{figure}
    \centering
    \includegraphics[width=0.48\linewidth]{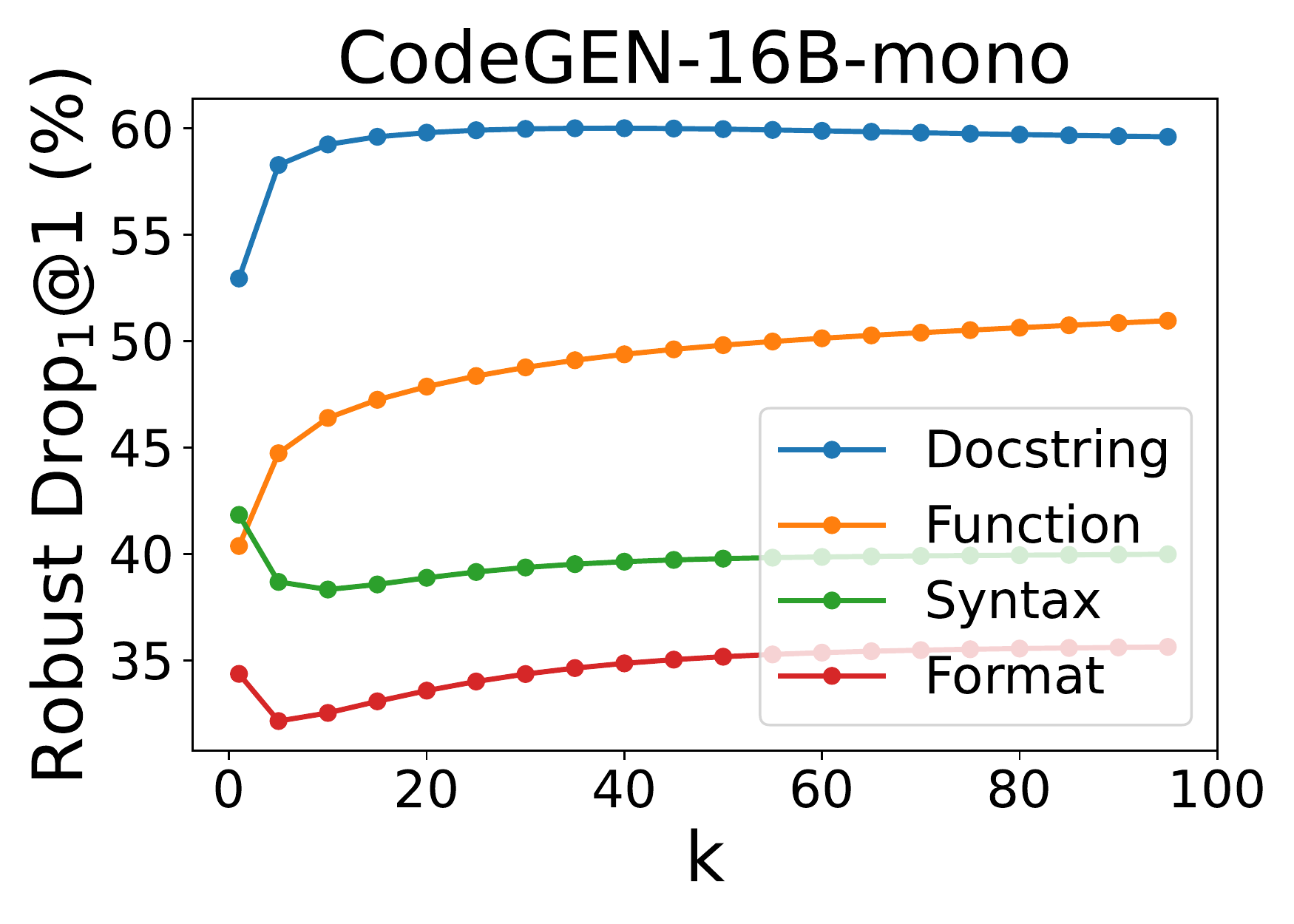}
    \includegraphics[width=0.48\linewidth]{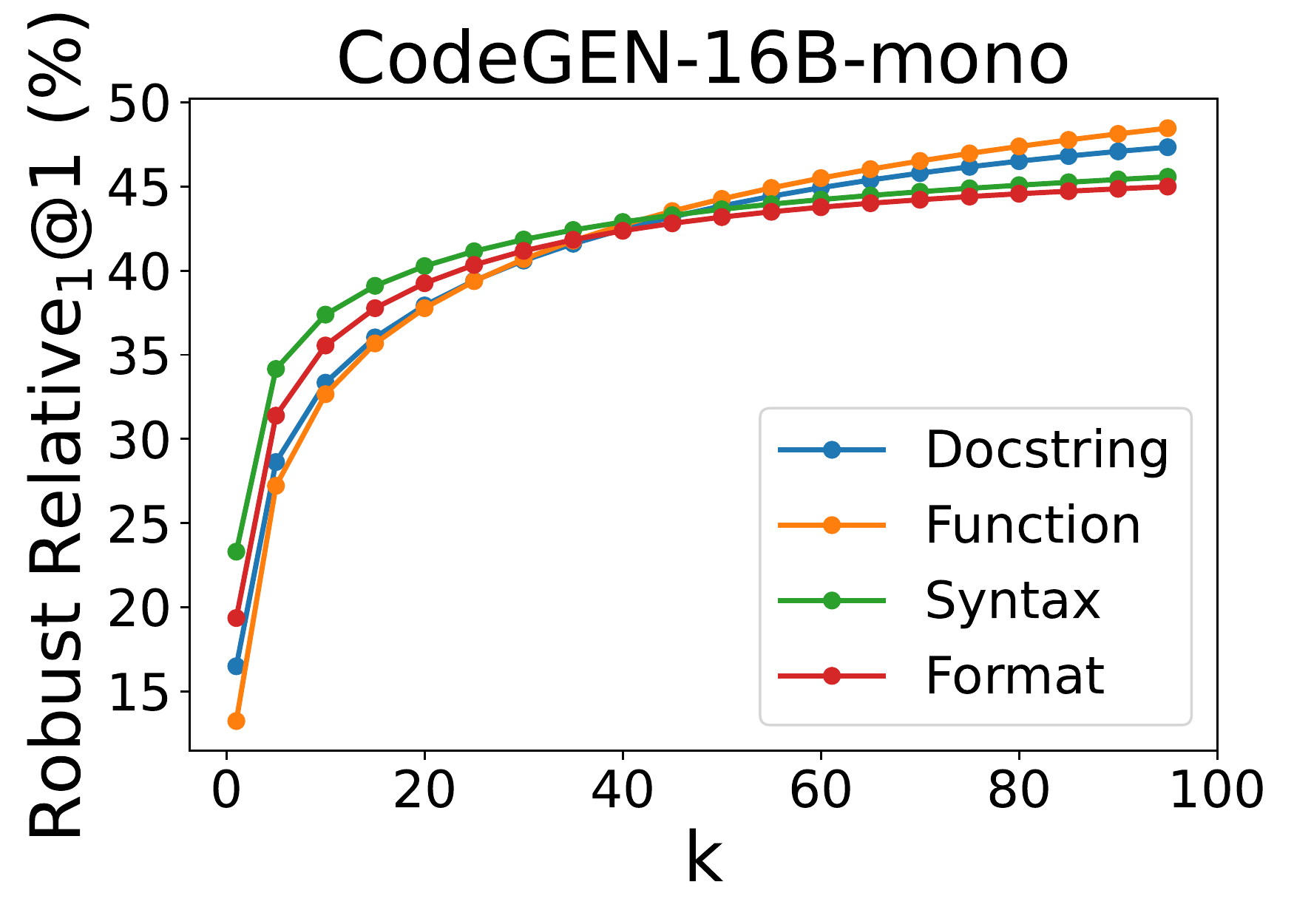}
    \vspace{-10pt}
    \caption{Robust Drop$_1$@k and Robust Relative$_1$@k under different $k$ using sampling $n=100$. Robust Drop remains stable while Robust Relative increases with k.}
    \label{fig: k}
    \vspace{-10pt}
\end{figure}

\vspace{-5pt}
\paragraph{Stable RD@k and increasing RR@k under different $k$.} Pass@k allows the model to have k trials and model performance is often reported with different k. With the sampling setting of $n=100$, we plot the RD$_1$@k and RR$_1$@k in \cref{fig: k}. Interestingly, we observe that RD@k stays stable across different k while RR@k increases with k. This is because larger k leads to higher nominal pass@k and RP@k but their relative ratio stays similar leading to stable RD. On the other hand, larger k involves more samples potentially changing results on perturbed datasets causing larger RR. Similar trends on CodeGen-2B and 6B in \cref{appd: k} further confirm the observations.

\subsection{Perturbation Sample Quality}
\label{subsec: quality}

\vspace{10pt} 
\begin{table}[t]
\centering
\footnotesize
\scalebox{0.80}{
\begin{tabular}{l|r|r} \toprule
                      & \multicolumn{1}{c}{HumanEval} & \multicolumn{1}{c}{MBPP} \\ \midrule
Naturalness (Nominal) $\uparrow$   & 0.92                          & 0.92                     \\ 
Naturalness (Perturbed) $\uparrow$ & 0.75                          & 0.80                     \\
Semantics Similarity $\uparrow$ & 0.92                          & 0.92                     \\
\bottomrule
\end{tabular}
}
\vspace{-5pt}
\caption{Human evaluation for practical naturalness and semantic similarity by 5 annotators. Either 0, 0.5, or 1 is assigned to each data point indicating quality level.}
\label{tab: human_eval}
\vspace{-5pt}
\end{table}

\begin{table}[t]
\centering
\footnotesize
\scalebox{0.80}{
\setlength{\tabcolsep}{3pt}
\begin{tabular}{l|rr|rr} \toprule
                  & \multicolumn{2}{c}{HumanEval}                           & \multicolumn{2}{|c}{MBPP}                                \\
                  & \multicolumn{1}{l}{Syntax} & \multicolumn{1}{l}{Format} & \multicolumn{1}{|l}{Syntax} & \multicolumn{1}{l}{Format} \\ \midrule
CodeBLEU (syntax) $\uparrow$   & 0.95                      & 0.98                      & 0.93                      & 0.96                      \\
CodeBLEU (dataflow) $\uparrow$ & 0.94                      & 1.00                      & 0.92                      & 1.00              \\ \bottomrule        
\end{tabular}
}
\vspace{-5pt}
\caption{Average CodeBLEU syntax and format scores between non-perturbed codes and perturbed ones with our syntax and format transformations.\vspace{-10pt}}
\label{tab: codegleu}
\end{table}

\vspace{-5pt}
\paragraph{Human evaluation.}
\label{subsec: human_eval}

To verify the naturalness of the perturbations in \recode, we randomly sample and shuffle $100$ and $50$ perturbed and non-perturbed MBPP and HumanEval data points and create a shuffle mix of $300$ samples. Each sample is shown to 5 human annotators who are familiar with Python and who are asked to rate naturalness out of 0: not natural, 0.5: possible to appear in practice but rare, and 1: natural. The scores for naturalness drop 14\% on average for our perturbed data where drops mainly come from typos by Butterfingers, CharCaseChanges, SwapCharacter, etc.

In addition, we randomly sample $100$ and $50$ pairs perturbed and non-perturbed MBPP and HumanEval data points. Each pair is shown to 5 human annotators who are asked to rate semantics out of 0: totally changed, 0.5: slightly changed, and 1: exactly preserved. Results are in \cref{tab: human_eval} {(\cref{appd: human_evaluation} for more details).} Notably, the majority vote (at least three out of five) is 1 for 90\% of data points. {We further provide automatic evaluation below to support the quality of our perturbed datasets but human evaluation is in general more reliable.}

\vspace{-5pt}
\paragraph{Docstring/function names similarity.} 
We measure the sentence cosine similarity between perturbed and non-perturbed docstrings and function names. We obtain the embeddings by sentence transformers using model \texttt{all-mpnet-base-v2}\footnote{Model embedding quality in \url{https://www.sbert.net}}~\cite{song2020mpnet}. Note that we split each function name into words to get sentence embeddings. On average, we have 0.93 and 0.81 for docstring and function name perturbations, showing that they well preserve the semantics. Scores for some function name perturbations are sensitive to typos due to the lack of sentence context (e.g., 0.21 for \texttt{interperse} and \texttt{intErpErse}). \cref{appd: sentrans} summarizes detailed numbers for each perturbation.

\vspace{-5pt}
\paragraph{Code syntax/format similarity.} In \cref{tab: codegleu}, we also measure the code similarity using CodeBLEU scores~\cite{lu2021codexglue} for perturbed and non-perturbed data involving code syntax/format transformations. Here we consider the CodeBLEU score with syntax and dataflow separately as the evaluation metrics. On average, we have score 0.96 and 0.97 for CodeBLEU syntax and dataflow, showing good quality of perturbed datasets. Note that a few perturbations should expect low CodeBLEU scores: \texttt{doc2comments} transforms docstrings into comments causing changes of syntax; \texttt{Deadcode insertion} and \texttt{for-while switch} involve new if-conditions, loops, and new variables causing changes of code syntax and dataflow. Please refer to \cref{appd: codebleu} for details.

%% file: conclusion.tex
\section{Conclusion}
\label{sec: conclusion}

In this paper, we propose \texttt{ReCode}, a comprehensive robustness evaluation benchmark for code generation models. We collect and customize over 30 natural transformations under categories of docstrings, function names, code syntax, and code format perturbations. These transformations are carefully selected and designed to be natural in practice and preserve the semantic meaning after perturbations. We further propose general worst-case robustness metrics to give a unified overview of the model robustness performance. We empirically demonstrate our ReCode benchmark on popular models including CodeGen, InCoder, and GPT-J using HumanEval and MBPP datasets and function completion tasks derived from them. With human evaluation, over 90\% of our perturbed data are confirmed to preserve the original semantic meaning; sentence similarity and CodeBLEU scores additionally support the quality of perturbations in \texttt{ReCode}.

%% file: appendix.tex
\section{Transformation Details and Qualitative Examples}
\label{appd: transformation}

In this section, we give detailed descriptions and settings for each type of perturbations that are included in our ReCode benchmark with qualitative examples for illustrations.

\subsection{Natural Transformations on Docstrings}

For natural transformations on docstrings, we aim to perturb the docstrings to their variances that preserve the semantics and also appear natural in practice. Specifically, we will first extract and perturb the docstrings with the following natural transformations in each prompt, and then attach their perturbed versions to the prompt. To preserve semantics for the code generation task prompts, we extract a blacklist of program keywords using tree-sitter as discussed in~\cref{subsec: text} that are excluded from perturbations. We extend most transformations from NL-Augmenter~\cite{dhole2021nl}, a standard library designed for data augmentation and robustness evaluation on text. We list some qualitative examples in \cref{tab: docstring_example}.

\paragraph{BackTranslation.} BackTranslation paraphrases the docstrings by translating them to another language (in this case, German) and then back to English. It is a common method for data augmentation in generating sentence variances with the same semantics~\cite{li2019improving,sugiyama2019data}. Overall, it can reliably generate high quality perturbed docstrings. We use the default implementation in NL-Augmenter~\cite{dhole2021nl}. BackTranslation contains no randomness in transformations.

\paragraph{ButterFingers.} ButterFingers transformation randomly selects characters of the docstrings and perturbs each of them to a random subset of similar characters, it is from~\cite{dhole2021nl} and is also used in~\cite{mille2021automatic}. Since this transformation tends to introduce character-level typos, we set randomness for perturbing each character to be low as 0.05 for naturalness consideration.

\paragraph{ChangeCharCase.}  ChangeCharCase transformation randomly changes the selected characters to upper case in the docstrings. We use the default probability 0.35 where majority annotators vote 0.5 for naturalness in the setting of \cref{subsec: human_eval}.

\paragraph{EnglishInflectionalVariation.}
This transformation randomly selects words in the docstring and change them to a random inflection variance. This can be from plural to singular (or vice versa) for nouns and tense changes for verbs. To maintain naturalness, the perturbation is constrained to be the same Part of Speech (POS) tag in the Penn Treebank~\cite{marcus-etal-1993-building}. 

\paragraph{SwapCharacters.}
This transformation randomly selects pairs of adjacent characters in the docstring and swap them. This represents a common type of typos by humans. To ensure naturalness, we set the probability as 0.05 for making the swap.

\paragraph{SynonymInsertion.}
This transformation randomly select words in the docstrings and inserts their synonyms in WordNet~\cite{miller1995wordnet}. Punctuations and stopwords are excluded. We set the probability to be 0.35 considering low success rate after keywords filtering.

\paragraph{SynonymSubstitution.}
This transformation randomly selects words in the docstring and replaces each one with a synonym from WordNet~\cite{miller1995wordnet}. Similar to SynonymInsertion, we set the probability as 0.35 to balance naturalness and perturbation success rates.

\paragraph{TenseTransformationPast.}
This is a deterministic transformation that converts sentences in the docstring to past tense.

\paragraph{TenseTransformationFuture.}
This is a deterministic transformation that converts sentences in the docstring to future tense.

\paragraph{Whitespace.}
This transformation inserts or deletes a single white space at randomly selected locations in the docstring.
This represents a common type of typos by humans. Folowing NL-Augmenter, we use probability 0.1 for adding whitespaces and 0.05 for removing whitespaces.

\subsection{Natural Transformations on Function Names}

These transformations modify the name of the target function to generate.
Any references to the function name in the prompt, e.g., in docstring, are also modified to maintain consistency. Qualitative examples can be found in \cref{tab: func_examples}.

\paragraph{CamelCase.}
A function name is often composed of multiple words.
If the original function name concatenates the words in camel-case style, this transformation changes it to snake-case, and vice versa. This transformation is deterministic.

\paragraph{ButterFingers.} ButterFingers transformation randomly selects characters of the docstrings and perturbs each of them to a random subset of similar characters, it is from~\cite{dhole2021nl} and is also used in~\cite{mille2021automatic}. Since this transformation tends to introduce character-level typos, we set randomness for perturbing each character to be low as 0.05 for naturalness consideration.

\paragraph{SwapCharacters.}
This transformation randomly selects pairs of adjacent characters in the function name and swap each pair.
This represents a common type of typos by humans.
To control naturalness, we set the probability to be 0.05, same setting as the docstring perturbations.

\paragraph{ChangeCharCase.}  ChangeCharCase transformation randomly changes the selected characters to upper case in the docstrings. We use the default probability 0.35 where majority annotators vote 0.5 for naturalness in the setting of \cref{subsec: human_eval}.

\paragraph{InflectionalVariation.}
This transformation randomly selects words in the function name and applies a random inflection on them.
This can be from plural to singular (or vice versa) for nouns and tense change for verbs. To control naturalness, the perturbation is constrained to be the same Part of Speech (POS) tag in the Penn Treebank~\cite{marcus-etal-1993-building}. 

\paragraph{SynonymSubstitution.}
This transformation randomly selects words in the docstring and replaces each one with a synonym from WordNet~\cite{miller1995wordnet}. Similar to SynonymInsertion, we set the probability as 0.35 to balance naturalness and perturbation success rates.

\subsection{Natural Transformations on Code Syntax}

\begin{figure}[!hbt]
    \centering
    \includegraphics[width=0.6\linewidth]{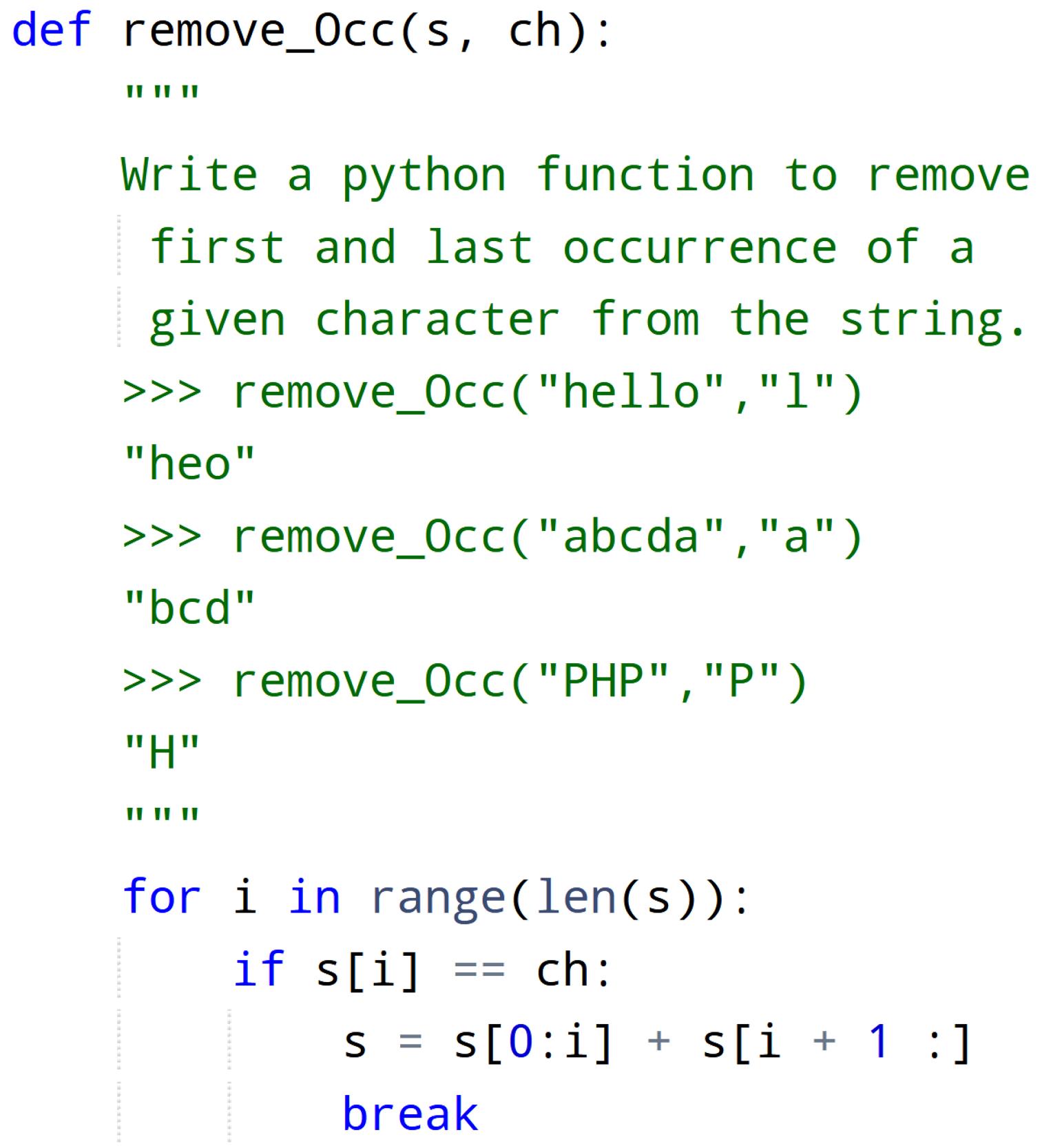}
    \caption{An baseline example of the prompt with partial code derived from original MBPP prompt for robustness evaluation on code.}
    \label{fig: appd_partial}
\end{figure}

These transformations modify the code content in the prompt. We derived function completion tasks with half the code from the canonical solutions such that the following code transformations and robustness evaluation can be performed. To guarantee fair comparisons to the nominal baseline, we make sure that we have the same block of code before and after code perturbations. In the following part we show qualitative examples on the same MBPP sample baseline (~\cref{fig: appd_partial}).

\paragraph{DeadCodeInserter.}
This transformation inserts a block of useless code at a random location.
The added block can be a loop of zero iteration or an if condition that is always false.
The code content inside the dummy loop or if condition is randomly selected from the adjacent code statements with limited tree-sitter node sizes.

\begin{figure}[!hbt]
    \centering
    \includegraphics[width=0.6\linewidth]{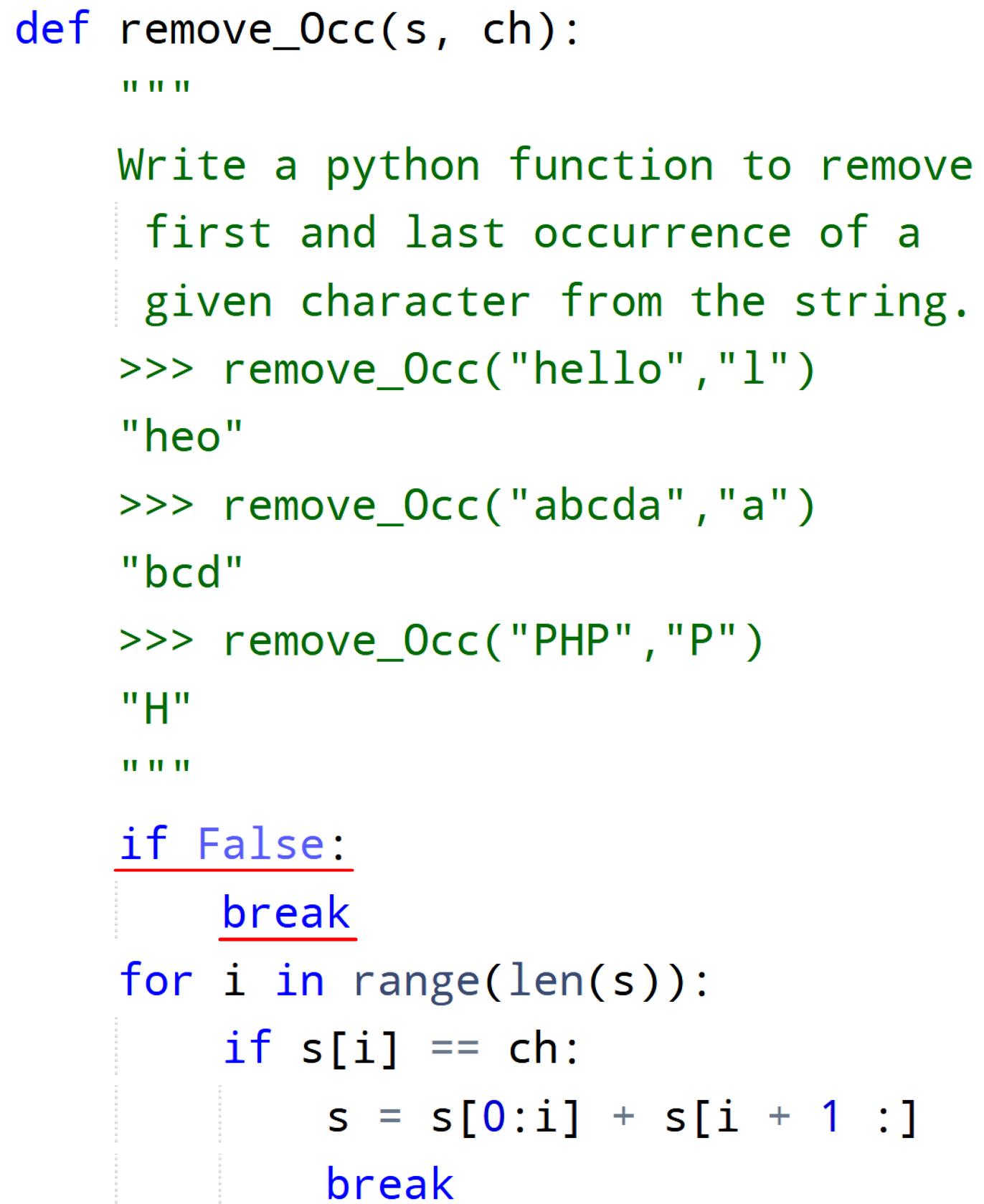}
    \caption{An example of the DeadCodeInserter perturbation.}
    \label{fig: deadcodeinserter}
\end{figure}

\paragraph{For-While Switch.}
This transformation randomly selects a for-loop or while-loop in the prompt and transforms it to its equivalent counterpart.
\begin{figure}[!hbt]
    \centering
    \includegraphics[width=0.6\linewidth]{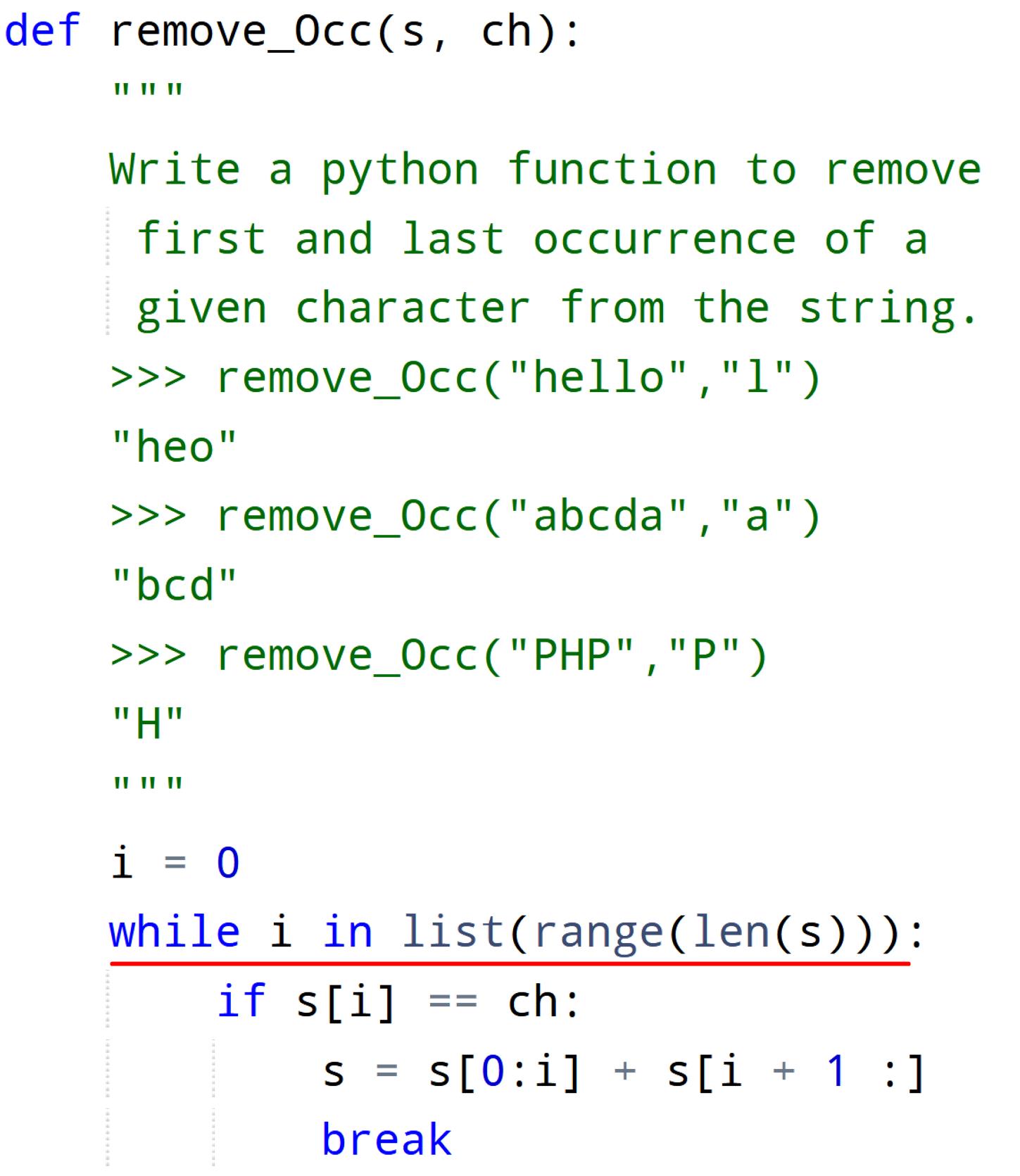}
    \caption{An example of the For-While Switch perturbation.}
    \label{fig: forwhile}
\end{figure}

\paragraph{OperandSwap.}
This transformation randomly selects a binary logical operation, swaps the two operands, and modifies the operator if necessary to maintain semantic equivalence.
\begin{figure}[!hbt]
    \centering
    \includegraphics[width=0.6\linewidth]{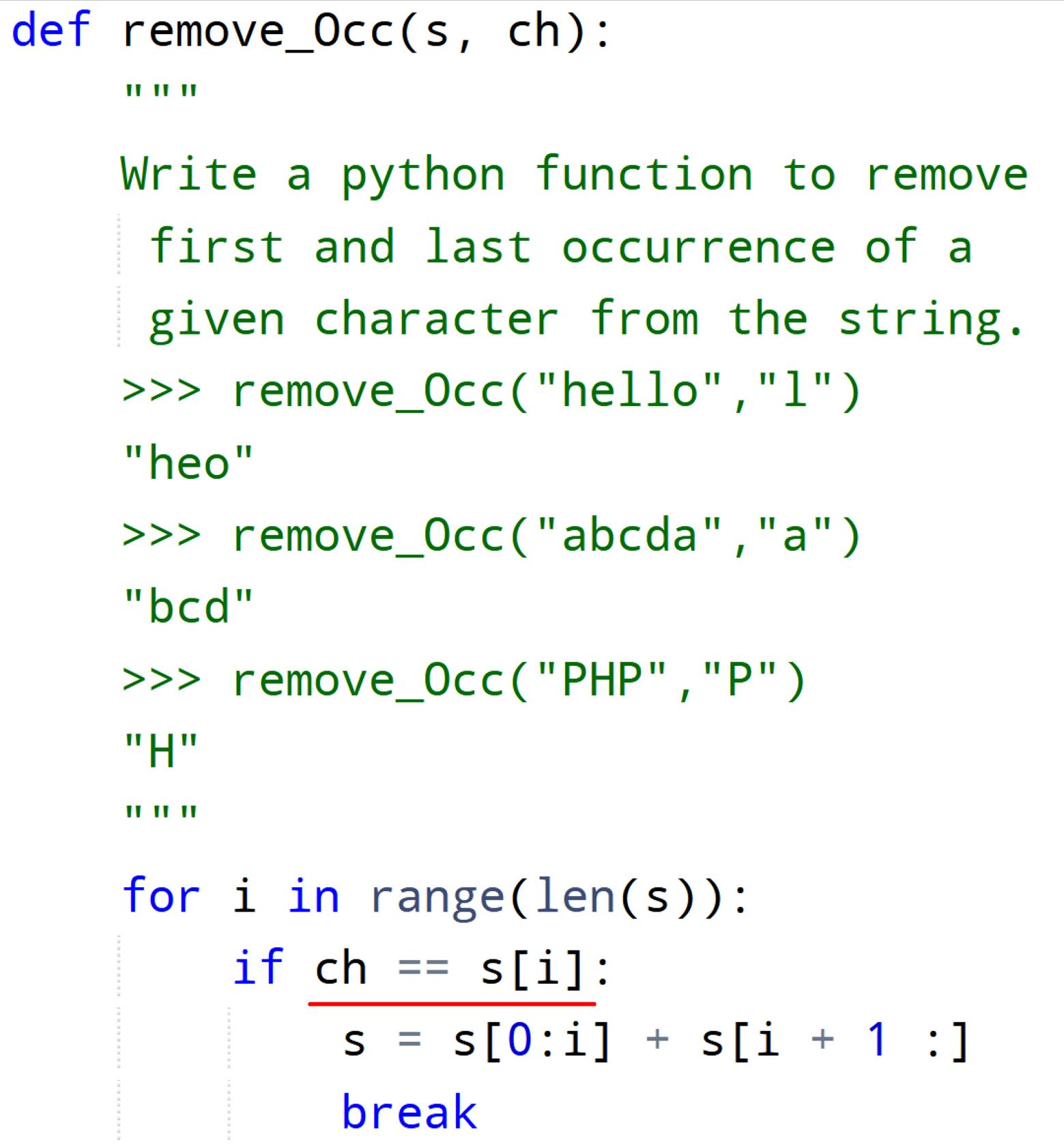}
    \caption{An example of the OperandSwap perturbation.}
    \label{fig: oprandswap}
\end{figure}

\paragraph{VarRenamerCB.}
This transformation selects the most frequently referenced variable name in the partial code and replaces it throughout the prompt with a new name obtained by CodeBERT~\cite{feng2020codebert}.
Specifically, we replace all occurrence of the variable name with a mask token, and then run CodeBERT inference to obtain candidate names at each location, where each candidate name comes with a probability score.
We pick the candidate name with the highest aggregated score across locations.
This transformation is inspired by~\cite{jha2022codeattack, li-etal-2020-bert-attack}.
\begin{figure}[!hbt]
    \centering
    \includegraphics[width=0.75\linewidth]{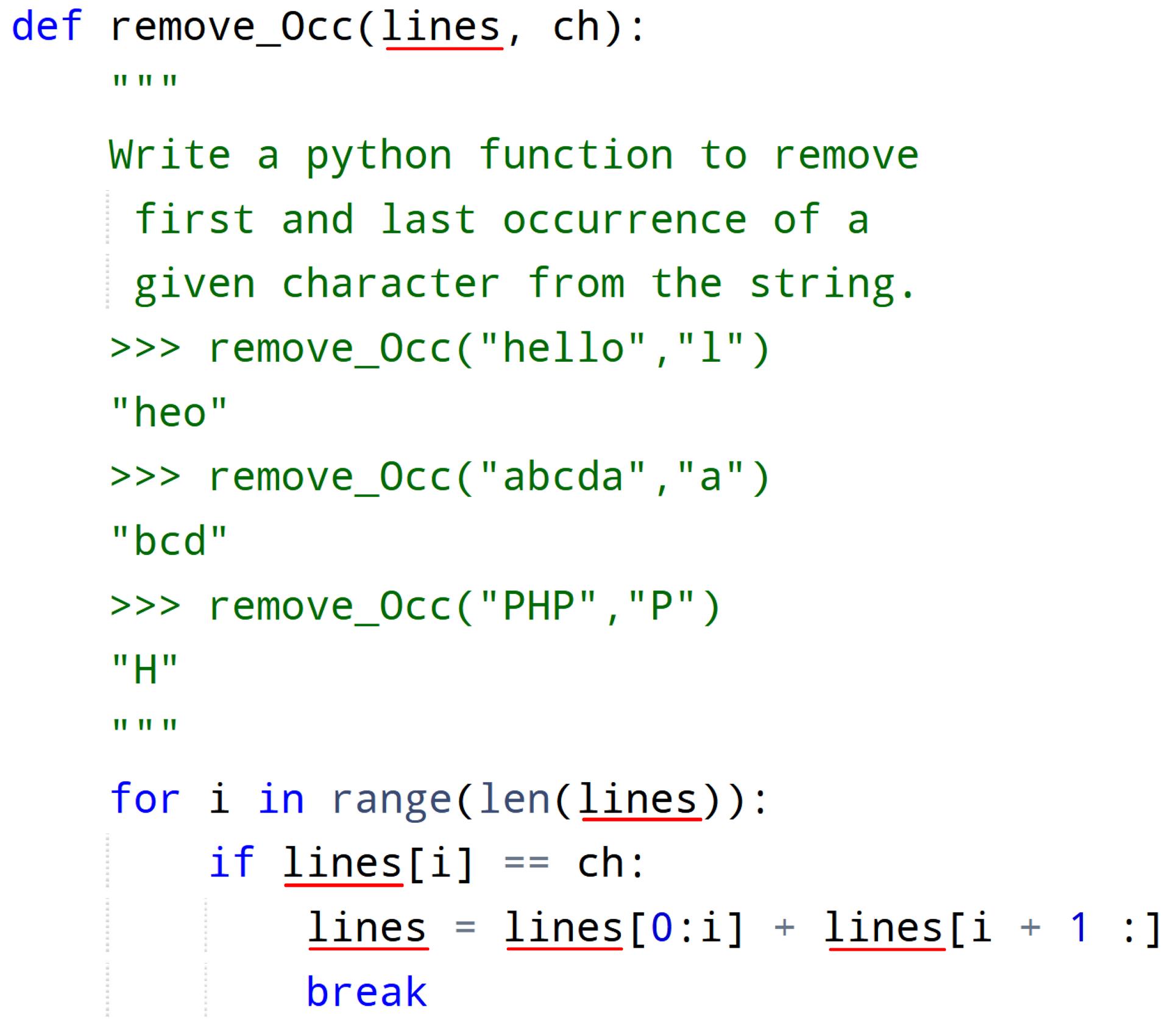}
    \caption{An example of the VarRenamerCB perturbation.}
    \label{fig: renamecb}
\end{figure}

\paragraph{VarRenamerNaive.}
This transformation selects the most frequently referenced variable name in the partial code and replaces it with "VAR\_0".
This is the original implementation in the NatGen package.
This transformation is deterministic.
\begin{figure}[!hbt]
    \centering
    \includegraphics[width=0.6\linewidth]{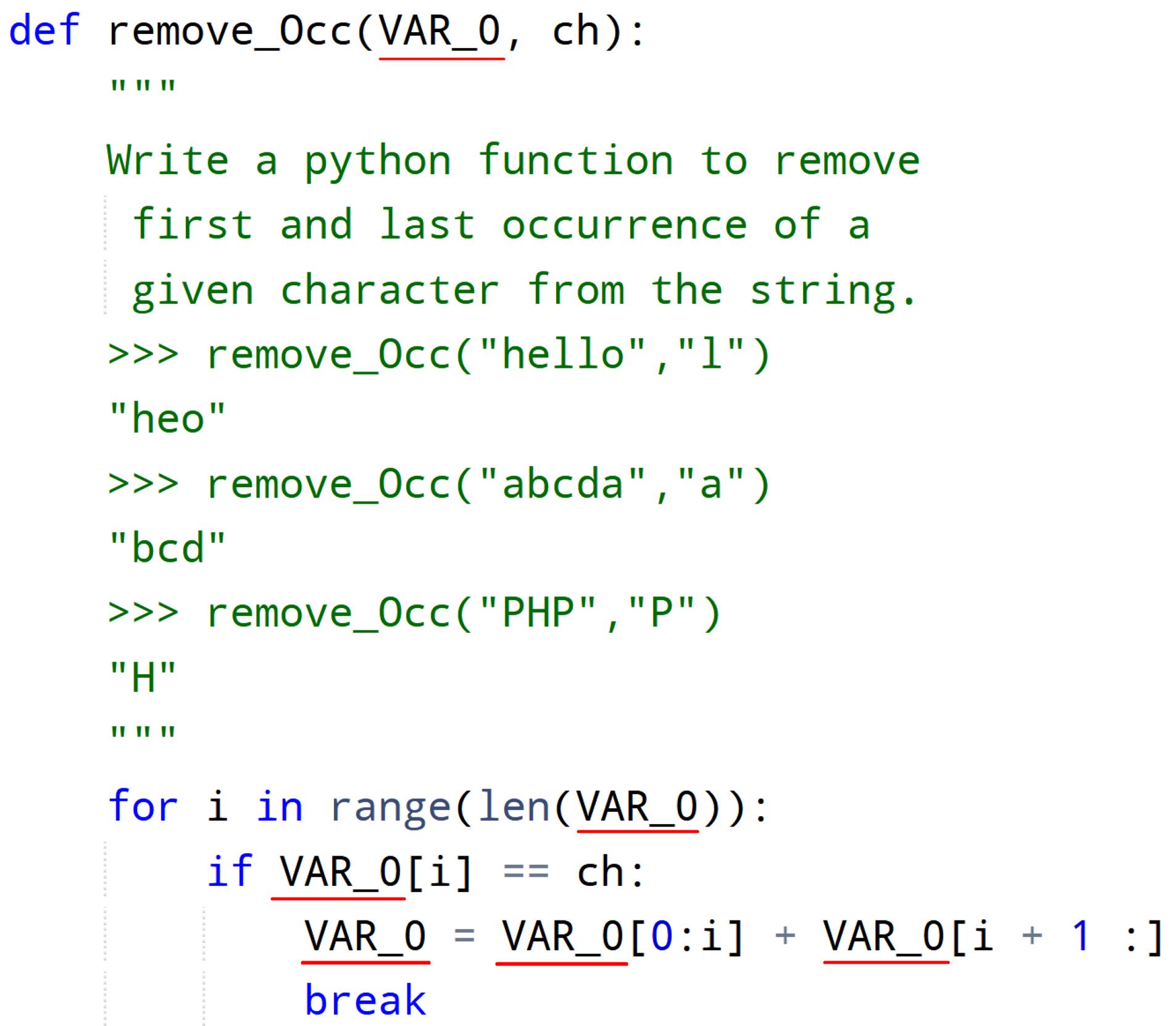}
    \caption{An example of the VarRenamerNaive perturbation.}
    \label{fig: renamenaive}
\end{figure}

\paragraph{VarRenamerRN.}
This transformation selects the most frequently referenced variable name in the partial code and replaces it with a random string with half alphabetic and half numeric characters.
\begin{figure}[!hbt]
    \centering
    \includegraphics[width=0.6\linewidth]{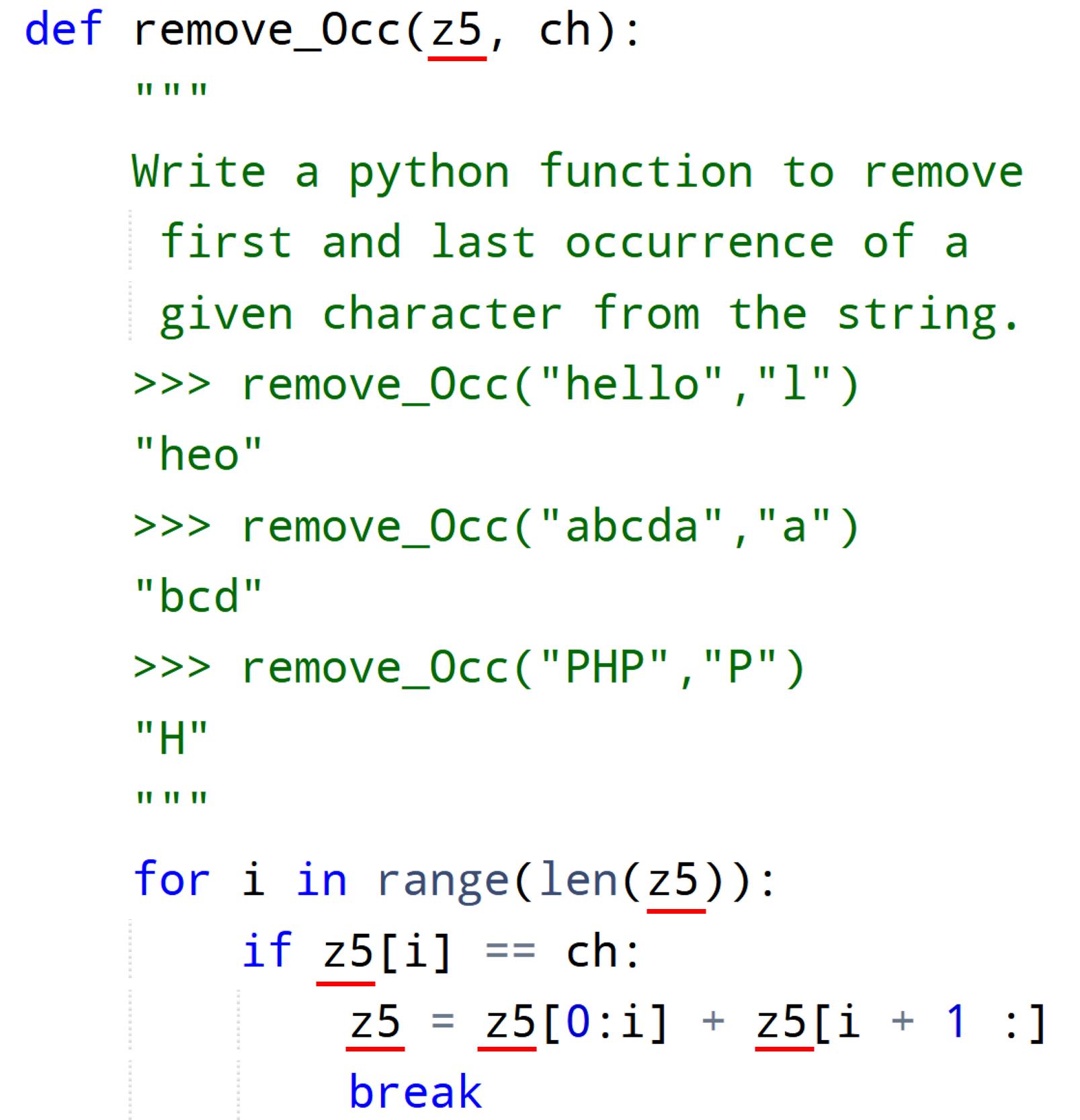}
    \caption{An example of the VarRenamerRN perturbation.}
    \label{fig: renamern}
\end{figure}

\subsection{Natural Transformations on Code Format}

\paragraph{Tab-Indent.}
This transformation replaces any space indents with tabs or replaces tabs with 4 spaces for indent-sensitive languages like Python. This transformation is deterministic.
\begin{figure}[!hbt]
    \centering
    \includegraphics[width=0.6\linewidth]{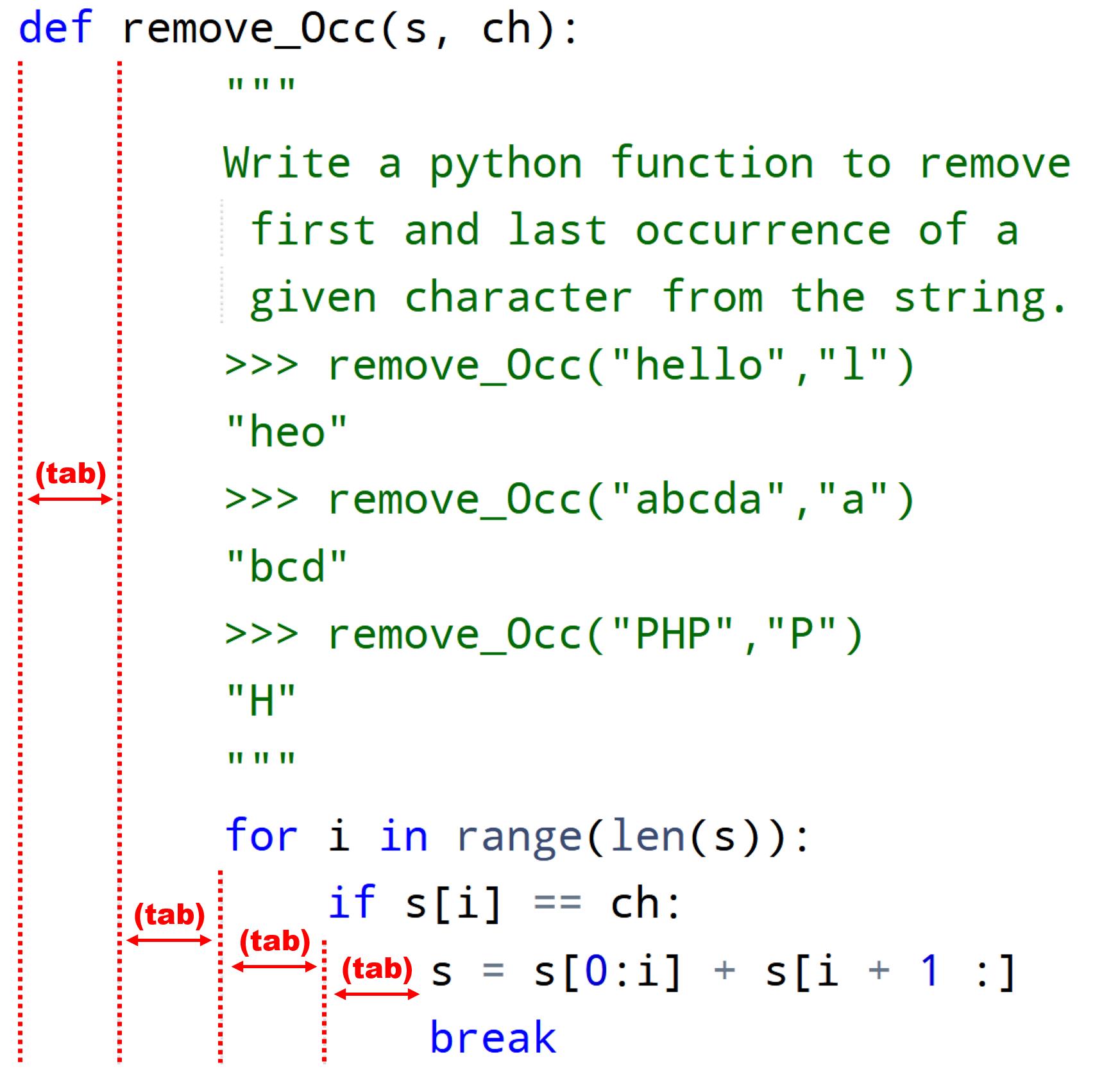}
    \caption{An example of the Tab-Indent perturbation.}
    \label{fig: tab}
\end{figure}

\paragraph{Line Split.}
This transformation splits the longest line in the partial code into two lines.
This transformation is deterministic.
\begin{figure}[!hbt]
    \centering
    \includegraphics[width=0.6\linewidth]{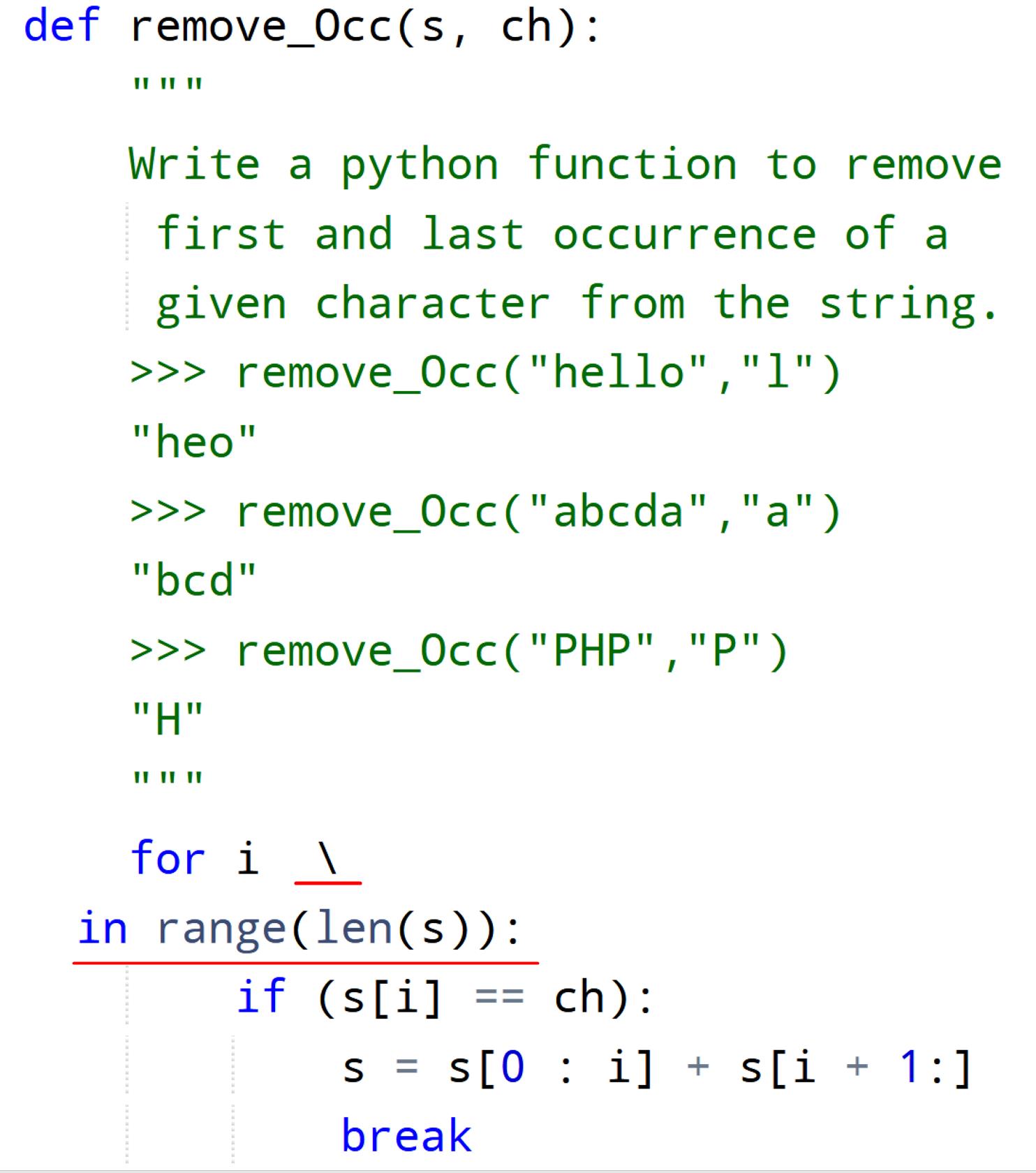}
    \caption{An example of the Line Split perturbation.}
    \label{fig: linesplit}
\end{figure}

\paragraph{Doc2Comments.}
This transformation changes the style of the documentation in the prompt.
For Python, it converts docstring (e.g., \texttt{""" docstring """}) to commented lines (e.g., \texttt{\# docstring}) and vice versa.
For Java, it converts comments in the format of \texttt{/* docstring */} to \texttt{// docstring} and vice versa.
This transformation is deterministic.
\begin{figure}[!hbt]
    \centering
    \includegraphics[width=0.6\linewidth]{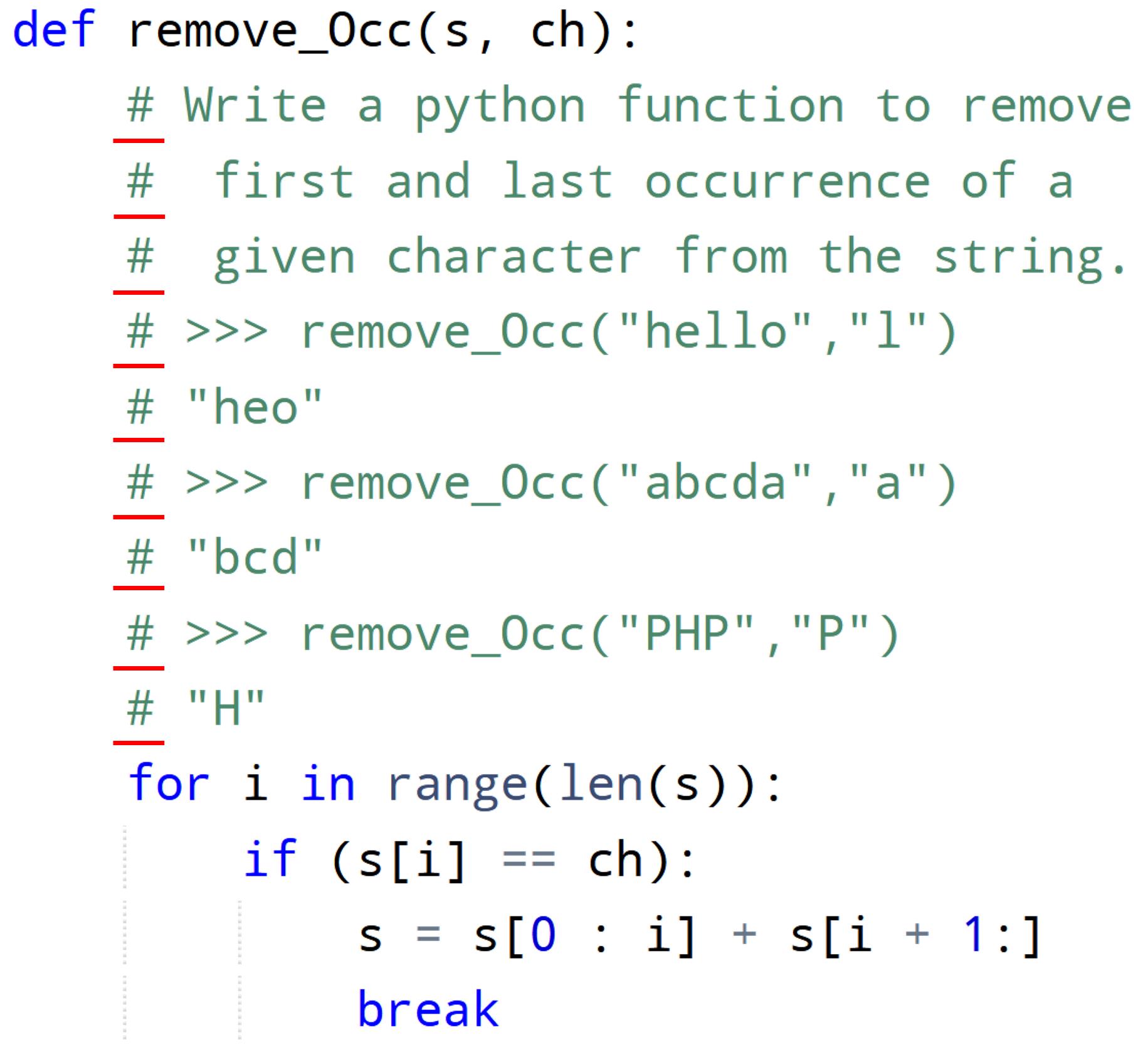}
    \caption{An example of the Doc2Comments perturbation.}
    \label{fig: doc2com}
\end{figure}

\paragraph{NewlineRandom.}
This transformation inserts empty lines at randomly selected positions.
\begin{figure}[!hbt]
    \centering
    \includegraphics[width=0.6\linewidth]{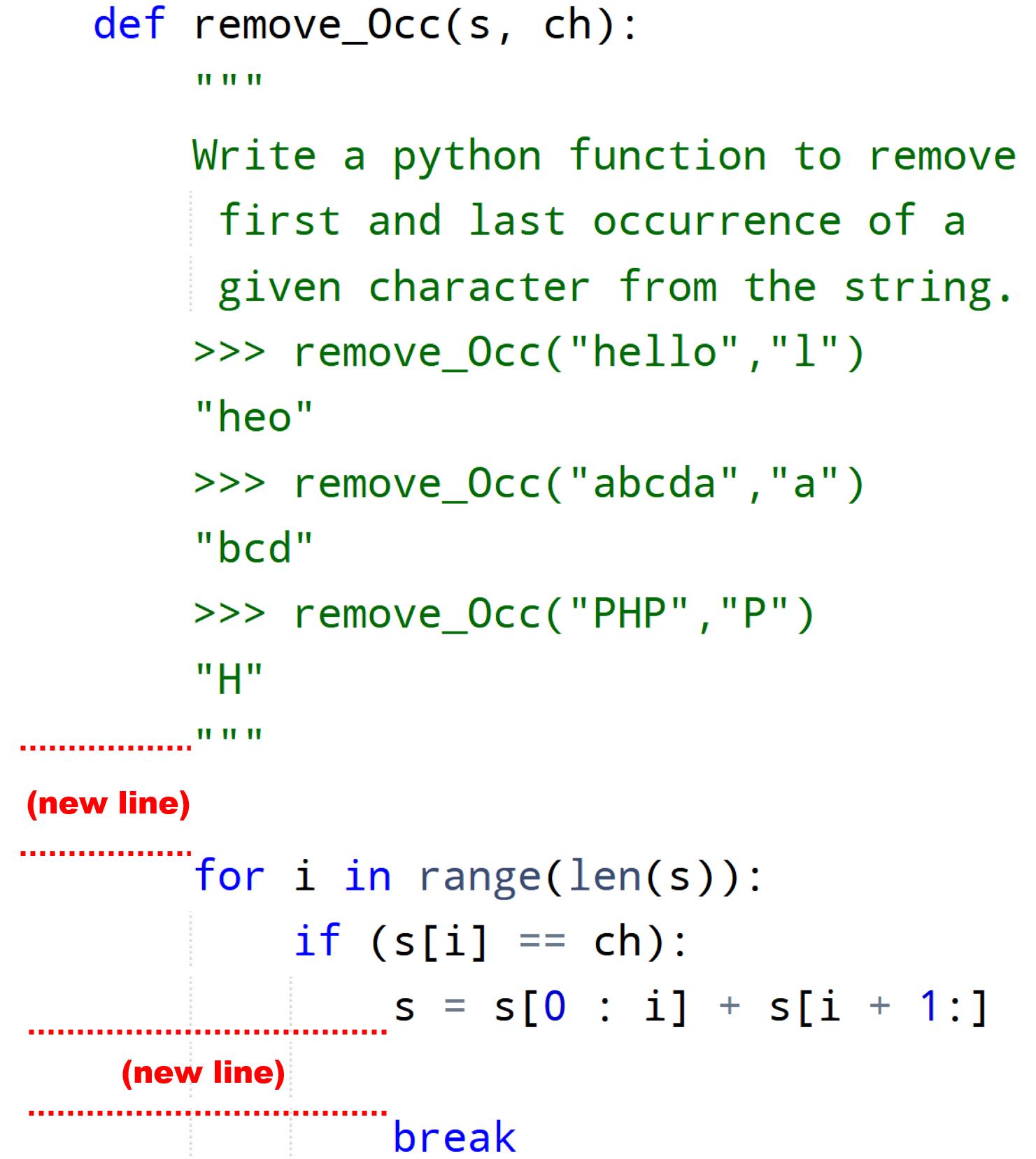}
    \caption{An example of the NewlineRandom perturbation.}
    \label{fig: NewlineRandom}
\end{figure}

\paragraph{NewlineAfterCode.}
This transformation inserts an empty line at the end of the prompt.
This transformation is deterministic.
\begin{figure}[!hbt]
    \centering
    \includegraphics[width=0.6\linewidth]{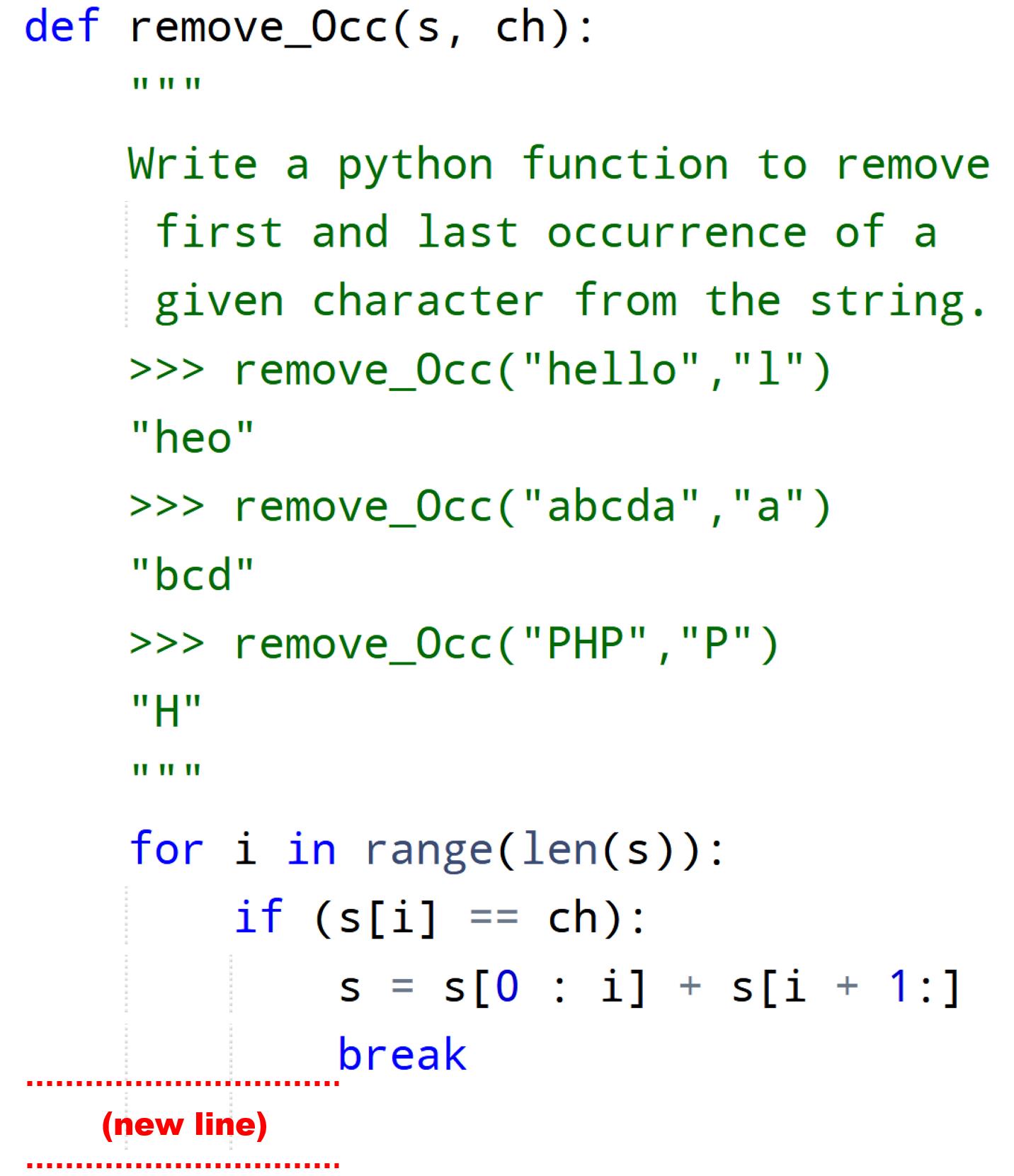}
    \caption{An example of the NewlineAfterCode perturbation.}
    \label{fig: NewlineAfterCode}
\end{figure}

\paragraph{NewlineAfterDoc.}
This transformation inserts an empty line between the docstring and the partial code.
This transformation is deterministic.
\begin{figure}[!hbt]
    \centering
    \includegraphics[width=0.6\linewidth]{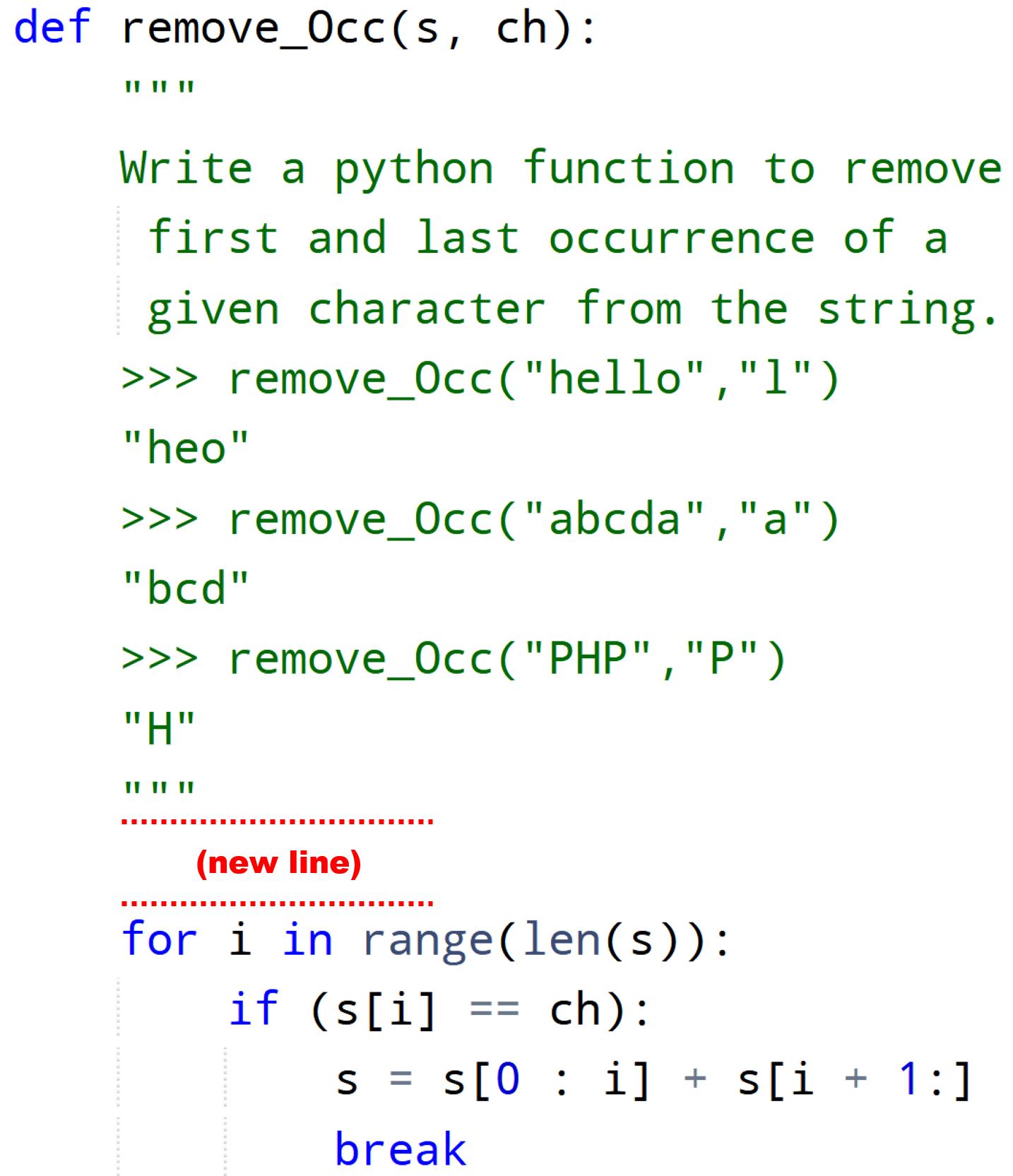}
    \caption{An example of the NewlineAfterDoc perturbation.}
    \label{fig: NewlineAfterDoc}
\end{figure}

\section{Limitations}
\label{appd: limitations}
\texttt{ReCode} benchmark has several limitations: (1) It contains perturbed datasets based on HumanEval and MBPP which focuses on Python function completion use cases. 
Therefore, we only perform evaluation on Python language and not be able to capture robustness in a wide variety of code completion use cases. However, our transformations are generalizable and could be easily extended to other languages and also other code-related datasets.
We encourage researchers to apply and extend \texttt{ReCode} benchmark to additional languages and other code-related tasks; 
(2) \texttt{ReCode} benchmark is designed for robustness evaluation and can not mitigate the lack of robustness. Given that our benchmark can be used to generate comprehensive collection of perturbed data, we believe that it can be used for training data augmentation to enhance model robustness. We will consider corresponding robust training strategy design and evaluation in future work.

\section{Failure Case Study under Perturbations}
\label{appd: error_analysis}
In this section, we showcase and analyze some failure cases on CodeGen-16B-mono and perturbed HumanEval datasets under three top perturbations that will cause significant performance drops. 

\begin{figure}[!hbt]
    \begin{subfigure}[b]{0.44\textwidth}
        \includegraphics[width=\linewidth]{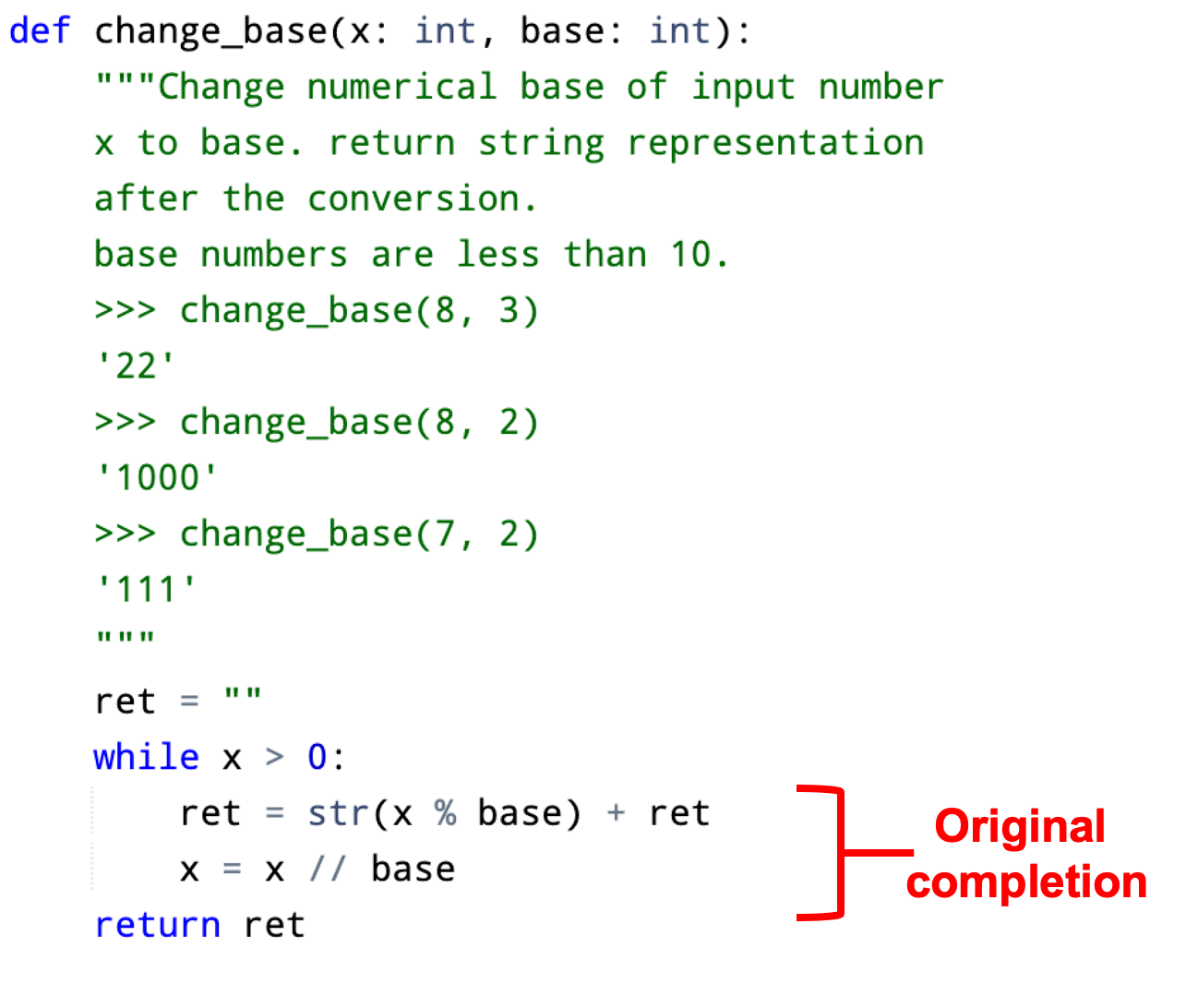}
            \caption{Correct completion without perturbation.}
            \label{subfig: appd_deadcoden}
    \end{subfigure}
    \begin{subfigure}[b]{0.4\textwidth}
        \includegraphics[width=\linewidth]{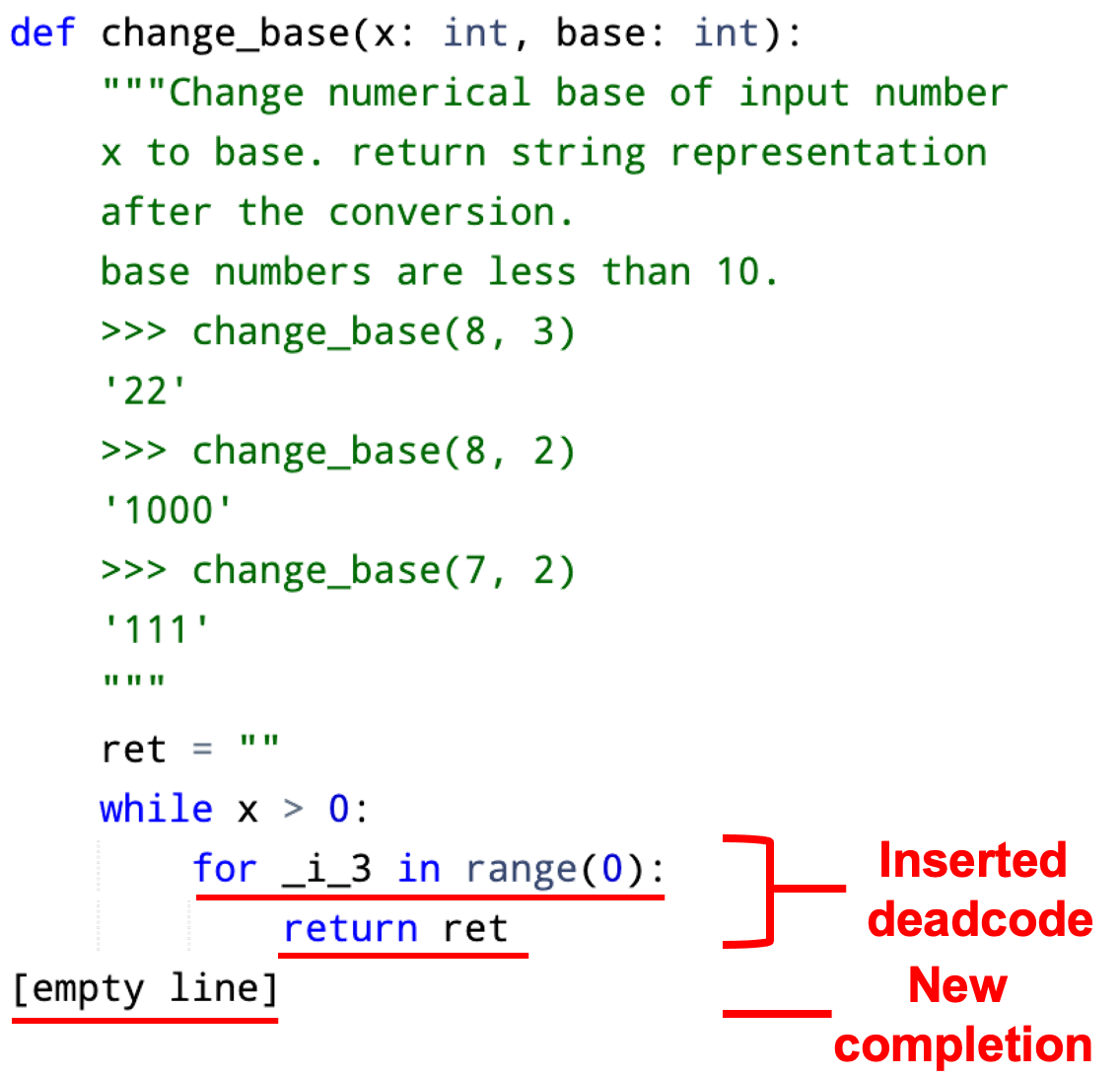}
                \caption{Wrong completion perturbed by \texttt{deadcode insertion}.}
        \label{fig: appd_deadcodep}
    \end{subfigure}
    \caption{HumanEval showcase 1 illustrating failure case under \texttt{deadcode insertion}.}
    \label{fig: appd_deadcode}
\end{figure}

DeadCode insertion is one of the most effective perturbations. It can commonly mislead the model predictions with the inserted dead code, especially when the completions are required right after the inserted dead code. \cref{fig: appd_deadcode} shows an failure example where CodeGen-mono-16B only predicts a newline after inserted meaningless for loop, which might be mislead by the inserted return statement.

\begin{figure}[!hbt]
    \begin{subfigure}[b]{0.49\textwidth}
        \includegraphics[width=\linewidth]{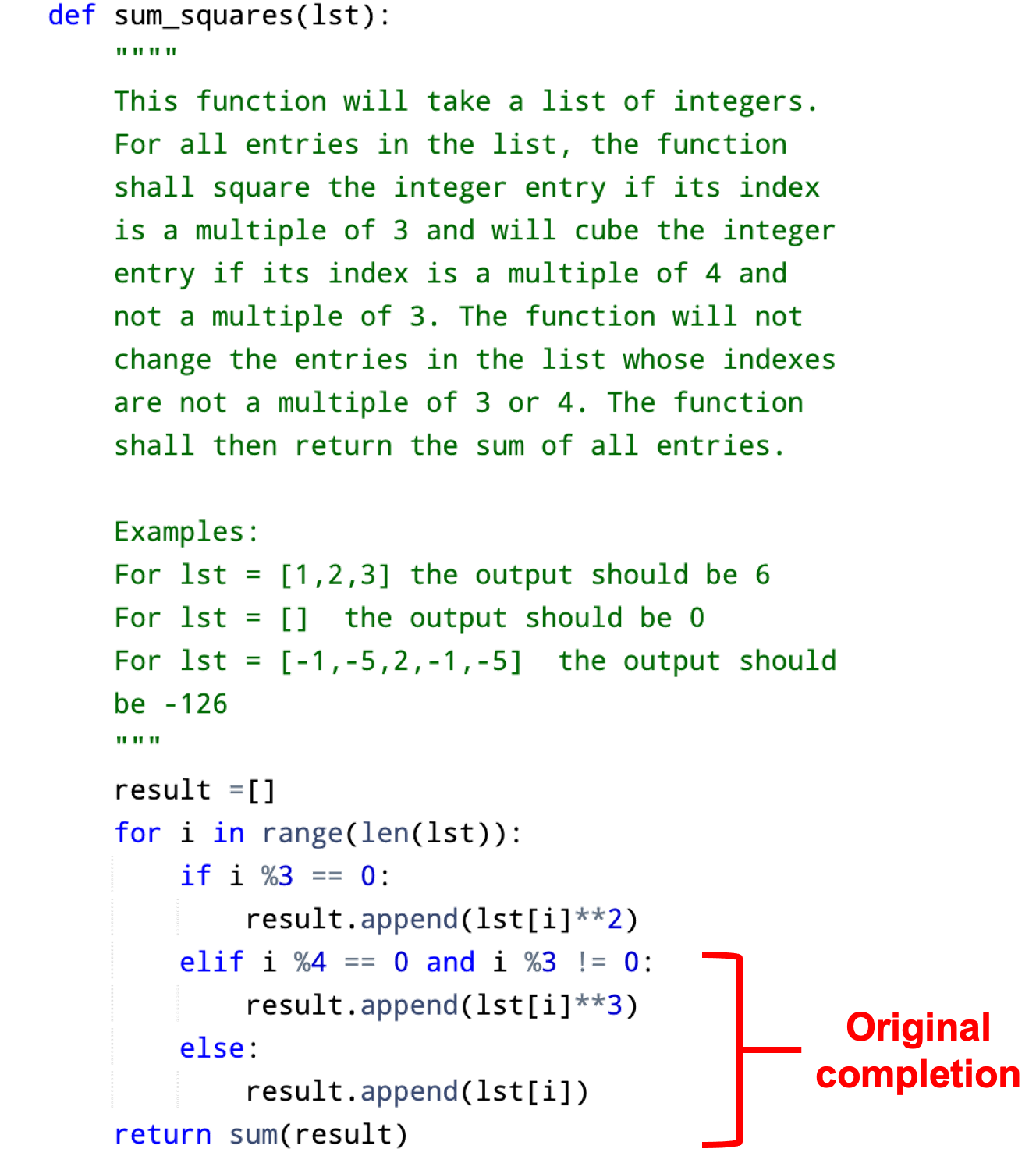}
            \caption{Correct completion without perturbation.}
            \label{subfig: appd_newlinen}
    \end{subfigure}
    \begin{subfigure}[b]{0.49\textwidth}
        \includegraphics[width=\linewidth]{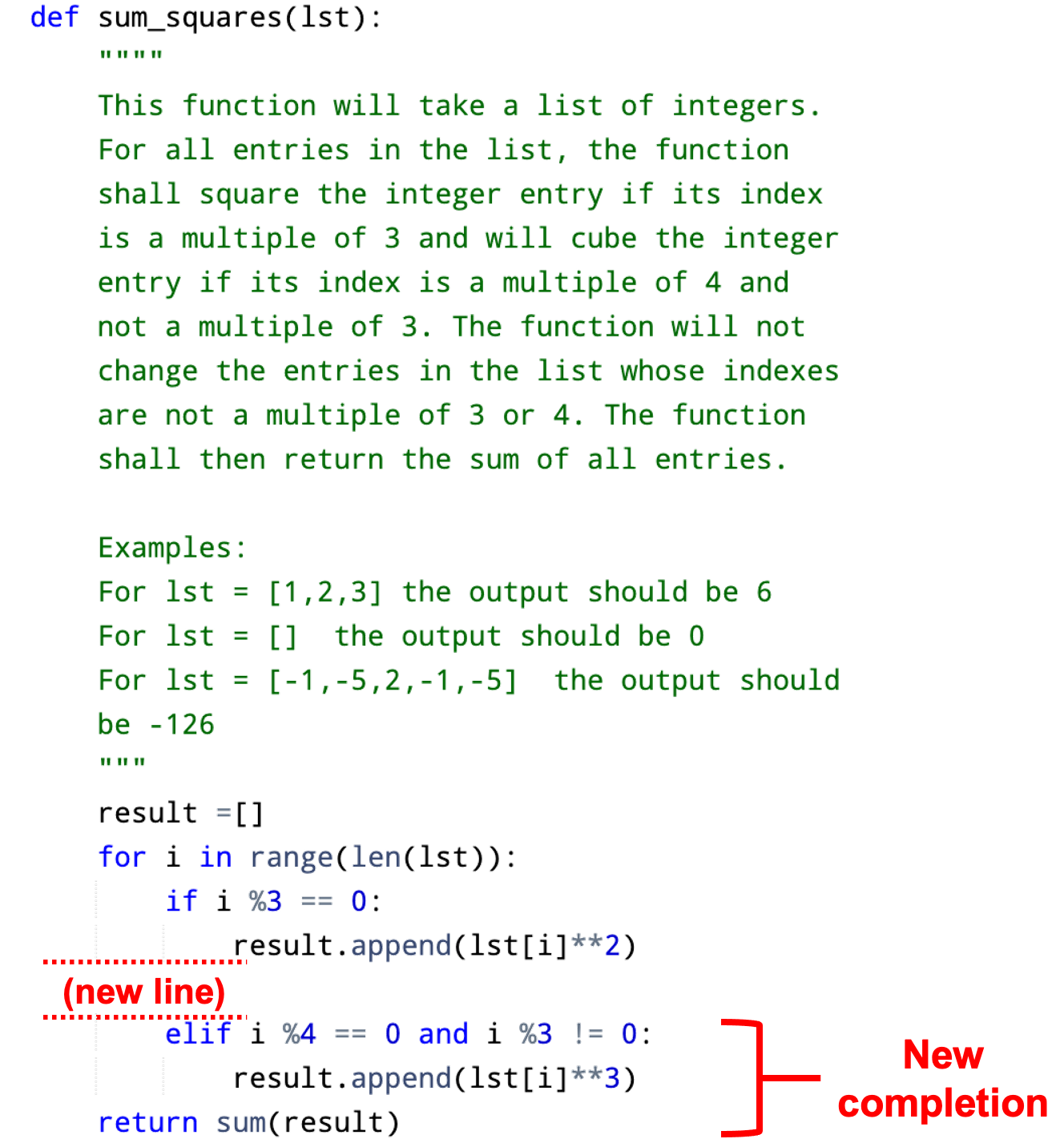}
                \caption{Wrong completion perturbed by \texttt{NewlineAfterCode} insertion.}
        \label{fig: appd_newlinep}
    \end{subfigure}
    \vspace{-20pt}
    \caption{HumanEval showcase 2 illustrating failure case under \texttt{NewlineAfterCode} insertion.}
    \label{fig: appd_newline}
    \vspace{-10pt}
\end{figure}

\cref{fig: appd_newline} shows a failure example of CodeGen-16B-mono on a prompt where an empty newline is inserted right before completion. Such simple perturbation causes wrong predictions for the following if-else conditions. It is especially effective when the required completion code is complicated.

\begin{figure}[!hbt]
    \begin{subfigure}[b]{0.49\textwidth}
        \includegraphics[width=\linewidth]{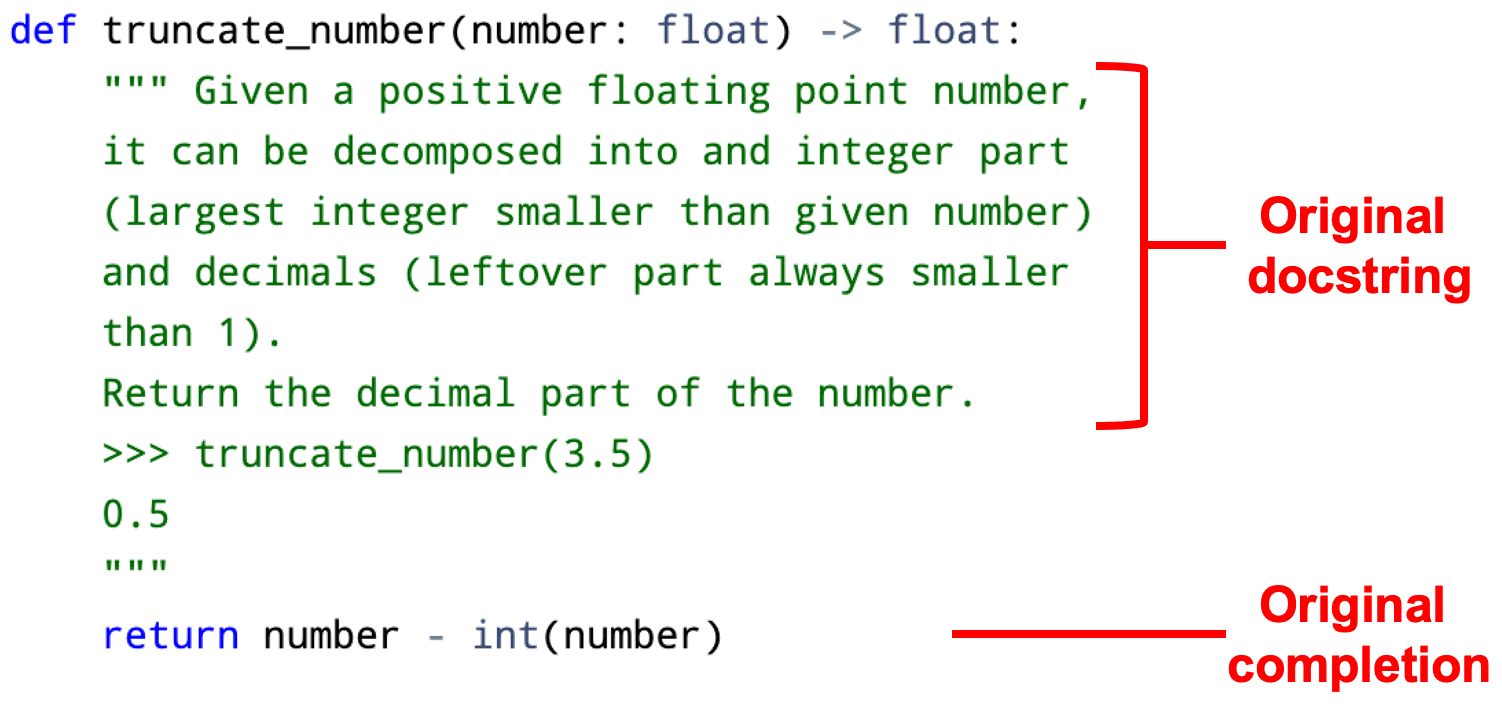}
            \caption{Correct completion without perturbation.}
            \label{subfig: appd_buttern}
    \end{subfigure}
    \begin{subfigure}[b]{0.49\textwidth}
        \includegraphics[width=\linewidth]{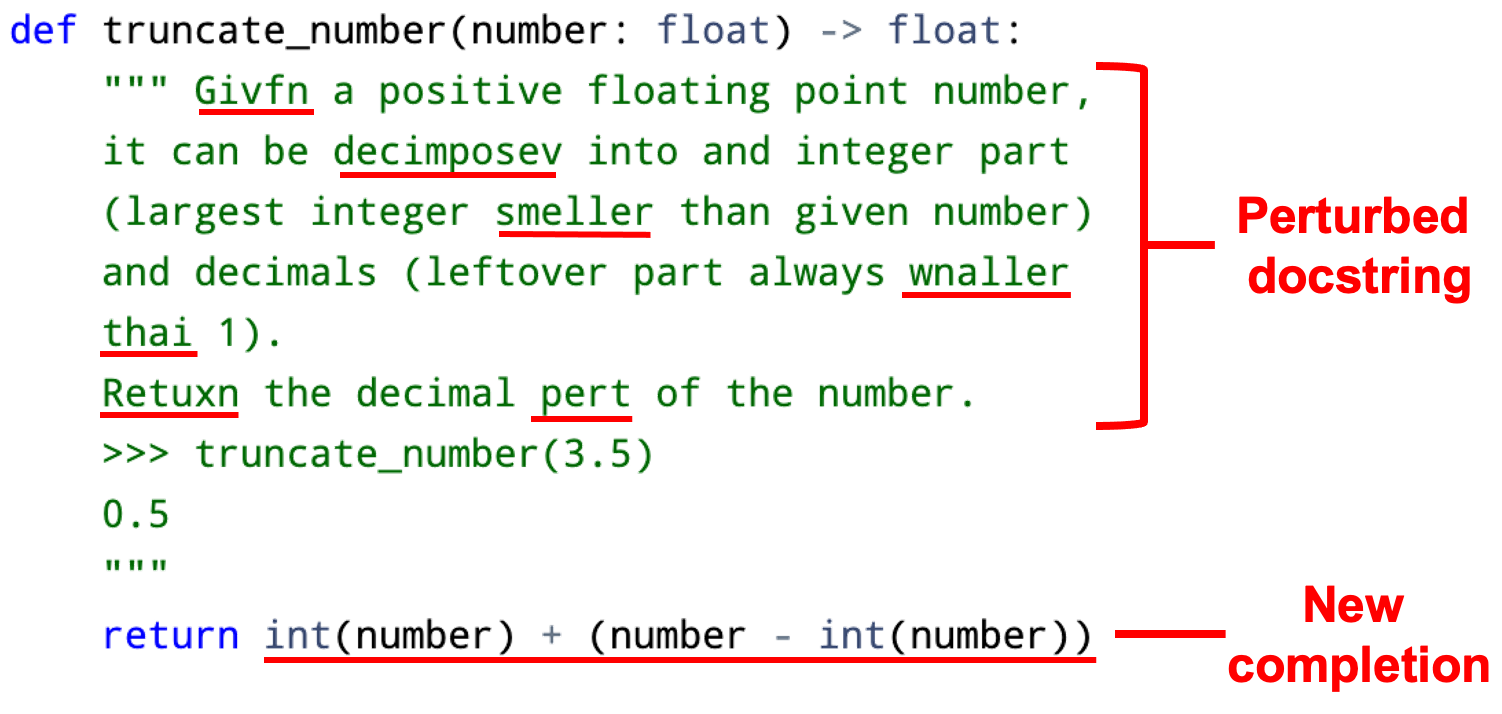}
                \caption{Wrong completion perturbed by \texttt{ButterFingers}.}
        \label{fig: appd_butterp}
    \end{subfigure}
    \vspace{-20pt}
    \caption{HumanEval showcase 3 illustrating failure case under \texttt{ButterFingers} perturbations on docstrings.}
    \label{fig: appd_butter}
    \vspace{-10pt}
\end{figure}

ButterFingers perturbation on docstring causes large performance drops as well. \cref{fig: appd_butter} shows another falure example on CodeGen-16B-mono. The typos introduced in the perturbation might cause the model to misunderstand the targeted docstrings, leading to wrong model completions.

\section{Perturbation Sample Quality}

\subsection{Details for Human Evaluation}
\label{appd: human_evaluation}

The annotators are all recruited from software engineers online who have good experience in Python via strict coding interview. To guarantee the reliability of the human evaluation results, we first conducted annotation trials with our annotators. We gave them clear definitions for each level of naturalness and semantic similarity.

We measure the inter-annotator agreement rate Fless Kappa in \cref{appd: tab_kappa}. The overall average Fleiss Kappa for the annotations is $0.52, 0.36$ for semantic and naturalness measurements on perturbed samples. The confidence interval (95\%) with bootstrap sampling (10K samples) is $[0.515, 0.528]$ and $[0.358, 0.364]$, indicating that our annotation reaches “moderate agreement” and thus our annotations are reliable~\cite{gwet2014handbook}.
 The scores from annotators are not perfectly consistent especially for naturalness since people have different preferences for code.

\begin{table}[ht]
\centering
\footnotesize
\begin{tabular}{lrr} \toprule
Fleiss Kappa          & \multicolumn{1}{c}{HumanEval} & \multicolumn{1}{c}{MBPP} \\ \midrule
Naturalness (Nominal) $\uparrow$  & 0.362                         & 0.301                    \\
Naturalness (Perturbed) $\uparrow$ & 0.435                         & 0.326                    \\
Semantics Similarity $\uparrow$  & 0.658                         & 0.461                   \\ \bottomrule
\end{tabular}
\caption{Fleiss Kappa of human evaluation.}
\label{appd: tab_kappa}
\end{table}

\subsection{Sentence Transformers for Docstring/Function Names Similarity}
\label{appd: sentrans}

In this subsection, we give experimental details for measuring the sentence similarity of perturbed and unperturbed data points using sentence transformers.

To measure the similarity scores for the docstring perturbations, we first extract docstrings from each pair of perturbed and unperturbed data points, and we use sentence transformer \texttt{all-mpnet-base-v2}~\cite{song2020mpnet} to predict an embedding vector for each docstring. Then cosine similarity is calculated and reported for each pair of perturbed and unperturbed datapoints.

Same process cannot be directly applied to function name perturbations since function names are concatenations of words instead of common sentences, barely seen by the sentence transformer training data. In order get more accurate sentence embeddings for function names, we first split each name into words (e.g., \texttt{has\_close\_elements} to \texttt{has close elements}) and then calculate the corresponding cosine similarities. 

In Table~\ref{tab: appd_sentrans}, we present the detailed results for each type of perturbations for sentence similarity. On average, we can have 0.93 and 0.92 similarity scores for docstring perturbations and 0.80 and 0.81 for function name perturbations on the HumanEval and MBPP datasets. The overall high similarity numbers provide support that our perturbations have good quality in naturalness and semantic preservation from the unperturbed inputs.

Some function name perturbations including ButterFinger, SynonymSubstitution, and CharCaseChange have relatively low sentence similarity. This is mainly because the function names only include keywords without complete sentence context and thus minor changes to each words could potentially cause large change in measured cosine similarity. For instance, character case changes on function name \texttt{intersperse} to \texttt{intErspErse} which lacks of context only has 0.21 similarity. On the other hand, the function names with more context has much higher scores, e.g., 1.0 similarity score for \texttt{has\_close\_elements} and \texttt{has\_ClosE\_Elements}.

\begin{table*}[t]
\centering
\footnotesize
\begin{tabular}{l|l|r|r} \toprule
Categories                  & Perturbations                           & \multicolumn{1}{l}{HumanEval} & \multicolumn{1}{l}{MBPP} \\ \midrule
\multirow{10}{*}{Docstring} & BackTranslation                         & 0.91                          & 0.95                     \\
                            & ButterFingers               & 0.87                          & 0.89                     \\
                            & ChangeCharCase                          & 1.00                          & 1.00                     \\
                            & EnglishInflectionalVariation            & 0.96                          & 0.93                     \\
                            & SwapCharacters              & 0.90                          & 0.87                     \\
                            & SynonymInsertion                        & 0.91                          & 0.88                     \\
                            & SynonymSubstitution                     & 0.88                          & 0.84                     \\
                            & TenseTransformationPast                 & 0.98                          & 1.00                     \\
                            & TenseTransformationFuture               & 0.97                          & 0.97                     \\
                            & Whitespace                  & 0.90                          & 0.86                     \\ \midrule
\multirow{6}{*}{Function}   &  CamelCase             & 1.00                          & 1.00                     \\
                            &  ButterFingers         & 0.57                          & 0.57                     \\
                            &  SwapCharacters       & 0.75                          & 0.75                     \\
                            &  ChangeCharCase     & 0.86                          & 0.96                     \\
                            &  InflectionalVariation & 0.94                          & 0.93                     \\
                            &  SynonymSubstition    & 0.68                          & 0.64                    
\\ \bottomrule
\end{tabular}
\caption{Cosine similarity for each type of perturbations where perturbed and unperturbed docstrings/function names are embedded by the SOTA sentence transformer.}
\label{tab: appd_sentrans}
\end{table*}

\subsection{CodeBLEU Scores for Code Similarity}
\label{appd: codebleu}

Here we present the experimental details for the CodeBLEU syntax and dataflow scores to quantitatively measure the quality of our code syntax and format transformations. 

The measurement is straightforward. The unperturbed baseline is each data point from our customized partial code datasets derived from HumanEval and MBPP. The perturbed one is the same data point transformed by each type of our perturbations. The CodeBLEU syntax and dataflow scores are then directly measured using the CodeXGLUE~\cite{lu2021codexglue} implementation.\footnote{https://github.com/microsoft/CodeXGLUE}

In Table~\ref{tab: appd_codegleu}, we present the detailed CodeBLEU results for each type of perturbations. The average numbers are summarized in Table~\ref{tab: codegleu}. Overall, 77\% and 89\% of our transformations have over 0.9 CodeBLEU syntax and dataflow scores, showing good quality in preserving semantics from the unperturbed code.
 
However, CodeBLEU syntax and dataflow are not perfect in quantitatively measuring naturalness and semantic preservation for the perturbations and thus some perturbations have expected relatively low scores: \texttt{Doc2Comments} transforms docstrings into comments causing changes of syntax; \texttt{Deadcode insertion} and \texttt{for-while switch} involve new if-conditions, loops, and new variables causing changes of code syntax and dataflow.

\begin{table*}[t]
\centering
\footnotesize
\begin{tabular}{l|l|rr|rr} \toprule
\multirow{3}{*}{Categories} & \multirow{3}{*}{Perturbations} & \multicolumn{2}{|c}{HumanEval} & \multicolumn{2}{|c}{MBPP} \\
                            &                                & CodeBLEU     & CodeBLEU       & CodeBLEU   & CodeBLEU    \\
                            &                                & (syntax)     & (dataflow)     & (syntax)   & (dataflow)  \\ \midrule
\multirow{6}{*}{Syntax}     & DeadCodeInserter               & 0.85         & 0.79           & 0.72       & 0.67        \\
                            & For-While Switch            & 0.92         & 0.90           & 0.84       & 0.86        \\
                            & OperandSwap                    & 0.91         & 1.00           & 0.90       & 1.00        \\
                            & VarRenamerCB                   & 1.00         & 0.99           & 0.93       & 0.99        \\
                            & VarRenamerNaive                & 1.00         & 0.99           & 0.93       & 0.99        \\
                            & VarRenamerRN                   & 1.00         & 0.99           & 0.93       & 0.99        \\ \midrule
\multirow{7}{*}{Format}     & Tab-Indent                    & 1.00         & 1.00           & 1.00       & 1.00        \\
                            & Line Split                   & 1.00         & 1.00           & 1.00       & 1.00        \\
                            & Doc2Comments                   & 0.84         & 1.00           & 0.76       & 1.00        \\
                            & NewlineRandom                     & 1.00         & 1.00           & 1.00       & 1.00        \\
                            & NewlineAfterCode           & 1.00         & 1.00           & 1.00       & 1.00        \\
                            & NewlineAfterDoc            & 1.00         & 1.00           & 1.00       & 1.00        \\
                            \bottomrule
\end{tabular}
\caption{CodeBLEU syntax and format similarity scores between unperturbed codes and perturbed ones with each type of our syntax and format transformations.}
\label{tab: appd_codegleu}
\end{table*}

\section{Additional Results}
\label{appd: additiona_results}

\subsection{Fine-grained Robustness Evaluation}
\label{appd: finegrained}

We present the robustness evaluation for each type of perturbations from Table~\ref{appd: tab_doc_humaneval} to~\ref{appd: tab_format_mbpp}, . The evaluation setting is the same as Table~\ref{tab: main_humaneval} and~\ref{tab: main_mbpp} where we evaluate various sizes of CodeGen~\cite{CodeGen}, InCoder~\cite{incoder}, and GPT-J~\cite{gpt-j} with greedy sampling. For each type of perturbations, we randomly generate $s=5$ different perturbed datasets derived from HumanEval and MBPP. For perturbations without randomness, only one single version of perturbed dataset is evaluated. The list of indeterministic perturbations can be found in Appendix~\ref{appd: transformation}.

\begin{table*}[t]
\centering
\footnotesize
\setlength{\tabcolsep}{3pt}
\scalebox{0.95}{
\begin{tabular}{r|r|rr|rr|rr|rr|r} \toprule
\multirow{2}{*}{HumanEval}   & \multirow{2}{*}{Metric} & CodeGen & CodeGen  & CodeGen & CodeGen  & CodeGen  & CodeGen   & InCoder & InCoder & GPT-J \\
                             &                         & 2B mono & 2B multi & 6B mono & 6B multi & 16B mono & 16B multi & 1B      & 6B      & 6B    \\ \midrule
Nominal                          & RP$_5$@1$\uparrow$       & 0.232           & 0.140            & 0.262           & 0.195            & \textbf{0.305}            & 0.195             & 0.104      & 0.152      & 0.122    \\ \midrule
\multirow{3}{*}{BackTranslation} & RP$_5$@1$\uparrow$       & 0.213           & 0.116            & 0.238           & 0.159            & \textbf{0.244}            & 0.152             & 0.098      & 0.134      & 0.098    \\
                                 & RD$_5$@1(\%)$\downarrow$     & 7.89            & 17.39            & 9.30            & 18.75            & 20.00            & 21.88             & \textbf{5.88}       & 12.00      & 20.00    \\
                                 & RR$_5$@1(\%)$\downarrow$ & 4.27            & 6.10             & 8.54            & 6.10             & 10.98            & 5.49              & \textbf{3.05}       & 3.05       & 3.66     \\ \midrule
\multirow{3}{*}{ButterFingers}        & RP$_5$@1$\uparrow$       & 0.165           & 0.098            & 0.171           & 0.122            & \textbf{0.189}            & 0.116             & 0.067      & 0.098      & 0.067    \\
        & RD$_5$@1(\%)$\downarrow$     & \textbf{28.95}           & 30.43            & 34.88           & 37.50            & 38.00            & 40.62             & 35.29      & 36.00      & 45.00    \\
        & RR$_5$@1(\%)$\downarrow$ & 10.37           & 7.32             & 15.85           & 10.37            & 20.12            & 12.20             & 7.32       & 9.15       & \textbf{6.71}     \\ \midrule
\multirow{3}{*}{ChangeCharCase}                   & RP$_5$@1$\uparrow$       & 0.152           & 0.079            & 0.152           & 0.104            & \textbf{0.177}            & 0.122             & 0.037      & 0.098      & 0.049    \\
                   & RD$_5$@1(\%)$\downarrow$     & \textbf{34.21}           & 43.48            & 41.86           & 46.88            & 42.00            & 37.50             & 64.71      & 36.00      & 60.00    \\
                   & RR$_5$@1(\%)$\downarrow$ & 12.80           & 10.98            & 15.85           & 9.76             & 17.68            & 9.15              & 10.37      & \textbf{7.32}       & 7.93     \\ \midrule
\multirow{3}{*}{\shortstack[r]{EnglishInflectional\\Variation}}     & RP$_5$@1$\uparrow$       & 0.207           & 0.134            & 0.226           & 0.171            & \textbf{0.268}            & 0.177             & 0.091      & 0.146      & 0.104    \\
     & RD$_5$@1(\%)$\downarrow$     & 10.53           & 4.35             & 13.95           & 12.50            & 12.00            & 9.38              & 11.76      & \textbf{4.00}       & 15.00    \\
     & RR$_5$@1(\%)$\downarrow$ & 3.66            & 3.05             & 8.54            & 6.10             & 7.93             & 4.27              & \textbf{1.22}       & 1.83       & 3.05     \\ \midrule
\multirow{3}{*}{\shortstack[r]{SwapCharacters\\Perturbation}}       & RP$_5$@1$\uparrow$       & 0.159           & 0.098            & 0.183           & 0.128            & \textbf{0.207}            & 0.134             & 0.085      & 0.104      & 0.067    \\
       & RD$_5$@1(\%)$\downarrow$     & 31.58           & 30.43            & 30.23           & 34.38            & 32.00            & 31.25             & \textbf{17.65}      & 32.00      & 45.00    \\
       & RR$_5$@1(\%)$\downarrow$ & 12.20           & 7.32             & 12.80           & 8.54             & 17.07            & 10.37             & \textbf{4.88}       & 10.37      & 6.10     \\ \midrule
\multirow{3}{*}{Synonym Insertion}                 & RP$_5$@1$\uparrow$       & 0.183           & 0.104            & 0.159           & 0.128            & \textbf{0.226}            & 0.128             & 0.067      & 0.104      & 0.079    \\
                 & RD$_5$@1(\%)$\downarrow$     & \textbf{21.05}           & 26.09            & 39.53           & 34.38            & 26.00            & 34.38             & 35.29      & 32.00      & 35.00    \\
                 & RR$_5$@1(\%)$\downarrow$ & 7.32            & \textbf{4.88}             & 14.63           & 8.54             & 15.85            & 9.15              & 6.10       & 9.15       & 5.49     \\ \midrule
\multirow{3}{*}{\shortstack[r]{Synonym\\Substitution}}              & RP$_5$@1$\uparrow$       & 0.146           & 0.091            & 0.159           & 0.104            & \textbf{0.201}            & 0.140             & 0.073      & 0.079      & 0.061    \\
              & RD$_5$@1(\%)$\downarrow$     & 36.84           & 34.78            & 39.53           & 46.88            & 34.00            & \textbf{28.12}             & 29.41      & 48.00      & 50.00    \\
              & RR$_5$@1(\%)$\downarrow$ & 10.37           & 6.71             & 17.07           & 10.98            & 15.24            & 7.93              & \textbf{4.88}       & 9.76       & 6.71     \\ \midrule
\multirow{3}{*}{\shortstack[r]{TenseTransformation\\Past}}          & RP$_5$@1$\uparrow$       & 0.250           & 0.146            & 0.238           & 0.189            & \textbf{0.305}            & 0.171             & 0.110      & 0.134      & 0.110    \\
          & RD$_5$@1(\%)$\downarrow$     & \textbf{-7.89}           & -4.35            & 9.30            & 3.13             & 0.00             & 12.50             & -5.88      & 12.00      & 10.00    \\
          & RR$_5$@1(\%)$\downarrow$ & 3.05            & 1.83             & 6.10            & 5.49             & 7.32             & 2.44              & 1.83       & 1.83       & \textbf{1.22}     \\ \midrule
\multirow{3}{*}{\shortstack[r]{TenseTransformation\\Future}}        & RP$_5$@1$\uparrow$       & 0.238           & 0.122            & 0.250           & 0.183            & \textbf{0.311}            & 0.171             & 0.085      & 0.146      & 0.110    \\
        & RD$_5$@1(\%)$\downarrow$     & \textbf{-2.63}           & 13.04            & 4.65            & 6.25             & -2.00            & 12.50             & 17.65      & 4.00       & 10.00    \\
        & RR$_5$@1(\%)$\downarrow$ & 4.27            & 4.27             & 4.88            & 4.88             & 6.71             & 3.66              & 1.83       & 1.83       & \textbf{1.22}     \\ \midrule
\multirow{3}{*}{\shortstack[r]{Whitespace\\Perturbation}}           & RP$_5$@1$\uparrow$       & 0.146           & 0.085            & 0.146           & 0.122            & \textbf{0.177}            & 0.122             & 0.073      & 0.091      & 0.049    \\
           & RD$_5$@1(\%)$\downarrow$     & 36.84           & 39.13            & 44.19           & 37.50            & 42.00            & 37.50             & \textbf{29.41}      & 40.00      & 60.00    \\
           & RR$_5$@1(\%)$\downarrow$ & 14.02           & 9.76             & 15.85           & 9.76             & 22.56            & 10.37             & \textbf{6.10}       & 10.98      & 7.32    \\ \bottomrule 
\end{tabular}
}
\caption{Robustness evaluation for each type of docstring perturbations on HumanEval. }
\label{appd: tab_doc_humaneval}
\end{table*}

\begin{table*}[t]
\centering
\footnotesize
\setlength{\tabcolsep}{3pt}
\scalebox{0.95}{
\begin{tabular}{r|r|rr|rr|rr|rr|r} \toprule
\multirow{2}{*}{MBPP}   & \multirow{2}{*}{Metric} & CodeGen & CodeGen  & CodeGen & CodeGen  & CodeGen  & CodeGen   & InCoder & InCoder & GPT-J \\
                             &                         & 2B mono & 2B multi & 6B mono & 6B multi & 16B mono & 16B multi & 1B      & 6B      & 6B    \\ \midrule
Nominal                      & RP$_5$@1$\uparrow$       & 0.317           & 0.191            & 0.361           & 0.221            & \textbf{0.407}            & 0.241             & 0.128      & 0.199      & 0.133    \\ \midrule
\multirow{3}{*}{\shortstack[r]{BackTranslation}}              & RP$_5$@1$\uparrow$       & 0.304           & 0.186            & 0.360           & 0.222            & \textbf{0.387}            & 0.230             & 0.119      & 0.177      & 0.128    \\
              & RD$_5$@1(\%)$\downarrow$     & 4.21            & 2.69             & 0.28            & \textbf{-0.47}            & 4.80             & 4.68              & 7.20       & 11.34      & 3.85     \\
              & RR$_5$@1(\%)$\downarrow$ & 6.26            & 6.06             & 7.91            & 6.06             & 6.26             & 7.08              & \textbf{4.00}       & 5.95       & 5.44     \\ \midrule
\multirow{3}{*}{\shortstack[r]{ButterFingers\\Perturbation}}    & RP$_5$@1$\uparrow$       & 0.210           & 0.092            & 0.240           & 0.100            & \textbf{0.280}            & 0.126             & 0.044      & 0.082      & 0.057    \\
    & RD$_5$@1(\%)$\downarrow$     & 33.66           & 51.61            & 33.52           & 54.88            & \textbf{31.06}            & 47.66             & 65.60      & 58.76      & 56.92    \\
    & RR$_5$@1(\%)$\downarrow$ & 20.43           & 21.66            & 23.72           & 22.07            & 25.26            & 23.31             & \textbf{14.48}      & 20.12      & 16.32    \\ \midrule
\multirow{3}{*}{\shortstack[r]{ChangeCharCase}}               & RP$_5$@1$\uparrow$       & 0.187           & 0.087            & 0.220           & 0.105            & \textbf{0.266}            & 0.124             & 0.053      & 0.074      & 0.055    \\
               & RD$_5$@1(\%)$\downarrow$     & 41.10           & 54.30            & 39.20           & 52.56            & \textbf{34.60}            & 48.51             & 58.40      & 62.89      & 58.46    \\
               & RR$_5$@1(\%)$\downarrow$ & 22.07           & 20.74            & 27.21           & 20.84            & 26.28            & 25.87             & \textbf{13.24}      & 21.46      & 17.45    \\ \midrule
\multirow{3}{*}{\shortstack[r]{EnglishInflectional\\Variation}} & RP$_5$@1$\uparrow$       & 0.306           & 0.161            & 0.334           & 0.198            & \textbf{0.399}            & 0.214             & 0.103      & 0.179      & 0.113    \\
 & RD$_5$@1(\%)$\downarrow$     & 3.56            & 15.59            & 7.67            & 10.23            & \textbf{1.77}             & 11.49             & 20.00      & 10.31      & 15.38    \\
 & RR$_5$@1(\%)$\downarrow$ & 8.93            & 10.78            & 11.40           & 9.65             & 10.68            & 12.53             & 7.08       & 9.45       & \textbf{6.78}     \\ \midrule
\multirow{3}{*}{\shortstack[r]{SwapCharacters\\Perturbation}}   & RP$_5$@1$\uparrow$       & 0.232           & 0.115            & 0.266           & 0.123            & \textbf{0.304}            & 0.149             & 0.059      & 0.108      & 0.063    \\
   & RD$_5$@1(\%)$\downarrow$     & 26.86           & 39.78            & 26.42           & 44.19            & \textbf{25.25}            & 38.30             & 54.40      & 45.88      & 53.08    \\
   & RR$_5$@1(\%)$\downarrow$ & 16.53           & 15.81            & 19.61           & 18.99            & 20.84            & 20.43             & \textbf{12.42}      & 14.99      & 15.30    \\ \midrule
\multirow{3}{*}{\shortstack[r]{Synonym\\Insertion}}             & RP$_5$@1$\uparrow$       & 0.238           & 0.111            & 0.263           & 0.103            & \textbf{0.290}            & 0.121             & 0.052      & 0.101      & 0.055    \\
             & RD$_5$@1(\%)$\downarrow$     & \textbf{24.92}           & 41.94            & 27.27           & 53.49            & 28.79            & 49.79             & 59.20      & 49.48      & 58.46    \\
             & RR$_5$@1(\%)$\downarrow$ & 16.63           & 18.99            & 21.25           & 20.53            & 24.44            & 24.85             & \textbf{13.14}      & 17.56      & 14.99    \\ \midrule
\multirow{3}{*}{\shortstack[r]{Synonym\\Substitution}}          & RP$_5$@1$\uparrow$       & 0.193           & 0.099            & 0.213           & 0.079            & \textbf{0.233}            & 0.092             & 0.027      & 0.064      & 0.031    \\
          & RD$_5$@1(\%)$\downarrow$     & \textbf{39.16}           & 48.39            & 41.19           & 64.19            & 42.68            & 61.70             & 79.20      & 68.04      & 76.92    \\
          & RR$_5$@1(\%)$\downarrow$ & 22.79           & 18.17            & 27.31           & 23.61            & 30.18            & 26.90             & \textbf{16.22}      & 22.38      & 17.45    \\ \midrule
\multirow{3}{*}{\shortstack[r]{TenseTransformation\\Past}}      & RP$_5$@1$\uparrow$       & 0.318           & 0.190            & 0.362           & 0.214            & \textbf{0.402}            & 0.238             & 0.120      & 0.197      & 0.141    \\
      & RD$_5$@1(\%)$\downarrow$     & -0.32           & 0.54             & -0.28           & 3.26             & 1.01             & 1.28              & 6.40       & 1.03       & \textbf{-5.38}    \\
      & RR$_5$@1(\%)$\downarrow$ & 2.16            & 2.36             & 4.00            & 3.18             & 3.29             & 2.16              & 1.64       & 2.26       & \textbf{1.54}     \\ \midrule
\multirow{3}{*}{\shortstack[r]{TenseTransformation\\Future}}    & RP$_5$@1$\uparrow$       & 0.314           & 0.197            & 0.369           & 0.218            & \textbf{0.400}            & 0.242             & 0.122      & 0.185      & 0.125    \\
    & RD$_5$@1(\%)$\downarrow$     & 0.97            & \textbf{-3.23}            & -1.99           & 1.40             & 1.52             & -0.43             & 4.80       & 7.22       & 6.15     \\
    & RR$_5$@1(\%)$\downarrow$ & 3.18            & 3.29             & 5.65            & 4.21             & 4.52             & 4.00              & 2.46       & 3.49       & \textbf{2.26}     \\ \midrule
\multirow{3}{*}{\shortstack[r]{Whitespace\\Perturbation}}       & RP$_5$@1$\uparrow$       & 0.214           & 0.107            & 0.252           & 0.106            & \textbf{0.287}            & 0.134             & 0.057      & 0.094      & 0.054    \\
       & RD$_5$@1(\%)$\downarrow$     & 32.69           & 44.09            & 30.40           & 52.09            & \textbf{29.29}            & 44.26             & 55.20      & 52.58      & 59.23    \\
       & RR$_5$@1(\%)$\downarrow$ & 20.64           & 17.66            & 21.56           & 20.64            & 25.46            & 23.31             & \textbf{12.42}      & 17.86      & 16.02   \\ \bottomrule
\end{tabular} 
}
\caption{Robustness evaluation for each type of docstring perturbations on MBPP. }
\label{appd: tab_doc_mbpp}
\end{table*}

\begin{table*}[t]
\centering
\footnotesize
\setlength{\tabcolsep}{3pt}
\scalebox{0.95}{
\begin{tabular}{r|r|rr|rr|rr|rr|r} \toprule
\multirow{2}{*}{HumanEval}   & \multirow{2}{*}{Metric} & CodeGen & CodeGen  & CodeGen & CodeGen  & CodeGen  & CodeGen   & InCoder & InCoder & GPT-J \\
&                         & 2B mono & 2B multi & 6B mono & 6B multi & 16B mono & 16B multi & 1B      & 6B      & 6B    \\ \midrule
Nominal                         & RP$_5$@1$\uparrow$       & 0.232           & 0.140            & 0.262           & 0.195            & \textbf{0.305}            & 0.195             & 0.104      & 0.152      & 0.122    \\ \midrule
\multirow{3}{*}{\shortstack[r]{CamelCase}}             & RP$_5$@1$\uparrow$       & 0.238           & 0.140            & 0.256           & 0.201            & \textbf{0.293}            & 0.165             & 0.098      & 0.152      & 0.116    \\
             & RD$_5$@1(\%)$\downarrow$     & -2.63           & 0.00             & 2.33            & \textbf{-3.13}            & 4.00             & 15.62             & 5.88       & 0.00       & 5.00     \\
             & RR$_5$@1(\%)$\downarrow$ & 1.83            & 1.22             & 3.05            & 3.05             & 3.66             & 3.05              & 3.05       & 1.22       & \textbf{0.61}     \\ \midrule
\multirow{3}{*}{\shortstack[r]{ButterFinger}}          & RP$_5$@1$\uparrow$       & 0.195           & 0.104            & 0.232           & 0.177            & \textbf{0.274}            & 0.159             & 0.098      & 0.140      & 0.091    \\
          & RD$_5$@1(\%)$\downarrow$     & 15.79           & 26.09            & 11.63           & 9.38             & 10.00            & 18.75             & \textbf{5.88}       & 8.00       & 25.00    \\
          & RR$_5$@1(\%)$\downarrow$ & 4.88            & 4.88             & 9.76            & 4.88             & 9.15             & 3.66              & 3.05       & \textbf{2.44}       & 3.05     \\ \midrule
\multirow{3}{*}{\shortstack[r]{SwapChar}}              & RP$_5$@1$\uparrow$       & 0.226           & 0.116            & 0.226           & 0.177            & \textbf{0.299}            & 0.183             & 0.073      & 0.146      & 0.116    \\
              & RD$_5$@1(\%)$\downarrow$     & 2.63            & 17.39            & 13.95           & 9.38             & \textbf{2.00}             & 6.25              & 29.41      & 4.00       & 5.00     \\
              & RR$_5$@1(\%)$\downarrow$ & 3.05            & 3.05             & 4.88            & 4.27             & 4.88             & 2.44              & 3.05       & 2.44       & \textbf{0.61}     \\ \midrule
\multirow{3}{*}{\shortstack[r]{ChangeCharCase}}            & RP$_5$@1$\uparrow$       & 0.207           & 0.122            & 0.213           & 0.140            & \textbf{0.256}            & 0.146             & 0.098      & 0.152      & 0.091    \\
            & RD$_5$@1(\%)$\downarrow$     & 10.53           & 13.04            & 18.60           & 28.12            & 16.00            & 25.00             & 5.88       & \textbf{0.00}       & 25.00    \\
            & RR$_5$@1(\%)$\downarrow$ & 7.32            & 5.49             & 10.37           & 7.93             & 10.98            & 4.88              & \textbf{4.27}       & 7.32       & 5.49     \\ \midrule
\multirow{3}{*}{\shortstack[r]{Inflectional\\Variation}} & RP$_5$@1$\uparrow$       & 0.232           & 0.134            & 0.262           & 0.195            & \textbf{0.305}            & 0.201             & 0.110      & 0.128      & 0.110    \\
 & RD$_5$@1(\%)$\downarrow$     & 0.00            & 4.35             & 0.00            & 0.00             & 0.00             & -3.13             & \textbf{-5.88}      & 16.00      & 10.00    \\
 & RR$_5$@1(\%)$\downarrow$ & 3.66            & 3.05             & 4.27            & 3.66             & 2.44             & \textbf{0.61}              & 1.83       & 2.44       & 1.22     \\ \midrule
\multirow{3}{*}{\shortstack[r]{Synonym\\Substitution}}            & RP$_5$@1$\uparrow$       & 0.195           & 0.098            & 0.232           & 0.159            & \textbf{0.305}            & 0.159             & 0.085      & 0.128      & 0.098    \\
            & RD$_5$@1(\%)$\downarrow$     & 15.79           & 30.43            & 11.63           & 18.75            & \textbf{0.00}             & 18.75             & 17.65      & 16.00      & 20.00    \\
            & RR$_5$@1(\%)$\downarrow$ & 7.32            & 6.71             & 7.93            & 6.10             & 7.32             & 3.66              & 3.05       & 4.88       & \textbf{2.44}    \\  \bottomrule
\end{tabular}
}
\caption{Robustness evaluation for each type of function name perturbations on HumanEval. }
\label{appd: tab_func_humaneval}
\end{table*}

\begin{table*}[t]
\centering
\footnotesize
\setlength{\tabcolsep}{3pt}
\scalebox{0.95}{
\begin{tabular}{r|r|rr|rr|rr|rr|r} \toprule
\multirow{2}{*}{MBPP}   & \multirow{2}{*}{Metric} & CodeGen & CodeGen  & CodeGen & CodeGen  & CodeGen  & CodeGen   & InCoder & InCoder & GPT-J \\
&                         & 2B mono & 2B multi & 6B mono & 6B multi & 16B mono & 16B multi & 1B      & 6B      & 6B    \\ \midrule
Nominal                         & RP$_5$@1$\uparrow$       & 0.317           & 0.191            & 0.361           & 0.221            & \textbf{0.407}            & 0.241             & 0.128      & 0.199      & 0.133    \\ \midrule
\multirow{3}{*}{\shortstack[r]{CamelCase}}             & RP$_5$@1$\uparrow$       & 0.316           & 0.196            & 0.367           & 0.219            & \textbf{0.408}            & 0.245             & 0.116      & 0.194      & 0.134    \\
             & RD$_5$@1(\%)$\downarrow$     & 0.32            & \textbf{-2.69}            & -1.42           & 0.93             & -0.25            & -1.70             & 9.60       & 2.58       & -0.77    \\
             & RR$_5$@1(\%)$\downarrow$ & 5.44            & 5.44             & 7.29            & 5.34             & 7.08             & 4.52              & 5.75       & 5.03       & \textbf{3.18}     \\ \midrule
\multirow{3}{*}{\shortstack[r]{ButterFinger}}          & RP$_5$@1$\uparrow$       & 0.312           & 0.185            & 0.370           & 0.203            & \textbf{0.412}            & 0.231             & 0.110      & 0.175      & 0.117    \\
          & RD$_5$@1(\%)$\downarrow$     & 1.62            & 3.23             & \textbf{-2.27}           & 7.91             & -1.26            & 4.26              & 14.40      & 12.37      & 12.31    \\
          & RR$_5$@1(\%)$\downarrow$ & 7.19            & 8.62             & 9.65            & 10.99            & 8.73             & 9.86              & \textbf{6.67}       & 8.11       & 6.98     \\ \midrule
\multirow{3}{*}{\shortstack[r]{SwapChar}}              & RP$_5$@1$\uparrow$       & 0.309           & 0.189            & 0.342           & 0.202            & \textbf{0.399}            & 0.237             & 0.116      & 0.171      & 0.113    \\
              & RD$_5$@1(\%)$\downarrow$     & 2.59            & \textbf{1.08}             & 5.40            & 8.37             & 1.77             & 1.70              & 9.60       & 13.92      & 15.38    \\
              & RR$_5$@1(\%)$\downarrow$ & 4.41            & 4.52             & 7.29            & 6.88             & 6.06             & 4.52              & \textbf{3.18}       & 5.24       & 4.21     \\ \midrule
\multirow{3}{*}{\shortstack[r]{ChangeCharCase}}            & RP$_5$@1$\uparrow$       & 0.295           & 0.179            & 0.346           & 0.192            & \textbf{0.400}            & 0.244             & 0.093      & 0.171      & 0.111    \\
            & RD$_5$@1(\%)$\downarrow$     & 7.12            & 6.45             & 4.26            & 13.02            & 1.52             & \textbf{-1.28}             & 27.20      & 13.92      & 16.92    \\
            & RR$_5$@1(\%)$\downarrow$ & 9.55            & 10.88            & 11.91           & 12.22            & 12.73            & 9.75              & \textbf{8.32}       & 10.57      & 9.45     \\ \midrule
\multirow{3}{*}{\shortstack[r]{Inflectional\\Variation}} & RP$_5$@1$\uparrow$       & 0.318           & 0.187            & 0.343           & 0.202            & \textbf{0.402}            & 0.243             & 0.128      & 0.188      & 0.125    \\
 & RD$_5$@1(\%)$\downarrow$     & -0.32           & 2.15             & 5.11            & 8.37             & 1.01             & \textbf{-0.85}             & 0.00       & 5.67       & 6.15     \\
 & RR$_5$@1(\%)$\downarrow$ & 3.08            & 4.31             & 6.88            & 5.75             & 5.95             & 4.31              & \textbf{2.46}       & 2.98       & 3.49     \\ \midrule
\multirow{3}{*}{\shortstack[r]{Synonym\\Substitution}}            & RP$_5$@1$\uparrow$       & 0.316           & 0.186            & 0.346           & 0.197            & \textbf{0.384}            & 0.243             & 0.105      & 0.164      & 0.117    \\
            & RD$_5$@1(\%)$\downarrow$     & 0.32            & 2.69             & 4.26            & 10.70            & 5.56             & \textbf{-0.85}             & 18.40      & 17.53      & 12.31    \\
            & RR$_5$@1(\%)$\downarrow$ & 6.88            & 7.49             & 10.88           & 10.47            & 9.96             & 9.86              & 7.70       & 8.52       & \textbf{6.88}    \\ \bottomrule
\end{tabular}
}
\caption{Robustness evaluation for each type of function name perturbations on MBPP. }
\label{appd: tab_func_mbpp}
\end{table*}

\begin{table*}[t]
\centering
\footnotesize
\setlength{\tabcolsep}{3pt}
\scalebox{0.95}{
\begin{tabular}{r|r|rr|rr|rr|rr|r} \toprule
\multirow{2}{*}{HumanEval}   & \multirow{2}{*}{Metric} & CodeGen & CodeGen  & CodeGen & CodeGen  & CodeGen  & CodeGen   & InCoder & InCoder & GPT-J \\
&                         & 2B mono & 2B multi & 6B mono & 6B multi & 16B mono & 16B multi & 1B      & 6B      & 6B    \\ \midrule
Nominal                  & RP$_5$@1$\uparrow$       & 0.402           & 0.293            & 0.518           & 0.366            & \textbf{0.549}            & 0.390             & 0.189      & 0.323      & 0.250    \\ \midrule
\multirow{3}{*}{\shortstack[r]{DeadCodeInserter}}    & RP$_5$@1$\uparrow$         & 0.116                    & 0.079                     & 0.152                    & 0.110                     & \textbf{0.159}                     & 0.091                      & 0.055               & 0.079               & 0.079             \\
    & RD$_5$@1(\%)$\downarrow$       & 71.21                    & 72.92                     & 70.59                    & 70.00                     & 71.11                     & 76.56                      & 70.97               & 75.47               & \textbf{68.29}             \\
    & RR$_5$@1(\%)$\downarrow$   & 37.80                    & 30.49                     & 41.46                    & 32.93                     & 45.12                     & 37.20                      & \textbf{17.07}               & 30.49               & 27.44             \\ \midrule
\multirow{3}{*}{\shortstack[r]{ForWhile\\TransformerFirst}} & RP$_5$@1$\uparrow$         & 0.384                    & 0.226                     & 0.500                    & 0.305                     & \textbf{0.537}                     & 0.384                      & 0.159               & 0.280               & 0.213             \\
 & RD$_5$@1(\%)$\downarrow$       & 4.55                     & 22.92                     & 3.53                     & 16.67                     & 2.22                      & \textbf{1.56}                       & 16.13               & 13.21               & 14.63             \\
 & RR$_5$@1(\%)$\downarrow$   & \textbf{5.49}                     & 6.71                      & 9.15                     & 8.54                      & 6.10                      & \textbf{5.49}                       & \textbf{5.49}                & 6.71                & 9.76             \\ \midrule
\multirow{3}{*}{\shortstack[r]{OperandSwap}}              & RP$_5$@1$\uparrow$       & 0.402           & 0.274            & 0.500           & 0.348            & \textbf{0.512}            & 0.354             & 0.171      & 0.311      & 0.220    \\
              & RD$_5$@1(\%)$\downarrow$     & \textbf{0.00}            & 6.25             & 3.53            & 5.00             & 6.67             & 9.38              & 9.68       & 3.77       & 12.20    \\
              & RR$_5$@1(\%)$\downarrow$ & 6.71            & \textbf{4.27}             & 6.71            & 6.10             & 5.49             & 6.71              & 6.10       & 7.93       & 7.32     \\ \midrule
\multirow{3}{*}{\shortstack[r]{VarRenamerCB}}             & RP$_5$@1$\uparrow$       & 0.415           & 0.268            & 0.476           & 0.329            & \textbf{0.518}            & 0.354             & 0.146      & 0.287      & 0.238    \\
             & RD$_5$@1(\%)$\downarrow$     & \textbf{-3.03}           & 8.33             & 8.24            & 10.00            & 5.56             & 9.38              & 22.58      & 11.32      & 4.88     \\
             & RR$_5$@1(\%)$\downarrow$ & \textbf{4.88}            & 6.10             & 6.71            & 8.54             & 5.49             & 7.32              & 7.93       & 8.54       & 4.88     \\ \midrule
\multirow{3}{*}{\shortstack[r]{VarRenamerNaive}}          & RP$_5$@1$\uparrow$       & 0.396           & 0.244            & 0.482           & 0.348            & \textbf{0.494}            & 0.341             & 0.177      & 0.280      & 0.220    \\
          & RD$_5$@1(\%)$\downarrow$     & \textbf{1.52}            & 16.67            & 7.06            & 5.00             & 10.00            & 12.50             & 6.45       & 13.21      & 12.20    \\
          & RR$_5$@1(\%)$\downarrow$ & \textbf{4.27}            & 9.76             & 7.32            & 9.15             & 6.71             & 8.54              & 9.76       & 10.37      & 5.49     \\ \midrule
\multirow{3}{*}{\shortstack[r]{VarRenamerRN}}             & RP$_5$@1$\uparrow$       & 0.366           & 0.207            & 0.421           & 0.280            & \textbf{0.470}            & 0.280             & 0.085      & 0.152      & 0.177    \\
             & RD$_5$@1(\%)$\downarrow$     & \textbf{9.09}            & 29.17            & 18.82           & 23.33            & 14.44            & 28.12             & 54.84      & 52.83      & 29.27    \\
             & RR$_5$@1(\%)$\downarrow$ & 12.20           & 14.63            & 14.02           & 12.80            & \textbf{11.59}            & 17.07             & 16.46      & 24.39      & 12.20   \\ \bottomrule
\end{tabular}
}
\caption{Robustness evaluation for each type of code syntax perturbations on HumanEval. }
\label{appd: tab_syntax_humaneval}
\end{table*}

\begin{table*}[t]
\centering
\footnotesize
\setlength{\tabcolsep}{3pt}
\scalebox{0.95}{
\begin{tabular}{r|r|rr|rr|rr|rr|r} \toprule
\multirow{2}{*}{MBPP}   & \multirow{2}{*}{Metric} & CodeGen & CodeGen  & CodeGen & CodeGen  & CodeGen  & CodeGen   & InCoder & InCoder & GPT-J \\
&                         & 2B mono & 2B multi & 6B mono & 6B multi & 16B mono & 16B multi & 1B      & 6B      & 6B    \\ \midrule
Nominal                  & RP$_5$@1$\uparrow$       & 0.450           & 0.285            & 0.535           & 0.331            & \textbf{0.571}            & 0.379             & 0.219      & 0.292      & 0.176    \\ \midrule
\multirow{3}{*}{\shortstack[r]{DeadCodeInserter}}    & RP$_5$@1$\uparrow$         & 0.043                    & 0.020                     & 0.044                    & 0.024                     & \textbf{0.055}                     & 0.025                      & 0.015               & 0.015               & 0.009             \\
    & RD$_5$@1(\%)$\downarrow$       & 90.41                    & 93.17                     & 91.75                    & 92.86                     & \textbf{90.29}                     & 93.50                      & 92.96               & 94.72               & 94.74             \\
    & RR$_5$@1(\%)$\downarrow$   & 52.05                    & 37.99                     & 57.39                    & 39.12                     & 60.57                     & 44.87                      & 29.26               & 37.78               & \textbf{24.95}             \\ \midrule
\multirow{3}{*}{\shortstack[r]{ForWhile\\TransformerFirst}} & RP$_5$@1$\uparrow$         & 0.432                    & 0.259                     & 0.497                    & 0.303                     & \textbf{0.532}                     & 0.346                      & 0.182               & 0.245               & 0.149             \\
 & RD$_5$@1(\%)$\downarrow$       & \textbf{3.88}                     & 9.35                      & 7.10                     & 8.39                      & 6.83                      & 8.67                       & 16.90               & 15.85               & 15.20             \\
 & RR$_5$@1(\%)$\downarrow$   & 13.66                    & 12.94                     & 11.60                    & 13.45                     & 11.70                     & 13.35                      & 12.73               & 16.53               & \textbf{9.24}        \\ \midrule
\multirow{3}{*}{\shortstack[r]{OperandSwap}}              & RP$_5$@1$\uparrow$       & 0.450           & 0.275            & 0.506           & 0.321            & \textbf{0.544}            & 0.379             & 0.225      & 0.276      & 0.211    \\
              & RD$_5$@1(\%)$\downarrow$     & 0.00            & 3.60             & 5.37            & 2.80             & 4.68             & 0.00              & -2.82      & 5.28       & \textbf{-20.47}   \\
              & RR$_5$@1(\%)$\downarrow$ & 13.24           & 11.81            & \textbf{10.57}           & 13.45            & 11.81            & 12.32             & 12.32      & 15.50      & 11.91    \\ \midrule
\multirow{3}{*}{\shortstack[r]{VarRenamerCB}}             & RP$_5$@1$\uparrow$       & 0.428           & 0.263            & 0.475           & 0.307            & \textbf{0.511}            & 0.359             & 0.194      & 0.247      & 0.207    \\
             & RD$_5$@1(\%)$\downarrow$     & 4.79            & 7.91             & 11.13           & 7.14             & 10.43            & 5.15              & 11.27      & 15.14      & \textbf{-18.13}   \\
             & RR$_5$@1(\%)$\downarrow$ & 15.30           & 13.96            & 15.20           & 15.50            & 14.17            & 14.07             & 13.14      & 16.12      & \textbf{12.83}    \\ \midrule
\multirow{3}{*}{\shortstack[r]{VarRenamerNaive}}          & RP$_5$@1$\uparrow$       & 0.417           & 0.240            & 0.461           & 0.286            & \textbf{0.513}            & 0.338             & 0.171      & 0.226      & 0.172    \\
          & RD$_5$@1(\%)$\downarrow$     & 7.31            & 15.83            & 13.82           & 13.35            & 10.07            & 10.84             & 21.60      & 22.54      & \textbf{1.75}     \\
          & RR$_5$@1(\%)$\downarrow$ & 16.63           & 15.61            & 17.04           & 17.97            & 14.78            & 16.43             & 14.17      & 18.07      & \textbf{13.24}    \\ \midrule
\multirow{3}{*}{\shortstack[r]{VarRenamerRN}}             & RP$_5$@1$\uparrow$       & 0.355           & 0.191            & 0.405           & 0.205            & \textbf{0.426}            & 0.259             & 0.114      & 0.168      & 0.114    \\
             & RD$_5$@1(\%)$\downarrow$     & \textbf{21.00}           & 33.09            & 24.38           & 37.89            & 25.36            & 31.71             & 47.89      & 42.25      & 35.09    \\
             & RR$_5$@1(\%)$\downarrow$ & 22.90           & 22.90            & 23.82           & 26.59            & 24.95            & 26.28             & 23.82      & 25.87      & \textbf{19.40}    \\ \bottomrule
\end{tabular}
}
\caption{Robustness evaluation for each type of code syntax perturbations on MBPP. }
\label{appd: tab_syntax_mbpp}
\end{table*}

\begin{table*}[t]
\centering
\footnotesize
\setlength{\tabcolsep}{3pt}
\scalebox{0.95}{
\begin{tabular}{r|r|rr|rr|rr|rr|r} \toprule
\multirow{2}{*}{HumanEval}   & \multirow{2}{*}{Metric} & CodeGen & CodeGen  & CodeGen & CodeGen  & CodeGen  & CodeGen   & InCoder & InCoder & GPT-J \\
&                         & 2B mono & 2B multi & 6B mono & 6B multi & 16B mono & 16B multi & 1B      & 6B      & 6B    \\ \midrule
Nominal              & RP$_5$@1$\uparrow$       & 0.402           & 0.293            & 0.518           & 0.366            & \textbf{0.549}            & 0.390             & 0.189      & 0.323      & 0.250    \\ \midrule
\multirow{3}{*}{\shortstack[r]{Tab-Indent}}          & RP$_5$@1$\uparrow$       & 0.415           & 0.305            & 0.518           & 0.354            & \textbf{0.561}            & 0.396             & 0.146      & 0.299      & 0.244    \\
          & RD$_5$@1(\%)$\downarrow$     & -3.03           & \textbf{-4.17}            & 0.00            & 3.33             & -2.22            & -1.56             & 22.58      & 7.55       & 2.44     \\
          & RR$_5$@1(\%)$\downarrow$ & \textbf{3.66}            & 4.88             & 8.54            & 4.88             & 3.66             & 4.27              & 7.93       & 9.76       & 7.93     \\ \midrule
\multirow{3}{*}{\shortstack[r]{Line Split}}         & RP$_5$@1$\uparrow$       & 0.384           & 0.274            & 0.500           & 0.378            & \textbf{0.524}            & 0.390             & 0.171      & 0.305      & 0.244    \\
         & RD$_5$@1(\%)$\downarrow$     & 4.55            & 6.25             & 3.53            & \textbf{-3.33}            & 4.44             & 0.00              & 9.68       & 5.66       & 2.44     \\
         & RR$_5$@1(\%)$\downarrow$ & 3.05            & 4.27             & 4.27            & 4.88             & 3.66             & \textbf{2.44}              & 3.05       & 6.71       & 4.27     \\ \midrule
\multirow{3}{*}{\shortstack[r]{Doc2Comments}}         & RP$_5$@1$\uparrow$       & 0.335           & 0.287            & 0.433           & 0.293            & \textbf{0.457}            & 0.335             & 0.146      & 0.293      & 0.195    \\
         & RD$_5$@1(\%)$\downarrow$     & 16.67           & \textbf{2.08}             & 16.47           & 20.00            & 16.67            & 14.06             & 22.58      & 9.43       & 21.95    \\
         & RR$_5$@1(\%)$\downarrow$ & 11.59           & 5.49             & 14.63           & 8.54             & 12.80            & 7.93              & \textbf{4.27}       & 5.49       & 10.37    \\ \midrule
\multirow{3}{*}{\shortstack[r]{NewlineRandom}}           & RP$_5$@1$\uparrow$       & 0.360           & 0.220            & 0.390           & 0.250            & \textbf{0.457}            & 0.299             & 0.152      & 0.232      & 0.171    \\
           & RD$_5$@1(\%)$\downarrow$     & \textbf{10.61}           & 25.00            & 24.71           & 31.67            & 16.67            & 23.44             & 19.35      & 28.30      & 31.71    \\
           & RR$_5$@1(\%)$\downarrow$ & 12.20           & 10.98            & 17.68           & 15.85            & 11.59            & 13.41             & \textbf{7.32}       & 15.85      & 12.20    \\ \midrule
\multirow{3}{*}{\shortstack[r]{NewlineAfterCode}} & RP$_5$@1$\uparrow$       & 0.409           & 0.262            & 0.494           & 0.311            & \textbf{0.537}            & 0.335             & 0.165      & 0.287      & 0.183    \\
 & RD$_5$@1(\%)$\downarrow$     &\textbf{ -1.52}           & 10.42            & 4.71            & 15.00            & 2.22             & 14.06             & 12.90      & 11.32      & 26.83    \\
 & RR$_5$@1(\%)$\downarrow$ & 6.71            & 7.93             & 8.54            & 9.15             & \textbf{3.66}             & 7.93              & 4.88       & 8.54       & 9.15     \\ \midrule
\multirow{3}{*}{\shortstack[r]{NewlineAfterDoc}}  & RP$_5$@1$\uparrow$       & 0.396           & 0.274            & 0.518           & 0.348            & \textbf{0.549}            & 0.384             & 0.183      & 0.311      & 0.244    \\
  & RD$_5$@1(\%)$\downarrow$     & 1.52            & 6.25             & 0.00            & 5.00             & \textbf{0.00}             & 1.56              & 3.23       & 3.77       & 2.44     \\
  & RR$_5$@1(\%)$\downarrow$ & 4.27            & 4.27             & 6.10            & 4.27             & 1.22             & 1.83              & \textbf{0.61}       & 3.66       & 4.27    \\ \bottomrule
\end{tabular}
}
\caption{Robustness evaluation for each type of code format perturbations on HumanEval. }
\label{appd: tab_format_humaneval}
\end{table*}

\begin{table*}[t]
\centering
\footnotesize
\setlength{\tabcolsep}{3pt}
\scalebox{0.95}{
\begin{tabular}{r|r|rr|rr|rr|rr|r} \toprule
\multirow{2}{*}{MBPP}   & \multirow{2}{*}{Metric} & CodeGen & CodeGen  & CodeGen & CodeGen  & CodeGen  & CodeGen   & InCoder & InCoder & GPT-J \\
&                         & 2B mono & 2B multi & 6B mono & 6B multi & 16B mono & 16B multi & 1B      & 6B      & 6B    \\ \midrule
Nominal              & RP$_5$@1$\uparrow$       & 0.450           & 0.285            & 0.535           & 0.331            & \textbf{0.571}            & 0.379             & 0.219      & 0.292      & 0.176    \\ \midrule
\multirow{3}{*}{\shortstack[r]{Tab-Indent}}          & RP$_5$@1$\uparrow$       & 0.452           & 0.302            & 0.530           & 0.339            & \textbf{0.566}            & 0.385             & 0.208      & 0.325      & 0.176    \\
          & RD$_5$@1(\%)$\downarrow$     & -0.46           & -5.76            & 0.96            & -2.48            & 0.90             & -1.63             & 4.69       & \textbf{-11.62}     & 0.00     \\
          & RR$_5$@1(\%)$\downarrow$ & 6.37            & 6.98             & \textbf{5.85}            & 8.01             & 6.88             & 6.78              & 9.24       & 12.22      & 7.60     \\ \midrule
\multirow{3}{*}{\shortstack[r]{Line Split}}         & RP$_5$@1$\uparrow$       & 0.445           & 0.275            & 0.524           & 0.326            & \textbf{0.556}            & 0.378             & 0.187      & 0.283      & 0.163    \\
         & RD$_5$@1(\%)$\downarrow$     & 1.14            & 3.60             & 2.11            & 1.24             & 2.52             & \textbf{0.27}              & 14.55      & 2.82       & 7.02     \\
         & RR$_5$@1(\%)$\downarrow$ & 4.41            & 6.37             & 4.41            & 6.16             & 5.54             & 6.26              & 6.06       & 6.78       & \textbf{3.90}     \\ \midrule
\multirow{3}{*}{\shortstack[r]{Doc2Comments}}         & RP$_5$@1$\uparrow$       & 0.435           & 0.269            & 0.476           & 0.299            & 0.\textbf{529}            & 0.342             & 0.169      & 0.264      & 0.172    \\
         & RD$_5$@1(\%)$\downarrow$     & 3.20            & 5.76             & 10.94           & 9.63             & 7.37             & 9.76              & 22.54      & 9.51       & \textbf{1.75}     \\
         & RR$_5$@1(\%)$\downarrow$ & \textbf{6.16}            & 8.62             & 8.32            & 9.14             & 8.93             & 11.29             & 7.19       & 8.32       & 6.47     \\ \midrule
\multirow{3}{*}{\shortstack[r]{NewlineRandom}}           & RP$_5$@1$\uparrow$       & 0.375           & 0.181            & 0.335           & 0.198            & \textbf{0.470}            & 0.262             & 0.123      & 0.159      & 0.104    \\
           & RD$_5$@1(\%)$\downarrow$     & \textbf{16.67}           & 36.69            & 37.43           & 40.06            & 17.63            & 30.89             & 43.66      & 45.42      & 40.94    \\
           & RR$_5$@1(\%)$\downarrow$ & 12.94           & 16.32            & 23.72           & 19.40            & 15.81            & 17.56             & 12.73      & 16.63      & \textbf{10.37}    \\ \midrule
\multirow{3}{*}{\shortstack[r]{NewlineAfterCode}} & RP$_5$@1$\uparrow$       & 0.406           & 0.238            & 0.379           & 0.240            & \textbf{0.525}            & 0.291             & 0.165      & 0.207      & 0.150    \\
 & RD$_5$@1(\%)$\downarrow$     & 9.82            & 16.55            & 29.17           & 27.33            & \textbf{8.09}             & 23.31             & 24.41      & 28.87      & 14.62    \\
 & RR$_5$@1(\%)$\downarrow$ & 9.14            & 10.27            & 19.51           & 13.55            & 10.99            & 14.17             & 9.03       & 12.73      & \textbf{7.91}     \\ \midrule
\multirow{3}{*}{\shortstack[r]{NewlineAfterDoc}}  & RP$_5$@1$\uparrow$       & 0.449           & 0.274            & 0.518           & 0.305            & \textbf{0.570}            & 0.378             & 0.180      & 0.242      & 0.153    \\
  & RD$_5$@1(\%)$\downarrow$     & 0.23            & 3.96             & 3.07            & 7.76             & \textbf{0.18}             & 0.27              & 17.84      & 16.90      & 12.87    \\
  & RR$_5$@1(\%)$\downarrow$ & \textbf{2.36}            & 4.41             & 4.11            & 7.08             & 4.62             & 4.41              & 4.72       & 6.57       & 4.31    \\ \bottomrule
\end{tabular}
}
\caption{Robustness evaluation for each type of code format perturbations on MBPP. }
\label{appd: tab_format_mbpp}
\end{table*}

\subsection{Additional Results for Different k}
\label{appd: k}

\begin{figure*}
    \centering
    \includegraphics[width=0.31\linewidth]{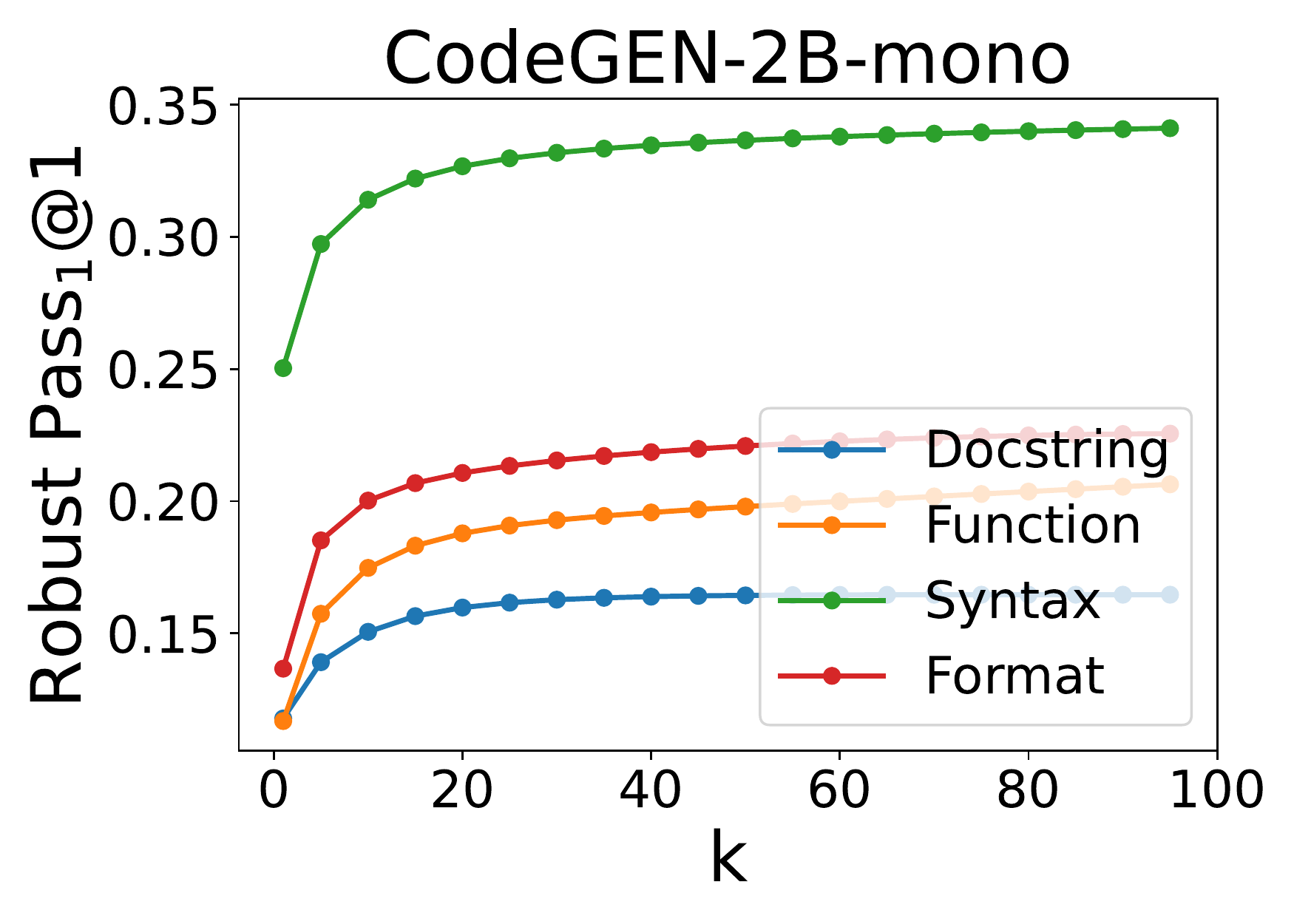}
    \includegraphics[width=0.31\linewidth]{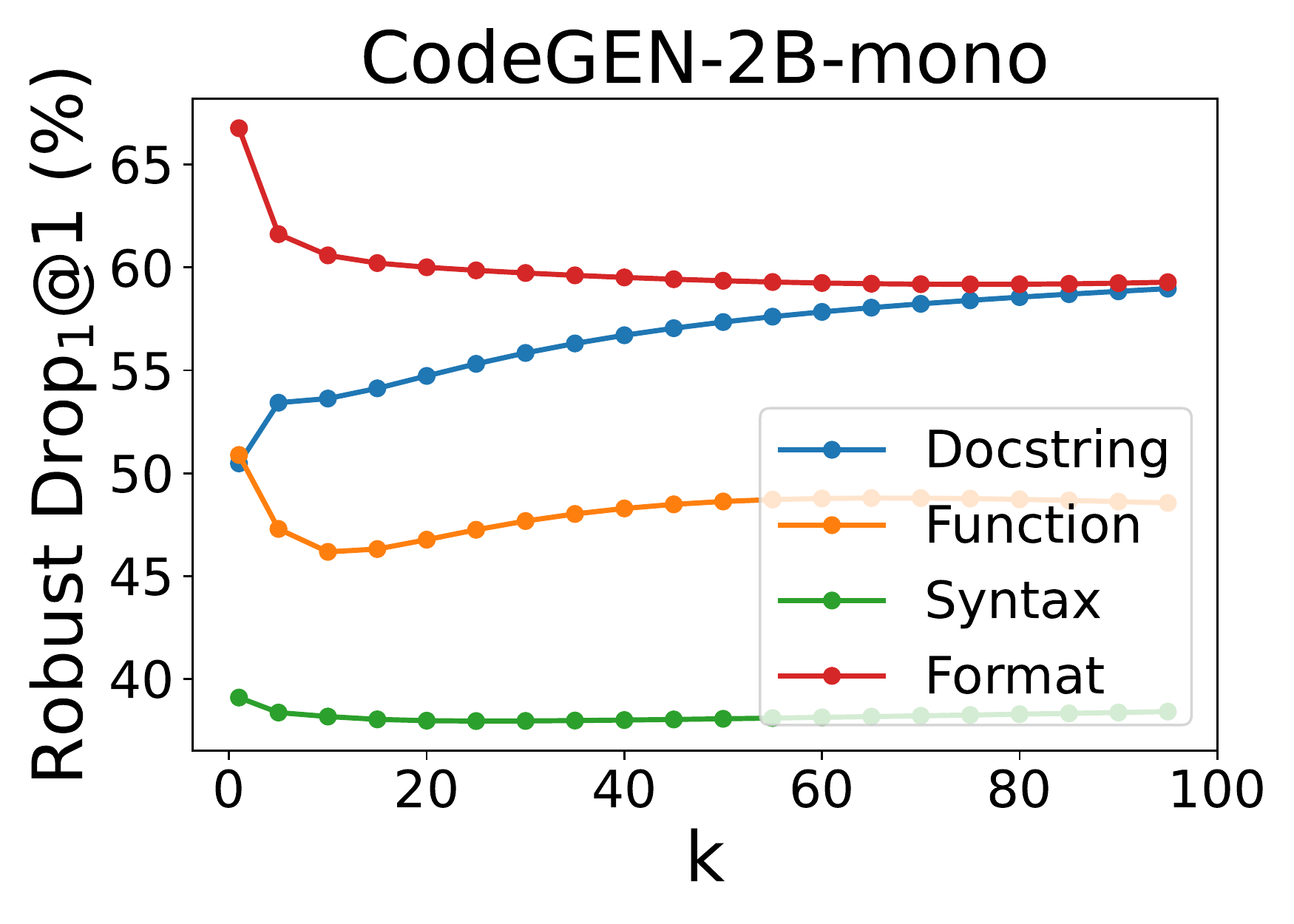}
    \includegraphics[width=0.31\linewidth]{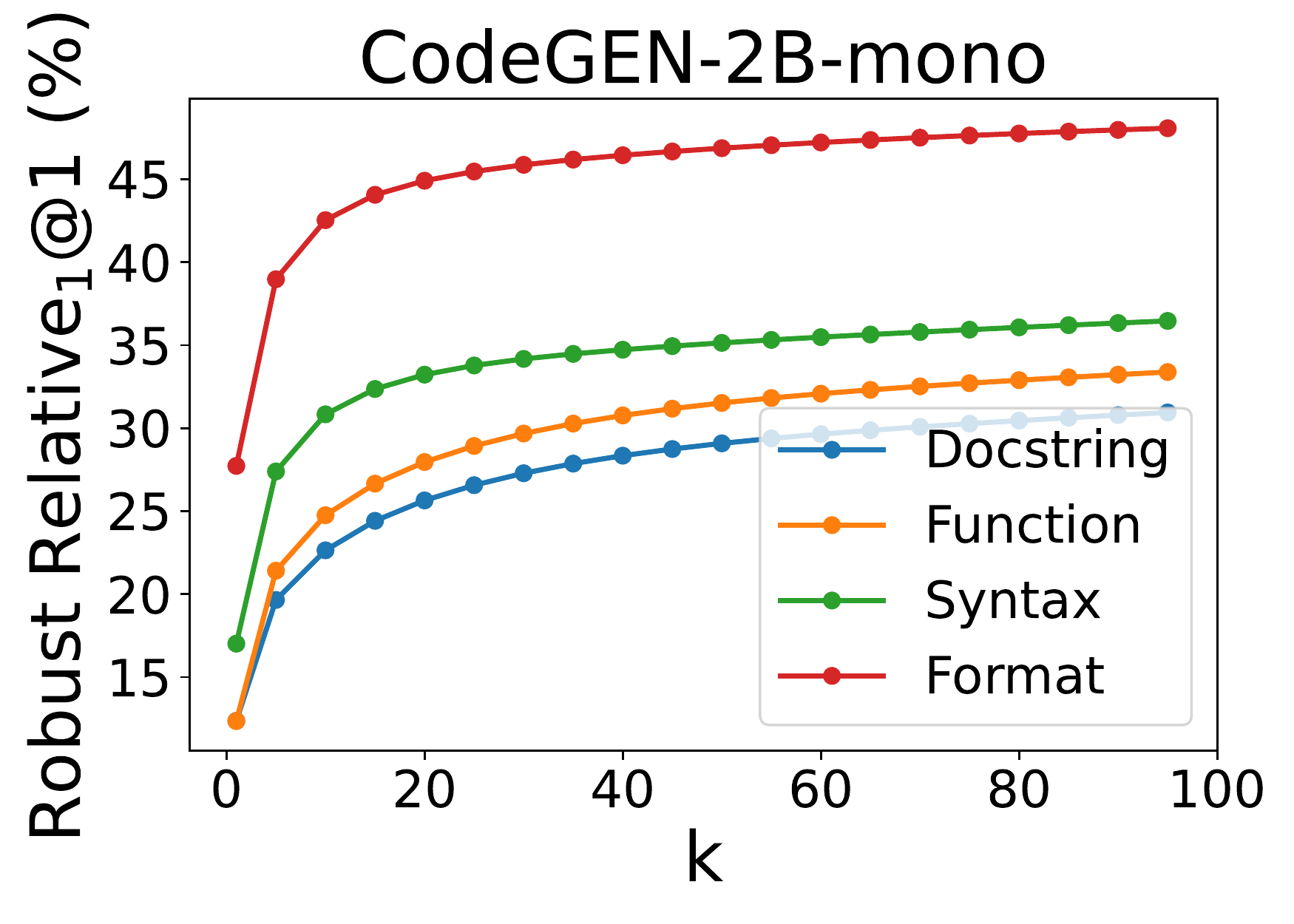}
    
    \includegraphics[width=0.31\linewidth]{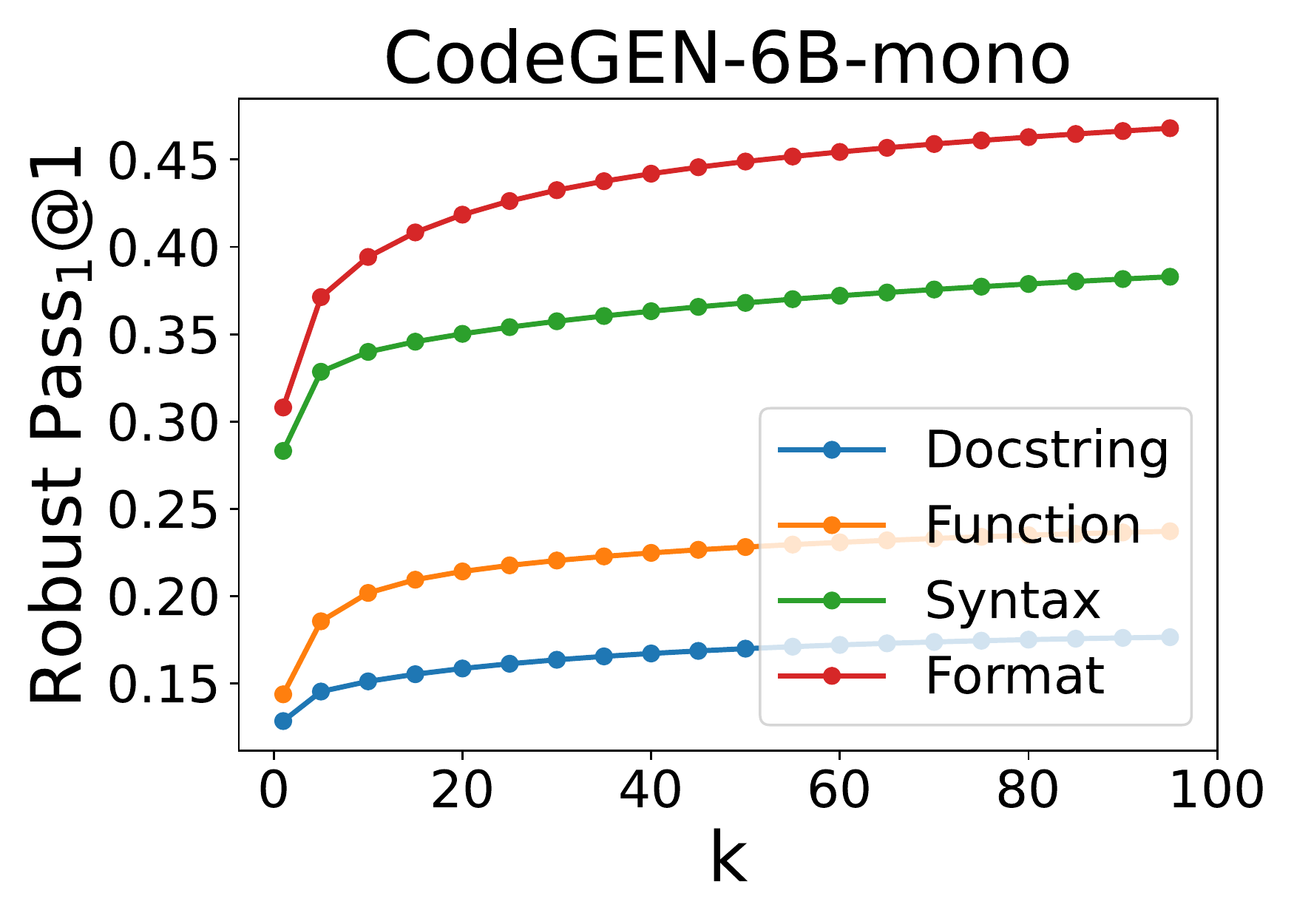}
    \includegraphics[width=0.31\linewidth]{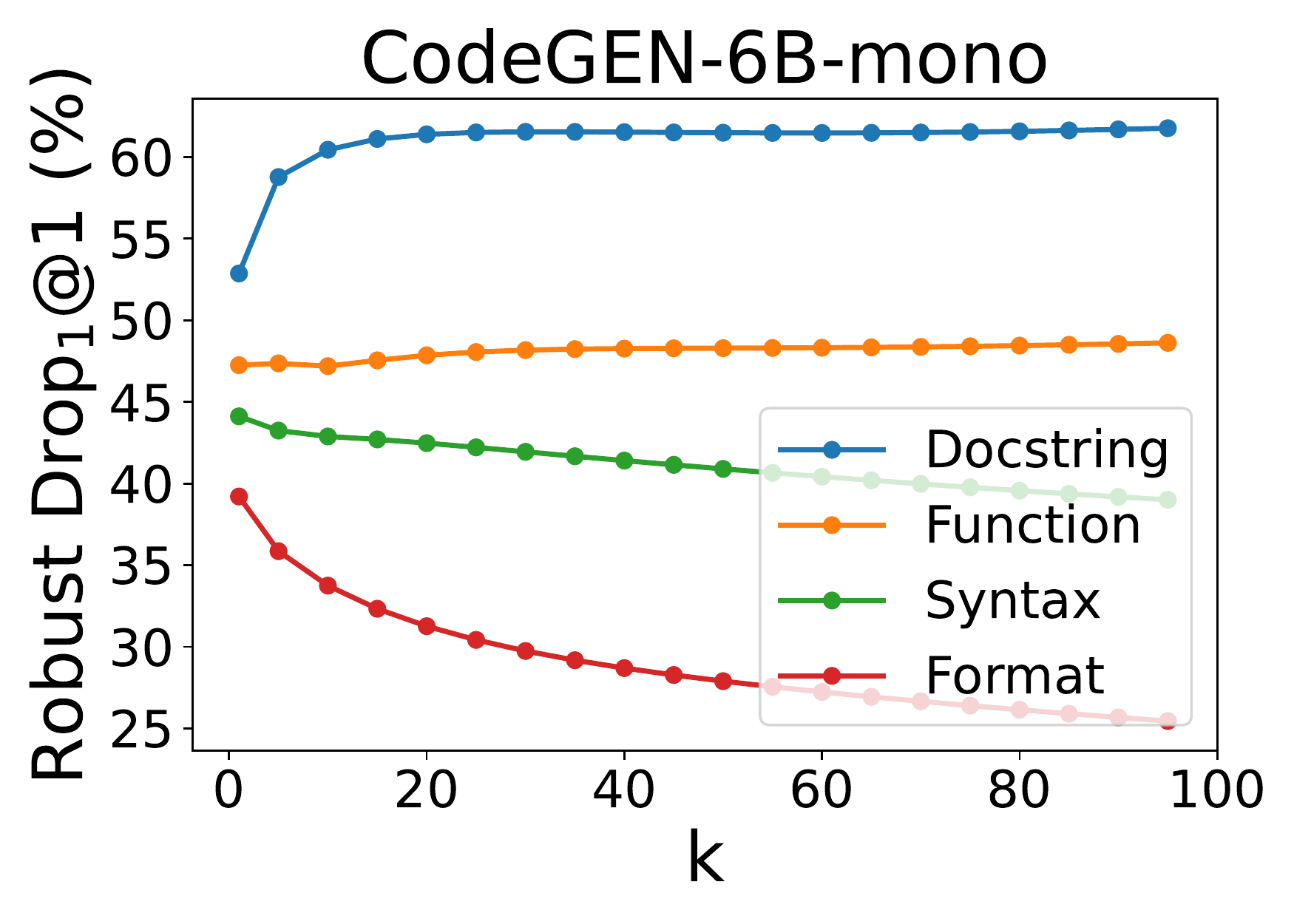}
    \includegraphics[width=0.31\linewidth]{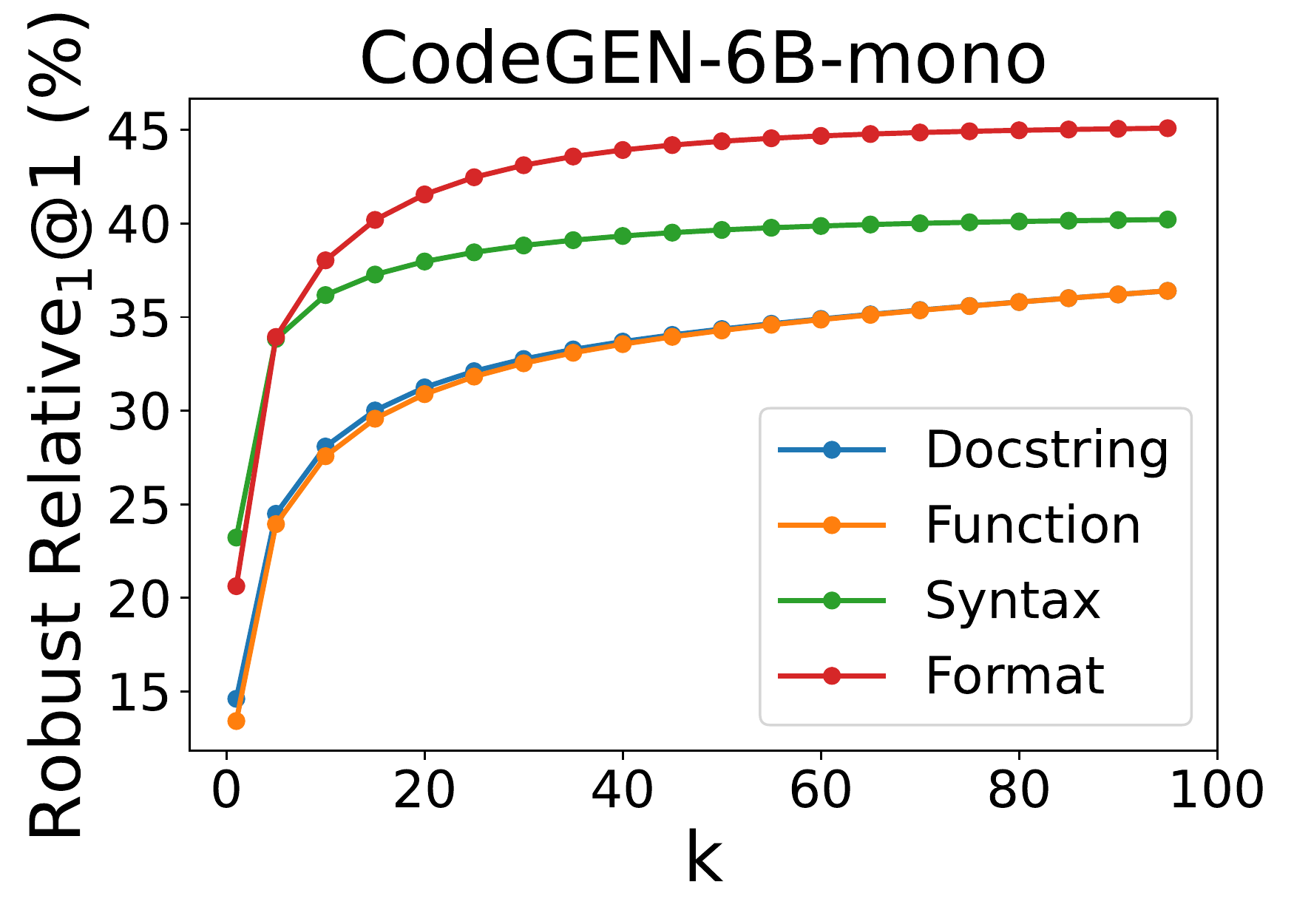}
    
    \includegraphics[width=0.31\linewidth]{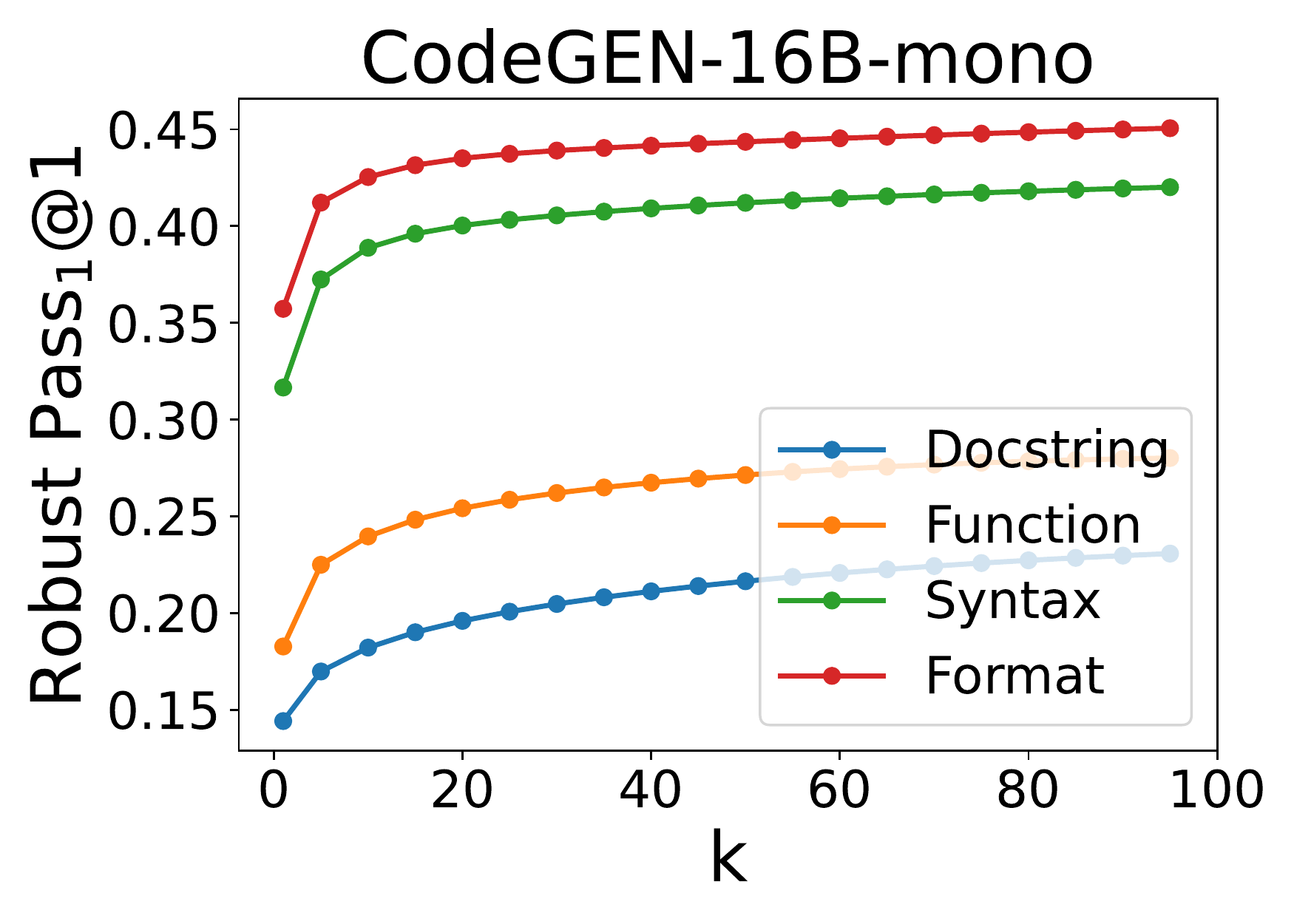}
    \includegraphics[width=0.31\linewidth]{images/ablation_study/CodeGen-16B-mono_drop.pdf}
    \includegraphics[width=0.31\linewidth]{images/ablation_study/CodeGen-16B-mono_relative.pdf}
    \caption{Robust Drop$_1$@1 and Robust Relative$_1$@1 on CodeGen-16B-mono under different $k$ using sampling $n=100$. Robust Drop remains stable while Robust Pass and Robust Relative increases with k.}
    \label{appd: figk}
\end{figure*}

As discussed in~\cref{subsec: ablation}, we observe that Robust Drop stays stable across different k while Robust Relative increases linearly with k. We present additional results on CodeGen-2B-mono, CodeGen-6B-mono along with CodeGen-16B-mono in~\cref{appd: figk}. We evaluate each model with large $n~ (n=100)$ using top-p sampling strategy with probability $0.95$ and temperature $0.2$.

\subsection{Additional Results for Large Sampling n}
\label{appd: n}

Larger sampling $n$ is commonly used for preventing model generation variances and providing accurate estimations. The evaluation cost increases linearly to $n$. Here we show that larger $n$ can also benefit our proposed three robustness metrics but not causing significant differences. In specific, we measure Robust Pass$_1$@1, Robust Drop$_1$@1, and Robust Relative$_1$@1 on CodeGen-16B-mono and HumanEval dataset. The model is run with $n=100$ using top-p sampling strategy with probability $0.95$ and temperature $0.2$. We present detailed results in~\cref{appd: tabn}.

\begin{table*}[ht]
\centering
\footnotesize
\setlength{\tabcolsep}{3pt}
\scalebox{0.95}{
\begin{tabular}{r|r|r|r|r} \toprule
\multicolumn{1}{l}{Category} & \multicolumn{1}{c|}{Metric} & $n=1$   & $n=10$  & $n=100$   \\ \midrule
\multirow{4}{*}{Docstring}                & Nominal$\uparrow$                      & 0.287 & 0.308 & 0.306 \\
                                          & RP$_1$@1$\uparrow$              & 0.128 & 0.140 & 0.143 \\
                                          & RD$_1$@1(\%)$\downarrow$                 & 55.32 & 54.46 & 53.34   \\
                                          & RR$_1$@1(\%)$\downarrow$             & 15.85 & 16.77 & 16.55   \\ \midrule
\multirow{4}{*}{Function}                 & Nominal$\uparrow$                    & 0.287 & 0.308 & 0.306   \\
                                          & RP$_1$@1$\uparrow$              & 0.183 & 0.180 & 0.183   \\
                                          & RD$_1$@1(\%)$\downarrow$                 & 36.17 & 41.39 & 40.37   \\
                                          & RR$_1$@1(\%)$\downarrow$             & 10.37 & 12.99 & 13.24   \\ \midrule
\multirow{4}{*}{Syntax}                   & Nominal$\uparrow$                    & 0.561 & 0.542 & 0.544   \\
                                          & RP$_1$@1$\uparrow$              & 0.220 & 0.234 & 0.244   \\
                                          & RD$_1$@1(\%)$\downarrow$                 & 60.87 & 56.81 & 55.19   \\
                                          & RR$_1$@1(\%)$\downarrow$             & 34.15 & 31.04 & 30.39   \\ \midrule
\multirow{4}{*}{Format}                   & Nominal$\uparrow$                    & 0.561 & 0.542 & 0.544   \\
                                          & RP$_1$@1$\uparrow$              & 0.341 & 0.352 & 0.357   \\
                                          & RD$_1$@1(\%)$\downarrow$                 & 39.13 & 34.98 & 34.36   \\
                                          & RR$_1$@1(\%)$\downarrow$             & 21.95 & 19.70 & 19.36  
\\ \bottomrule
\end{tabular}
}
\caption{Generation variances for three robustness metrics with different sampling $n$ on CodeGen-16B-mono and HumanEval.}
\label{appd: tabn}
\end{table*}